%% file: 25_rohit_scaling.tex
\documentclass{article} % For LaTeX2e
\usepackage{iclr2026_conference,times}

\input{math_commands.tex}

\usepackage{rohit}
\usepackage{pifont}

\usepackage{svg}
\usepackage{subcaption} % put this in the preamble
\usepackage{wrapfig}

\title{A Scalable Distributed Framework for Multimodal GigaVoxel Image Registration}

\author{%
  Rohit Jena$^{*,\dagger}$, Vedant Zope$^{\dagger}$, Pratik Chaudhari, James C. Gee\\
  University of Pennsylvania\\
  Philadelphia, PA 19104
}

\newcommand{\methodname}{{FFDP}}

\definecolor{ultramarine}{RGB}{0,32,96}

\iclrfinalcopy

\begin{document}

% \begingroup
% \renewcommand\thefootnote{\dagger}
% \footnotetext{These authors contributed equally.}
% \endgroup
\maketitle

\renewcommand\thefootnote{\fnsymbol{footnote}}
\footnotetext[1]{Corresponding author: \texttt{rjena@seas.upenn.edu}}
\footnotetext[2]{These authors contributed equally.}
% reset to normal numbering for the rest of the document (if you will use footnotes later)
\renewcommand\thefootnote{\arabic{footnote}}
\setcounter{footnote}{0}

\begin{abstract}
\input{sections/abstract}

\end{abstract}

% --------------------
\vspace*{-13pt}
\input{sections/introduction}
% --------------------
\input{sections/methods}
\input{sections/experiments}



\clearpage
\bibliography{iclr2026_conference}
\bibliographystyle{iclr2026_conference}
\clearpage

\appendix
\crefname{section}{Appendix}{Appendices}
\Crefname{section}{Appendix}{Appendices}

% \section{Appendix}
\input{sections/appendix}

\end{document}

%% file: math_commands.tex
\usepackage{amsmath,amsfonts,bm}

\def\eqref#1{equation~\ref{#1}}
\def\ceil#1{\lceil #1 \rceil}

\def\1{\bm{1}}

\DeclareMathAlphabet{\mathsfit}{\encodingdefault}{\sfdefault}{m}{sl}
\SetMathAlphabet{\mathsfit}{bold}{\encodingdefault}{\sfdefault}{bx}{n}

\DeclareMathOperator*{\argmin}{arg\,min}

%% file: sections/abstract.tex
% Image acquisition techniques and compute capabilities have advanced at an exponential rate in the past three decades, enabling large-scale studies in neuroscience, developmental biology, robotics, geoscience, and other AI-for-science disciplines.
% However, image registration algorithms that form the cornerstone of these applications have not scaled in tandem with these advancements, leading to a significant gap in the ability to \textit{accurately} register images at high resolutions.

In this work, we propose \textbf{\methodname}, a set of IO-aware non-GEMM fused kernels supplemented with a distributed framework for image registration at unprecedented scales.
Image registration is an inverse problem fundamental to biomedical and life sciences, but algorithms have not scaled in tandem with image acquisition capabilities.
Our framework complements existing model parallelism techniques proposed for large-scale transformer training by optimizing non-GEMM bottlenecks and enabling convolution-aware tensor sharding.
We demonstrate unprecedented capabilities by performing multimodal registration of a $100\um$ \textit{ex-vivo} human brain MRI volume at native resolution -- an inverse problem more than 570$\times$ larger than a standard clinical datum in about a minute using only 8 A6000 GPUs.
{\methodname} accelerates existing state-of-the-art optimization and deep learning registration pipelines by upto $6-7\times$ while reducing peak memory consumption by $20-59\%$. 
% Comparative analysis on a 250$\um$ dataset show that {\methodname} performs image registration of datum upto two orders of magnitude larger than existing SOTA on a single GPU without approximations like downsampling or image mosaicing.
Comparative analysis on a 250$\um$ dataset shows that {\methodname} can fit upto 64$\times$ larger problems than existing SOTA on a single GPU, and highlights both the performance and efficiency gains of {\methodname} compared to SOTA image registration methods.

% Results on a simulated 250$\um$ ex-vivo brain MRI dataset show that our method performs deformable registration of images of upto two orders of magnitude larger than existing methods on a single GPU without approximations like downsampling or image mosaicing, leading to upto an improvement of 31.6 points in inverse-weighted Dice score. %, and a $73.7\%$ reduction in Hausdorff distance.
%  -- leading to upto a $112\%$ improvement in inverse-weighted Dice score, and a $73.7\%$ reduction in Hausdorff distance at $250\um$.
% We perform ablation and weak scaling studies, followed by a first-of-its-kind demonstration: multimodal deformable matching to a $100\um$ \textit{ex-vivo} human brain MRI volume at native resolution -- an inverse problem more than 570$\times$ larger than a standard in-vivo MRI datum, with over \textit{11.8 billion} optimizable parameters -- completed in just \textit{under a minute} using only 8 A6000 GPUs.

%% file: sections/introduction.tex
\section{Introduction}
\label{sec:intro}

Image {registration} (also called `image alignment' or `image matching') is a non-linear inverse problem ubiquitous in biomedical and life sciences.
Given $d$-dimensional images $F: \Omega \to \reals^d$ and $M: \Omega \to \reals^d$ defined on domain $\Omega$ (usually a compact subset of $\reals^d$), image registration seeks to find a coordinate transform $\varphi: \Omega \to \Omega$ that deforms the moving image $M$ to look similar to the fixed image $F$. 
Mathematically, we minimize the following objective (\cref{fig:imagereg}):
\begin{equation}
    \varphi^* = \argmin_{\varphi \in G} L(\varphi) \doteq C(F, M \circ \varphi) + R(\varphi)
    \label{eq:image-reg}
\end{equation}
where $C$ is a cost or dissimilarity function, and $\circ$ is the interpolation operator, i.e. $(I \circ g) (x) = I(g(x))$ for all $x \in \Omega$. 
% In this paper, we consider $d=3$ but higher dimensional $d$ is possible, i.e. time-series data. 
Popular choices of $\varphi$ are affine and deformable transforms, i.e. $\varphi(x) = Ax + t$, and $\varphi(x) = x + u(x)$.
Modern registration pipelines \citep{synthmorph,fireants} consider an affine matching followed by a deformable matching step, resulting in a composite transform $\varphi(x) = Ax + t + u(x)$.
% We assume that the fixed image has $N$ voxels.
$u$ is called the displacement field, modeled as a grid of per-voxel vectors $u(x) \in \reals^d$. 
For an image of size $N$, the displacement field is a tensor of size $dN$.
We use $\mathbf{[x]}_\Omega$, $A\mathbf{[x]}_\Omega+t$, and $\mathbf{[u]}_\Omega$ to denote the identity grid, grid of affine transformed coordinates, and deformation grid defined on $\Omega$ respectively.
% totalling to a tensor of size $dN$ for a $d$-dimensional fixed image of size $N$.
% The transformation can belong to an algebraic group, say $G$, whose elements $g \in G$ act on the image by transforming the domain as $(I \circ g) (x) = I(g(x))$ for all $x \in \Omega$. The registration problem solves for
% where $C$ is a cost function, e.g., that matches the pixel intensities of the warped image with those of the fixed image, or local normalized cross-correlation or mutual information of image patches. 
Common choices of $C$ are mean squared error, Localized Normalized Cross Correlation \citep{avants_symmetric_2008}, and Mattes Mutual Information \citep{mattesmi}.
Common choices of $R$ include Sobolev norm of the gradient or warp fields \citep{beg2005computing,claire,avants_symmetric_2008-1}, total variation, and inverse-consistency~\citep{christensen2001consistent}.
To optimize \cref{eq:image-reg}, iterative methods optimize $\varphi^*$ directly using gradient descent, and deep learning methods learn a deep neural network $\varphi = f_\theta(F, M)$.
%total variation, , or volume-preserving~\cite{haber2004numerical}. 
% 
{Image registration establishes a common coordinate system, aligning scans across individuals and atlases \citep{hering2022learn2reg,oasis,murphy2011evaluation}. 
This alignment is a prerequisite for multimodal data fusion, cross-subject comparison, morphometric analysis \citep{das2009registration}, and construction of large-scale atlases \citep{allen}. 
Establishing such voxelwise correspondence is fundamental for studying anatomical variability, detecting pathological signatures \citep{Ravikumar2021}, and advancing precision medicine \citep{borner2022tissue, jonsson2022image}.
The saliency and centrality of the task across various biomedical and life science applications has spurred numerous methodological advances in the field, spanning more than three decades of research \citep{gee1993elastically,unigradicon}.
% Given its central role across biomedical and life science applications, registration has long been recognized as a core computational challenge, driving sustained methodological innovation for more than three decades \citep{gee1993elastically,unigradicon}.
% However, the scaling of these algorithms has not kept pace with the advancements in image acquisition capabilities, leading to a significant gap in the ability to accurately register images at high resolutions.
% Most biomedical workflows require a robust and accurate registration framework

% \begin{wrapfigure}{R}{0.8\linewidth}
\begin{figure}[t!]
    \centering
    \includegraphics[width=\linewidth]{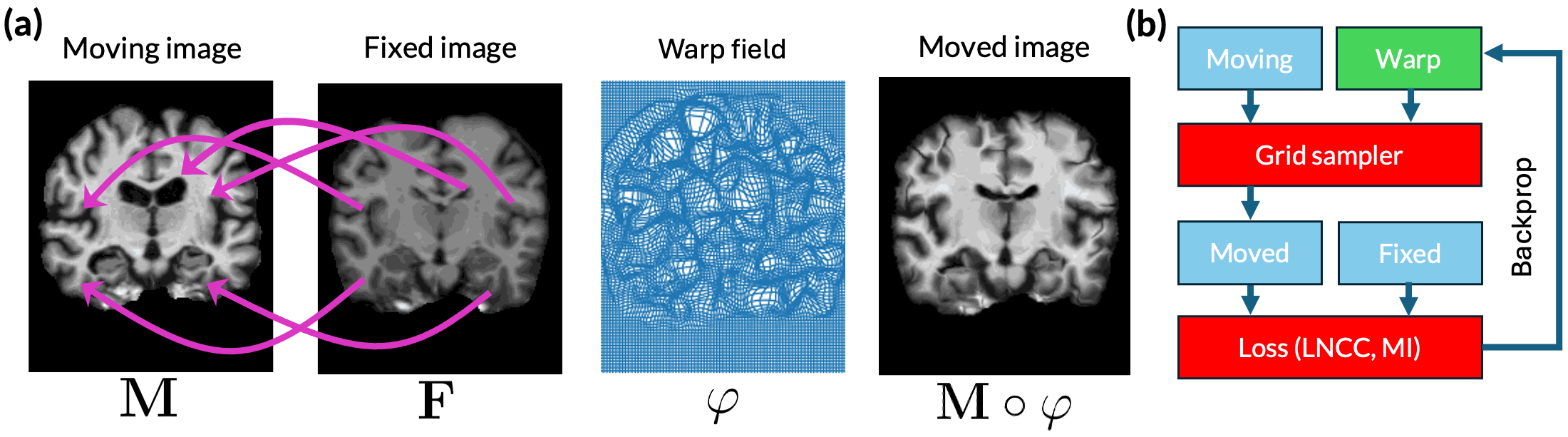}
    \caption{ 
    \textbf{Image Registration Problem.} \textbf{(a)}: The task is to find a coordinate transform that warps the moving image $M$ to the fixed image $F$. Individual corresponding points are shown as \textcolor{violet}{{violet}} arrows; the per-pixel coordinate transform is shown as a warp field $\varphi$, and the transformed image $M \circ \varphi$.
    \textbf{(b)}: A typical registration pipeline - the grid sampler warps the moving image, that is then compared to the fixed image using a loss function. \textcolor{ForestGreen}{\textbf{Green}} denotes the optimizable warp, \textcolor{red}{\textbf{red}} denotes the primary bottlenecks that we optimize in this paper.
    }
    \label{fig:imagereg}
\end{figure}

% The recent years have witnessed a tremendous growth in image acquisition capabilities for various biomedical and life science applications, including MRI \citep{ultrahighfield}, CT \citep{photoncountingct}, PET \citep{pet}, microscopy \citep{ultramicroscopy,wassie2019expansion}.
% The recent years have witnessed 
Over the past decade, there has been 
a tremendous growth in image acquisition capabilities for various biomedical and life science applications, including MRI, CT, PET, microscopy \citep{ultrahighfield,photoncountingct,pet,ultramicroscopy,wassie2019expansion}.
Ultra-high resolution imaging technology has enabled acquisition of images beyond three orders of magnitude larger than macroscopic biomedical domains \citep{waxholm,allen,dukeatlas,kleinfeld2011large}.
For instance, a typical clinical scan registration requires solving $\sim 20$M parameters, while a high-resolution ex-vivo human brain scan requires solving upto $11$B parameters.
However, current approaches work reliably only at the scale of \textit{macroscopic} biomedical domains ($\sim$ 50M warp parameters) 
% \citep{synthmorph,transmorph,vfa} 
and quickly run out of memory on larger problems due to high computational and memory requirements.
This leads to a significant gap in the ability to accurately register images at high resolutions.

Recent years have also witnessed tremendous innovations in large-scale transformer model training, including IO-aware fused operations to minimize latency and large memory overheads\citep{flashattn,flashattn2,thunderkittens}, 5D parallelism to distribute larger-than-memory models and inputs into multi-GPU/node setups\citep{megatronlm,sequenceparallel,sequenceparallel2,distflashattn,fsdp2scaling,torch2}.
Although most existing techniques are specialized for generalized matrix multiplication (GEMM) like operations only, 
the fundamental concepts utilized by these methods (IO-awareness, recomputing and aggregating intermediates on shared memory to minimize high bandwidth memory (HBM) storage, identifying partial aggregates across hosts to minimize communication overheads for distributed optimization) are broadly applicable to a wide class of problems of the non-GEMM nature. 

% Image registration in particular represents a complementary scaling challenge: its core operators are not dense matrix multiplies but voxelwise stencil and interpolation kernels, with highly irregular memory access and strong locality constraints. Thus, adapting these concepts requires rethinking how they apply to spatial, non-GEMM workloads. While transformer training scales in terms of network depth and parameters, registration scales in terms of image resolution and spatial size — often billions of voxels per scan — making it an equally pressing testbed for distributed algorithm design.

In this paper, we apply these concepts to scale image registration algorithms to match parity with the developments in both increasing resolution of image acquisition \textit{and} compute capabilities.  
% We revamp critical workhorse components that form the cornerstone of a variety of biomedical, clinical, life science, geoscience and robotics applications. 
% Our goal in this paper is twofold - 
To that end, our contributions are twofold.
First, we identify key compute and memory bottlenecks in image registration algorithms, and propose novel components 
that fit problems upto $64\times$ larger than existing algorithms on a single GPU. 
% Second, we propose a distributed framework by extending the proposed components to scale algorithms to an arbitrary number of GPUs, thereby scaling to arbitrary problem sizes. 
Second, we propose \textbf{F}lash \textbf{F}used \textbf{D}istributed \textbf{P}rimitives (\methodname), a distributed framework to scale registration to an arbitrary number of GPUs, thereby scaling to ultra high-resolution problems.
We present a first-of-its-kind demonstration: aligning a 250$\um$ in-vivo MRI \citep{lusebrink2017t1} to a $100\um$ \textit{ex-vivo} 
% single-echo FLASH volume of a human brain 
human brain FLASH volume 
\citep{edlow20197} -- a multimodal registration problem more than \textbf{570$\times$} larger than a standard clinical datum \citep{oasis}, with over 11.8B transform parameters -- completed in \textit{one minute} using only 8 A6000 GPUs.
{\methodname} accelerates existing traditional registration pipelines by upto $7.48\times$ while reducing memory consumption by upto $59\%$, and deep learning pipelines by upto 6.14$\times$ while consuming upto $24\%$ less memory.
We highlight the necessity of performing high-resolution registration by comparing our method with various SOTA optimization and deep learning baselines on a 250$\um$ T1-weighted MRI dataset, showing unprecedented performance and gains in efficiency.
% We show the efficacy of our method by comparing with various optimization and deep learning baselines on a 250$\um$ dataset, and showing upto 1000$\times$ efficiency gains.
% Finally, we demonstrate a first-of-its-kind result by mapping two $100\um$ \textit{ex-vivo} human brain MRI images -- a problem more than 570$\times$ the size of a OASIS brain subject on only \todo{xx} GPUs in only \todo{xx} minutes.
% Finally, 

%% file: sections/methods.tex
% \vspace*{-15pt}
\section{Fused Kernels for Memory Efficient Registration on a Single GPU}
\vspace*{-5pt}
\label{sec:prelim}

\paragraph{Bottlenecks of a deformable image registration pipeline}
Our primary objective is to identify compute and memory bottlenecks in \textit{large-scale} image matching tasks.
In identifying these bottlenecks, training-free optimization methods are better suited than deep networks since the latter has a much larger activation memory footprint, which forms the primary memory bottleneck \citep{ultrascaleplaybook}.
% For deep methods, the activation memory forms the primary memory bottleneck \citep{ultrascaleplaybook}.
For instance, for a 250$\um$ image pair, a standard deep learning method \citep{synthmorph} generates an activation map of size 27GB only after the first layer.
Extrapolating memory usage for clinical data, existing deep networks will require upto 1.2TB of GPU memory at inference to process these image volumes at native resolution.
In contrast, a training-free optimizer can fit this problem in less than 45GB of GPU memory. 
We use FireANTs ~\citep{fireants} as our base framework to identify compute and memory bottlenecks in a typical image registration problem.
We analyze the flamegraph of a typical clinical MRI registration task from the \textit{OASIS} brain dataset \citep{oasis} in \cref{fig:flamegraph}.
We identify three key memory bottlenecks in image matching pipelines (1) deformable interpolation and warp composition (2) cross-correlation loss, and (3) mutual information loss (see \cref{fig:overview}(right)). 
\footnote{A GPU’s memory hierarchy spans multiple tiers: registers (per-thread, single-cycle), shared memory/L1 cache (on-chip, tens of KB, low latency within a block), L2 cache (MBs, shared across SMs, moderate latency), and global memory (\textbf{HBM}).
Our work focuses on reducing HBM usage for key non-GEMM operations used in image registration, by maximizing register and shared memory usage while minimizing global memory traffic.
% Performance depends on maximizing register and shared memory usage while minimizing global memory traffic.
}
We first propose efficient designs to fit larger problems on a single GPU, and then extend the framework to distributed registration.

\input{figures/overview}

% \textbf{GPU memory hierarchy}

\vspace*{-5pt}
\subsection{Composite Implicit Grid Sampler}
\vspace*{-5pt}
\label{sec:gridsampler}
A fundamental operation used in image registration is the \textit{grid sampler}. %, which resamples the image given a grid of coordinates.
This operator allows us to warp an image $M$ using a deformation field $\varphi: \Omega \rightarrow \Omega$ and computes the image $M': M'(x) = M(\varphi(x))$.
% This operator is used to optimize affine, deformable and joint transforms.
Virtually every image registration pipeline uses this operation to warp the moving image using an affine, deformable, or composite transform.
For affine and composite transforms, the operator initializes a regular grid $\gridx_\Omega$, a grid of size $3N$. 
The affine grid $A\gridx_\Omega + t$ is another grid of size $3N$. 
If a deformable grid $\gridu_\Omega$ is optimized, then a third grid $A\gridx_\Omega + t + \gridu_\Omega$ is materialized, costing a total of $9N$ overhead for an image of size $N$.
% 
% To alleviate this, we propose a generalized implicit grid sampler operator which performs the operation
% To save this overhead, 
To consolidate these memory overheads, we propose a composite implicit grid sampler.
This is a fused CUDA kernel that performs the following operation:
$$\text{\texttt{fused\_grid\_sampler}}(I; A, t, \gridu, S, x_{\text{bounds}})(x) = I(Ax + t + Su(x))$$
where $A, S \in GL(d, \reals)$ are affine matrices, $t$ is a translation vector, $\gridu$ is the deformation grid, and $x_{\text{bounds}}$ are the bounds of the (implicit) identity grid $\gridx_\Omega$.
There are three benefits of this approach.
% First, if only an affine matrix $A$ is optimized, then we reduce an overhead of $6N$ since we never materialize the grids $\gridx_\Omega$ and $A\gridx_\Omega+t$.
% Second, when a deformation field is added, the per-pixel transformation $Ax + t + u(x)$ is computed within the CUDA kernel threads directly without materializing another grid of size $3N$.
% 
First, the kernel avoids materializing any additional grids in HBM, reducing the memory overhead of the kernel from $O(n)$ to $O(1)$ with no loss in runtime or accuracy.
Second, when the warp $\gridu_\Omega$ is sharded across hosts in a distributed setting, the identity grid $[\mathbf{x}]_\Omega$ needs to be sharded correctly too.
Since the identity grid is implicitly defined by its bounds  $x_{\text{bounds}} = (x_{\min}, x_{\max}) \in \mathbb{R}^{2d}$, our implementation can be easily used in a distributed optimization setting without instantiating partial shards $\gridx_\oh$.
% Finally, the matrix $S$ may look redundant since the deformation field is generally not multiplied with any independent affine component. 
Finally, the matrix $S$ is used to rescale the deformation field to sample from the coordinates of the sharded images $I_h$ which lie on the grid $\oh$ instead of $\Omega$ (see \cref{sec:ringsampler-rescale}) without initializing additional memory.
The backward pass is very similar to the existing PyTorch implementation, with the exception of the gradient of the affine matrix. 
We discuss the derivation and pseudocode of the forward and backward pass in the \cref{app:gridsampler}.

\input{figures/distributed}

\vspace*{-5pt}
\subsection{Implicit Parzen Windowing for Mutual Information}
\vspace*{-5pt}
% % Global Mutual Information (MI) loss is one of the most commonly used loss functions for \textit{multimodal} image registration\todo{make this better}. 
Mattes Mutual Information (MI) is one of the most commonly used loss functions for \textit{multimodal} image matching \citep{transmorph,ants,mattesmi}.
% Beyond multimodal image matching, MI is a cornerstone operation in computer vision \citep{miboundary,miregion}, contrastive learning \citep{micontrastive}, remote sensing \citep{miremotesensing}, graph learning \citep{migraph}, ecological and social community interactions \citep{mi_network,mi_network_corso}, and cosmological dynamics \citep{mi_galaxy}.
% 
For random variables $X$ and $Y$, MI is the KL divergence between the joint distribution $P(X,Y)$ and product of marginal distributions $P(X)P(Y)$ of the intensities of the two images.
For image matching, $X$ and $Y$ are the pixel intensities for the images $I, J$.
The distributions are estimated using a kernel density estimator:
\begin{align}
P_I(v) = \frac{1}{N}\sum_k \kappa(v - I_k), \quad P_{(I,J)}(v, w) = \frac{1}{N}\sum_{k} \kappa(v-I_k)\kappa(w - J_k) \label{eq:mi}
\end{align}
where $\kappa$ is a kernel function of choice.
Common choices of $\kappa$ are the Gaussian \citep{mattesmi_gaussian} and 3rd order B-Spline kernels \citep{mattesmi_bspline}.
% To compute the mutual information between the two variables, the distributions $P_I(v), P_J(w)$ and $P_{(I,J)}(u,w)$ are discretized at $B$ equally spaced bins from $[0,1]$.
To empirically compute the KL divergence, the distributions \cref{eq:mi} are discretized over $B$ equally spaced bins on the domain of $u \in I, v \in J$.
However, to compute the joint histogram of size $B^2$, this method requires materializing the entire Parzen Block $\Psi_I(j, k) = \kappa(b_j - I_k)$ of size $\textcolor{red}{2k_PBN}$, where $k_P$ is a kernel-dependent constant.
% Here, $k_P$ is a constant containing intermediate variables depending on the chosen kernel ($k_{\text{Gaussian}} = 4, k_{\text{BSpline}} = 14$).
Since $N >> B$ ($B$ is typically chosen to be 32), this operation becomes a significant memory bottleneck for large $N$.
For instance, a typical clinical image volume ($N \approx 30\text{MB}$) with 32 bins will consume \textbf{7.5GB} of HBM - a significantly huge cost that grows much faster for larger problems.

% We make the implementation efficient by leveraging
Our efficient implementation leverages
the fact that $B$ is small to avoid materializing the tensors $\Psi_I, \Psi_J \in \reals^{B\times N}$ altogether and use high-throughput shared memory to compute and accumulate the histogram entries and partial gradients for each image pixel.
We provide the detailed derivation in \cref{app:fusedmi}.
This leads to an efficient implementation that consumes $O(1)$ additional HBM instead of $O(N)$ (holding $B$ constant).
This leads to upto \textbf{98\%} lesser HBM usage for images considered in our experiments, and an asymptotic $100\%$ reduction in HBM usage for large images (\cref{fig:ablations}(top-right)). %, while being upto an order of magnitude faster than the vanilla PyTorch implementation.

\vspace*{-5pt}
\subsection{Efficient Implicit Fused Cross-Correlation}
\vspace*{-5pt}
% Local Normalized Cross-Correlation is used ubiquitously for intramodal image registration.
Local Normalized Cross-Correlation (LNCC) is used ubiquitously in signal and image processing as a similarity metric.
In deformable image registration, it is used as a robust similarity function to compare anatomical similarities \citep{transmorph,synthmorph,avants_symmetric_2008-1,wu2024neural}. 
% 
% Most LNCC implementations suffer from two major memory-bound bottlenecks.
% First, the loss function initializes many intermediate variables.
Most LNCC implementations are memory-bound due to the large number of intermediate variables.
Our analysis in \cref{app:fusedcc} shows that the computational graph adds {16}$\times$ HBM overhead, and upto another $16\times$ HBM overhead for computing gradients with respect to all intermediates.
% For large images, this becomes a huge memory bottleneck.

% Most of the arithmetic forward pass can be fused inside a kernel. 
To avoid these huge memory overheads, we fuse all the intermediate computation in a fused kernel.
Our fused forward pass requires only $5\times$ memory for storing all intermediates ($I, J, I^2, J^2, IJ$ convolved with matrix $w$).
In \cref{app:fusedcc-definition} we analytically derive the gradient and show that the input gradients can be computed by modifying the saved intermediates \textit{in-place}.
% , saving an requiring only $5\times$ memory for the forward and backward pass combined.
This leads upto a \textbf{76.5\% reduction} in memory (see \cref{tab:cc-ablation-table}) and outperforms even \texttt{torch.compile} implementations.
% We ablate on the memory requirements and runtime for various implementations in \cite{sec:exp}.

% \subsection{Interleaved CPU offloading}

\vspace*{-5pt}
\section{Extending image registration to multiple GPUs}
\vspace*{-5pt}
Our composite implicit grid sampler and improved loss functions allows optimizing problems with image sizes that are upto two magnitudes larger than other baselines on a single A6000 GPU (\cref{tab:exp1_allmethods_grouped}).
% However, many applications in life sciences like \todo{cite} require registration on images that do not fit on a single GPU.  
However, many applications using mesoscopic and microscopic data require registration of images that do not fit on a single GPU.
% In the following section, we propose
Inspired by distributed frameworks for LLM training \citep{megatronlm,zero} and initial work on distributed image registration \citep{claire}, we propose a distributed framework that allows sharding large images across multiple GPUs to efficiently scale to arbitrarily large problem sizes with any similarity loss function. % without approximating the optimization objective itself.
% We propose 
% a distributed framework that allows sharding large images across multiple GPUs to efficiently scale to arbitrarily large problem sizes without approximating the optimization objective itself.
% that can be solved on multiple GPUs without approximating the optimization itself.  

\textbf{Distributed Setting.}
For distributed registration with $H$ hosts or GPUs, we partition the domain $P(\Omega) = \{\Omega_1, \Omega_2, \ldots \Omega_H\}$ such that $\left|\Omega_i\right| = N/H$, $\Omega_i \cap \Omega_j = \phi \quad\forall i \ne j$ and $\large{\cup_{i}} \Omega_i = \Omega$.
We use $\mathbf{[x]}_\oh$, $A\mathbf{[x]}_\oh+t$, and $\mathbf{[u]}_\oh$ to denote the sharded tensors defined on domain $\oh$.

\subsection{Grid Parallel for Boundary-Synchronized Image Sharding}
Techniques like Tensor/Sequence/Expert/Context Parallel have been tremendously successful in distributed optimization by sharding large models and sequences across multiple GPUs \citep{megatronlm,sequenceparallel,ringattention,deepseekv3}.
However, these techniques work for transformer-like architectures and input sequences where the model parameters and activations do not require boundary synchronization.
In contrast, image registration contains operations that require boundary synchronization between image and grid shards to perform mathematically correct convolutions.
Examples of such operations include convolutions for calculating LNCC, total variation loss, Sobolev norm of the gradient and warp fields \citep{claire,avants_symmetric_2008-1,beg2005computing}.
% These operations require synchronized boundary values along the sharded dimensions.

To enable these functionalities and complement existing parallelism techniques, we propose `\textit{Grid Parallel}' (GP) as an abstraction on a tensor.
% GP partitions and shards a tensor equally across hosts, stores the sharded dimension and boundary coordinates as metadata, and provides synchronization operations to augment the image with sufficient boundary padding from neighboring shards prior to performing a convolution operation. 
GP shards a tensor across hosts, stores the sharded dimension and bounds as metadata, and provides synchronization operations to augment the tensor with sufficient boundary padding from neighboring shards prior to performing a convolution operation.
GP allows us to partition the fixed images, $\gridu$, and the optimizer state $\mathbf{[m_1]}, \mathbf{[m_2]}$ -- essentially sharding the entire problem across $H$ hosts while allowing the user to apply convolutional operations seamlessly. 
We compare the performance of GP with naive DTensor sharding in \cref{app:gridparallel}.
% We analyse the correctness of GP in \cref{todo}.

\input{figures/ringsampler}

\input{tables/faux-oasis-results}

\subsection{Distributed Ring Sampler}
Despite the sharding in GP, the moving image $M$ cannot be sharded across GPUs due to the random-access nature of the \texttt{grid\_sample} operation applied on $M$.
In general, the warp vector $\varphi(x)$ residing on GPU $i$ can point to coordinates that reside on the sharded image on GPU $j$ for any $j \ne i$. 
Even for neighboring coordinates $x_s, x_u \in \gridx_i$, the coordinates $\varphi(x_s)$ and $\varphi(x_u)$ can point to different shards $j_1 \ne j_2 \ne i$.
This is illustrated in \cref{fig:ring-sampler}(a).
Keeping the entire moving image in memory limits the maximum problem size to $N \le V$, where $V$ is the memory per GPU, regardless of the number of hosts $H$.
However, we want the maximum problem size to scale with $H$.
Therefore, we propose a distributed \texttt{grid\_sampler} that allows us to \textit{correctly} interpolate the moving image with sharded images scattered across multiple hosts without performing an \texttt{allgather} operation on the moving image.

% 
% If the total memory requirement for a registration problem is $\xi N$ where $\xi$ is a multiplier for intermediate variables, warp field, and the optimizer state, then the memory per GPU is $N_i = \left(1 + \frac{(\xi-1)}{H}\right)N$.
% Asymptotically, the largest problem that can fit on infinite hosts with per-host memory of $V$ is $\lim_{H \rightarrow \infty }N_i \le V \implies N \le V$. However, to fit arbitrarily large problems, we require $\lim_{H \rightarrow \infty }N_i = 0$. 
% Therefore, we must extend the grid sampler operation to interpolate on sharded images scattered across multiple hosts without performing an \textit{all-gather} operation on the image.
% to be compatible with sharded moving images.

% We propose an implementation that 
% Our approach tackles both these problems with only an additional $\frac{N}{H}$ memory overhead in the worst case.
Our approach leverages the key observation that (bi/tri)linear interpolation can be decomposed as an aggregate of partial sums of interpolated values on individual image shards.
\cref{fig:ring-sampler}(b) illustrates this example.
These individual image shards are sent across hosts in a ring topology, similar to \cite{ringattention}, and the partial sum is aggregated inplace. %, interweaving the \texttt{grid\_sample} operation on the shard and sending and fetching the next image shard.  
This operation only incurs an additional $N/H$ HBM overhead for fetching the sharded image from other hosts, scaling efficiently to arbitrary large problem sizes for sufficiently large $H$.
% 
% The value of the pixel, which depends on the nearest neighbors residing on different GPUs.
% , wherein the interpolated value depends on pixels that potentially reside on different GPUs. 
% These partially interpolated values can be obtained by fetching the image shards from other GPUs, applying the \texttt{grid\_sample} operation, and accumulating the results.
% Similar to \cref{ringattention}, we use a ring topology to pass image shards and aggregate the interpolated value. 
% 
The detailed derivation and correctness of this operation is shown in \cref{app:ringsampler}.

\vspace*{-5pt}
\subsection{Distributed Loss Functions}
\vspace*{-5pt}
% Certain loss functions require specialized treatment for loss functions.
Since the moved image and fixed image are sharded cross $H$ hosts, the loss function must take this into account to compute the loss function correctly. 

\textbf{Mean Squared Error (MSE)}. Since MSE is a per-pixel loss, we compute the individual MSE on host $h$ and perform an \texttt{allreduce} operation. 

\textbf{Localized Normalized Cross Correlation (LNCC)}. 
The LNCC computes per-pixel patch similarities for each pixel, using a convolution over its neighbors.
For sharded images, the patch statistics at the boundary requires a boundary synchronization with its neighboring shards which is provided by our GP implementation.
After computing the LNCC for all pixels in each shard, we perform another \texttt{allreduce} to compute the LNCC over the entire image.

\textbf{Mutual Information (MI)}. The MI loss computes the joint histograms $p_{(I,J)}(x, y)$ and marginals $p_I(x), p_J(y)$. However, these distributions are partial aggregates from the sharded images on each GPU. 
\cref{eq:mi} can be rewritten as $p_I(v) = \sum_h \frac{N_h}{N} \textcolor{red}{\left(\frac{1}{N_h}\sum_{k \in \Omega_h} \kappa(v - I_k)\right)}, p_{IJ}(v, w) = \sum_h \frac{N_h}{N} \textcolor{red}{\left(\frac{1}{N_h}\sum_{k \in \Omega_h} \kappa(v - I_k)\kappa(w - J_k)\right)}$, where the \textcolor{red}{red} terms correspond to the per-host histogram computation.
Performing an \texttt{allreduce} to compute the weighted average of these histograms (with weights $N_h/N$) results in a valid and correct joint and marginal distributions over all hosts. 
This also leads to only a $B^2 + 2B$ communication overhead regardless of $N$, making a distributed implementation highly practical.  

%% file: figures/overview.tex
\begin{figure}[t!]
    \centering
    \begin{minipage}{0.79\linewidth}
        %\includesvg[width=\linewidth]{images/overview-fireants} % no extension
        \includegraphics[width=\linewidth]{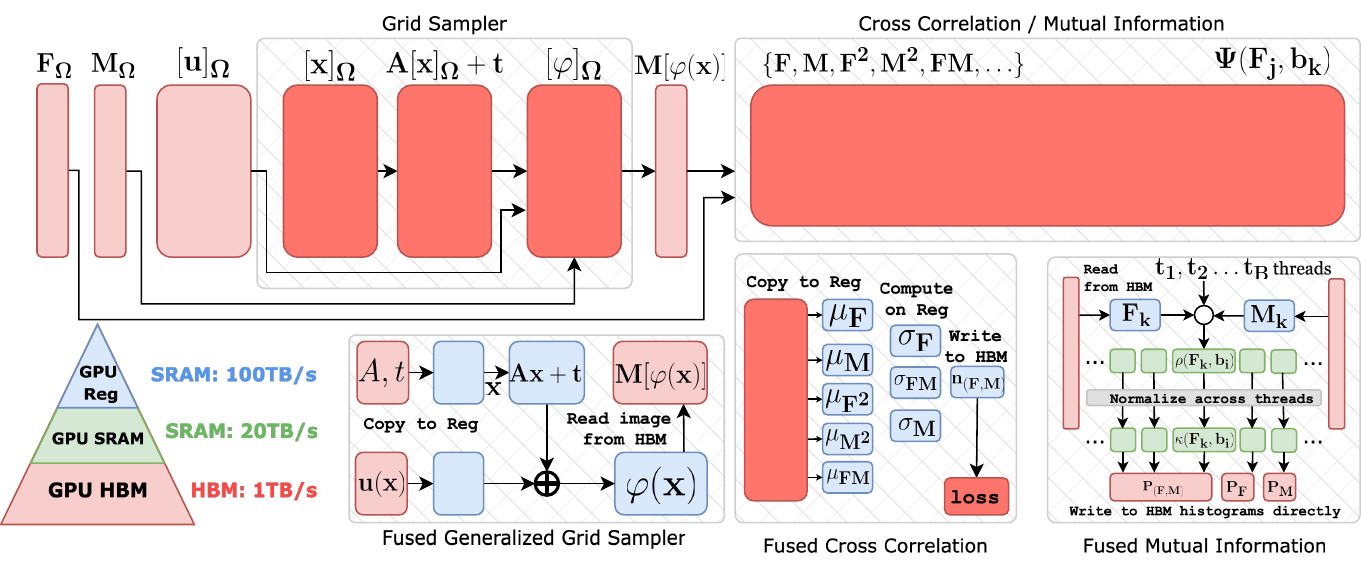} % no extension
    \end{minipage}
    \hfill
    \begin{minipage}{0.20\linewidth}
        \includegraphics[width=\linewidth]{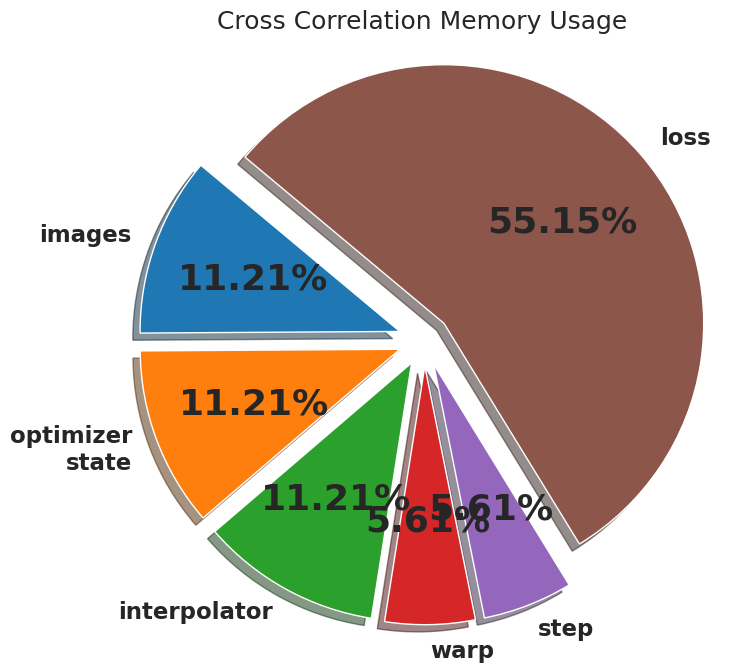}
        \vspace{0.5em}
        \includegraphics[width=\linewidth]{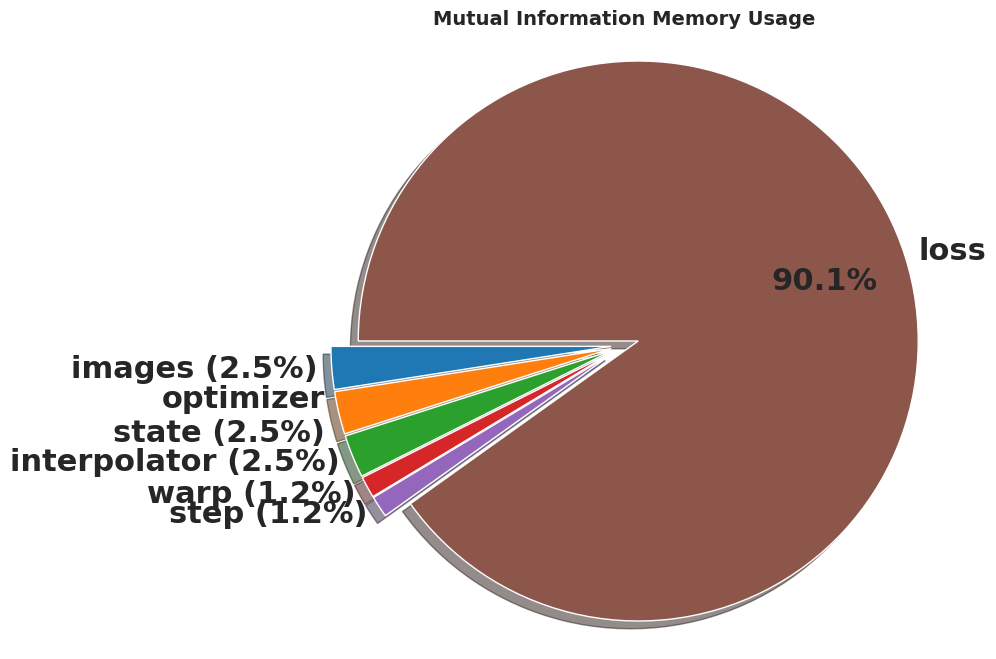}
    \end{minipage}
    \caption{\textbf{Left}: {\methodname} uses fused kernels to eliminate intermediate HBM memory usage (in {\textcolor{red}{\textbf{dark red}}}) for memory-bound workhorse operations (\texttt{grid\_sampler}, LNCC, MI) for large-scale image registration. For \texttt{grid\_sampler} and LNCC, additional intermediate per-pixel variables (warp coordinates, patchwise statistics) are computed per-pixel in registers (\textcolor{blue}{{blue}}). For MI, the Parzen Windowing and histogram aggregation is performed using shared memory (\textcolor{ForestGreen}{{green}}), avoiding large HBM overheads. \textbf{Right}: Pie charts show the breakdown of memory overheads for storing the image, grid, optimizer state, and intermediate variables for MI and LNCC losses.
    }
    \label{fig:overview}
\end{figure}

%% file: figures/distributed.tex
% \begin{wrapfigure}{R}{0.8\textwidth} % r = right, l = left
\begin{figure}[t!]
    \centering
    \includegraphics[width=0.64\linewidth]{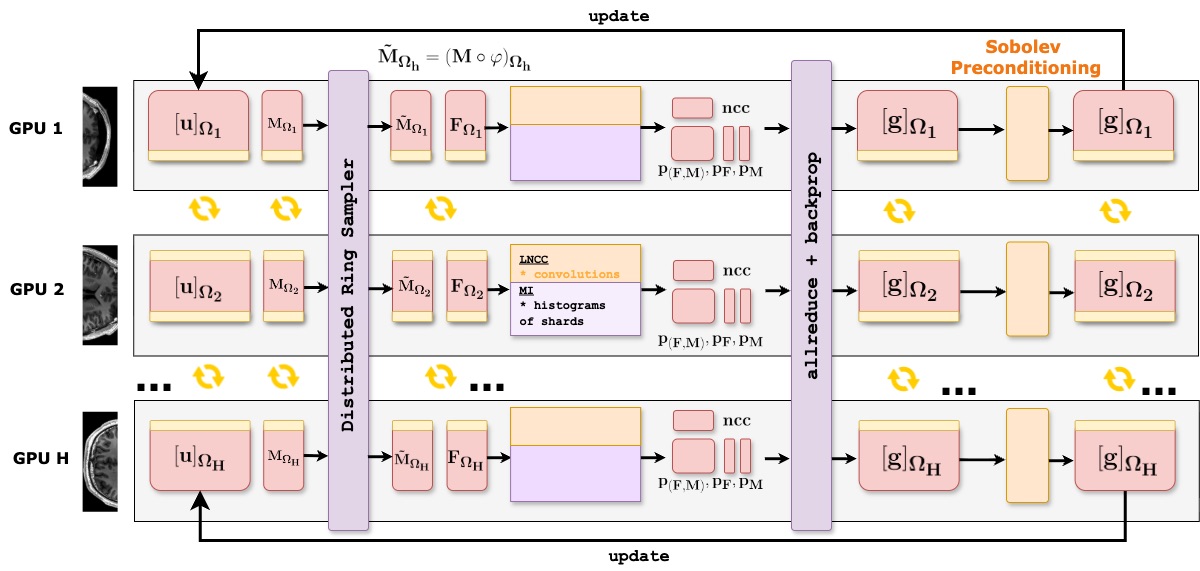}
    \includegraphics[width=0.35\linewidth]{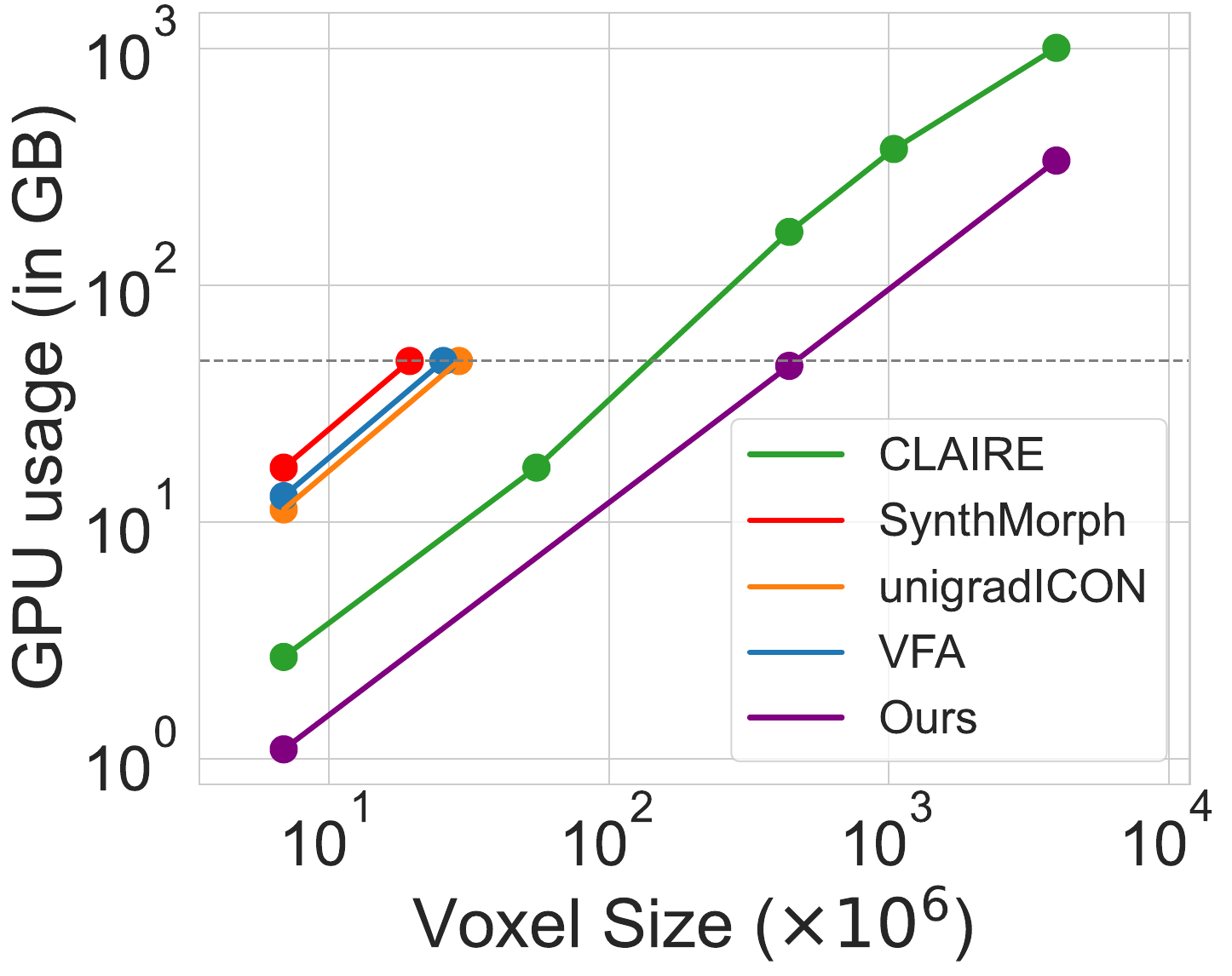}
    \caption{\small \textbf{Left}: Overview of our distributed framework. GridParallel (GP) shards the fixed and moving images $(F,M)$ and the warp field $\gridu$ across multiple GPUs.
    \textcolor{Dandelion}{Yellow} blocks and arrows denote synchronized halo boundaries between GPUs, enabling smoothing on images and warp fields without an allgather.
    The ring sampler (\textcolor{violet}{violet}) computes interpolated image shards on the fly, avoiding materialization of the full moving image.
    We then compute losses (MSE, LNCC, MI), compute gradients w.r.t. each warp shard, apply \textcolor{Dandelion}{Sobolev regularization} with GP, and update shards by gradient descent.
    \textbf{Right}: Scaling efficiency compared to deep methods and CLAIRE \citep{claire}, a distributed registration method. Most SOTA deep learning baselines require orders-of-magnitude more memory for the same problem size and scalability is limited to a single GPU (dotted line).
    Our framework scales to arbitrarily large problem sizes while using about $5\times$ less memory than CLAIRE. %and three orders of magnitude faster than CLAIRE.}
    }
    \label{fig:distributed}
% \end{wrapfigure}
\end{figure}

%% file: figures/ringsampler.tex
% \begin{wrapfigure}{L}{0.65\textwidth}
% % \begin{figure}[h]
%     \centering
%     % \includegraphics[width=\linewidth]{images/ringsampler.png} % no extension
%     % \includegraphics[width=0.8\linewidth]{images/ringsampler_v2.png} % no extension
%     \includegraphics[width=\linewidth]{images/ringsampler_v2.png} % no extension
%     \caption{Ring Sampler overview}
%     \label{fig:ring-sampler}
% % \end{figure}
% \end{wrapfigure}

\begin{figure}[t!]
    \centering
    \includegraphics[width=\linewidth]{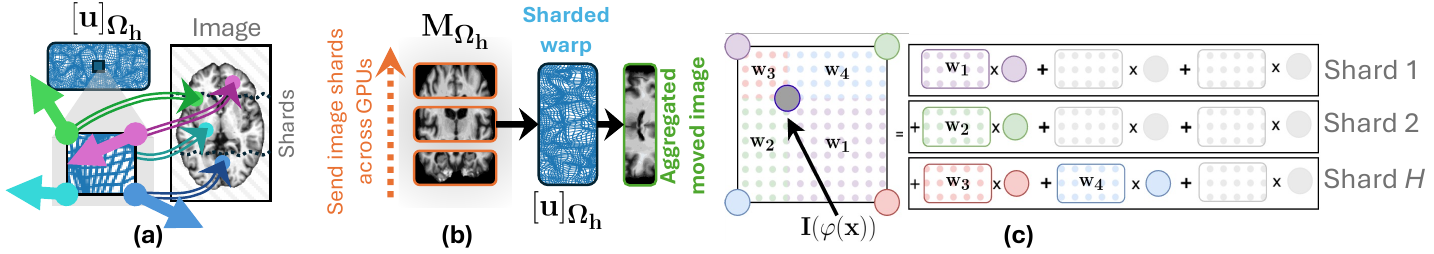} % no extension
    \caption{
    \textbf{(a)} Neighboring coordinates in the warp field may refer to pixel locations on arbitrary image shards due to the deformable nature of the warp field, making distributed interpolation non-trivial.
    \textbf{(b)} Ring Sampler interleaves fetching of \textcolor{orange}{image shards} and aggregating the \textcolor{ForestGreen}{partial sums} of interpolated values, avoiding a memory-expensive allgather.
    \textbf{(c)} Bilinear Interpolation is decomposed into partial sums over image shards, which are accumulated with a ring topology communication, similar to \citet{ringattention}.
    }
    \label{fig:ring-sampler}
\end{figure}

%% file: tables/faux-oasis-results.tex
% start
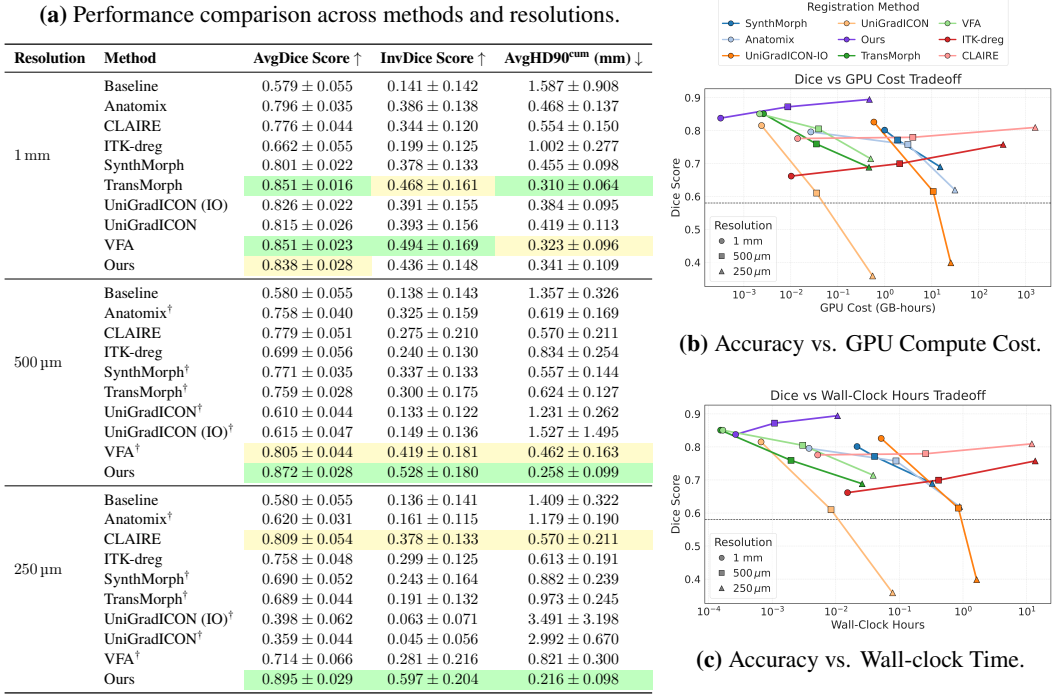
\begin{figure}[t!]
\label{tab:exp1-results}

\begin{minipage}{0.62\linewidth}
  % \begin{table*}[t]
  % \begin{wraptable}{L}{0.6\linewidth}
  \begin{subfigure}[t]{\linewidth}
  \centering
    \caption{Performance comparison across methods and resolutions.}
    \renewcommand{\arraystretch}{1.15}
    \resizebox{\linewidth}{!}{%
    \begin{tabular}{l
                    l
                    S[table-format=1.3(1)]
                    S[table-format=1.3(1)]
                    S[table-format=2.3(1)]}
      \toprule
      \textbf{Resolution} & \textbf{Method} &
        {\textbf{AvgDice Score}~$\uparrow$} &
        {\textbf{InvDice Score}~$\uparrow$} &
        {\textbf{AvgHD90$^{\text{cum}}$ (mm)}~$\downarrow$} \\
      \midrule
      \multirow{8}{*}{\SI{1}{\milli\meter}}
        & Baseline              & \num{0.579 +- 0.055} & \num{0.141 +- 0.142} & \num{1.587 +- 0.908} \\
        & Anatomix\gpus{1}              & \num{0.796 +- 0.035} & \num{0.386 +- 0.138} & \num{0.468 +- 0.137} \\
        & CLAIRE\gpus{1}             & \num{0.776 +- 0.044} & \num{0.344 +- 0.120} & \num{0.554 +- 0.150} \\
        & ITK-dreg\gpus{CPU}              & \num{0.662 +- 0.055} & \num{0.199 +- 0.125} & \num{1.002 +- 0.277} \\
        & SynthMorph\gpus{1}            & \num{0.801 +- 0.022} & \num{0.378 +- 0.133} & \num{0.455 +- 0.098} \\
        & TransMorph\gpus{1}            & \best{\num{0.851 +- 0.016}} & \second{\num{0.468 +- 0.161}} & \best{\num{0.310 +- 0.064}} \\
        & UniGradICON (IO)\gpus{1}      & \num{0.826 +- 0.022} & \num{0.391 +- 0.155} & \num{0.384 +- 0.095} \\
        & UniGradICON\gpus{1}           & \num{0.815 +- 0.026} & \num{0.393 +- 0.156} & \num{0.419 +- 0.113} \\
        & VFA\gpus{1}                   & \best{\num{0.851 +- 0.023}} & \best{\num{0.494 +- 0.169}} & \second{\num{0.323 +- 0.096}} \\
        & Ours\gpus{1}         & \second{\num{0.838 +- 0.028}} & \num{0.436 +- 0.148} & \num{0.341 +- 0.109} \\
      \midrule
      \multirow{8}{*}{\SI{500}{\micro\meter}}
        & Baseline          & \num{0.580 +- 0.055} & \num{0.138 +- 0.143} & \num{1.357 +- 0.326} \\
        & Anatomix\oom\gpus{1}          & \num{0.758 +- 0.040} & \num{0.325 +- 0.159} & \num{0.619 +- 0.169} \\
        & CLAIRE\gpus{1}              & \num{0.779 +- 0.051} & \num{0.275 +- 0.210} & \num{0.570 +- 0.211} \\
        & ITK-dreg\gpus{CPU}           & \num{0.699 +- 0.056} & \num{0.240 +- 0.130} & \num{0.834 +- 0.254} \\
        & SynthMorph\oom\gpus{1}        & \num{0.771 +- 0.035} & \num{0.337 +- 0.133} & \num{0.557 +- 0.144} \\
        & TransMorph\oom\gpus{1}        & \num{0.759 +- 0.028} & \num{0.300 +- 0.175} & \num{0.624 +- 0.127} \\
        & UniGradICON\oom\gpus{1}       & \num{0.610 +- 0.044} & \num{0.133 +- 0.122} & \num{1.231 +- 0.262} \\
        & UniGradICON (IO)\oom\gpus{1}  & \num{0.615 +- 0.047} & \num{0.149 +- 0.136} & \num{1.527 +- 1.495} \\
        & VFA\oom\gpus{1}               & \second{\num{0.805 +- 0.044}} & \second{\num{0.419 +- 0.181}} & \second{\num{0.462 +- 0.163}} \\
        & Ours\gpus{1}         & \best{\num{0.872 +- 0.028}} & \best{\num{0.528 +- 0.180}} & \best{\num{0.258 +- 0.099}} \\
      \midrule
      \multirow{8}{*}{\SI{250}{\micro\meter}}
        & Baseline          & \num{0.580 +- 0.055} & \num{0.136 +- 0.141} & \num{1.409+-0.322} \\
        & Anatomix\oom\gpus{1}          & \num{0.620 +- 0.031} & \num{0.161 +- 0.115} & \num{1.179 +- 0.190} \\
        & CLAIRE\gpus{4}              & \second{\num{0.809 +- 0.054}} & \second{\num{0.378 +- 0.133}} & \second{\num{0.570 +- 0.211}} \\
        & ITK-dreg\gpus{CPU}              & \num{0.758 +- 0.048} & \num{0.299 +- 0.125} & \num{0.613 +- 0.191} \\
        & SynthMorph\oom\gpus{1}        & \num{0.690 +- 0.052} & \num{0.243 +- 0.164} & \num{0.882 +- 0.239} \\
        & TransMorph\oom\gpus{1}        & \num{0.689 +- 0.044} & \num{0.191 +- 0.132} & \num{0.973 +- 0.245} \\
        & UniGradICON (IO)\oom\gpus{1}  & \num{0.398 +- 0.062} & \num{0.063 +- 0.071} & \num{3.491 +- 3.198} \\
        & UniGradICON\oom\gpus{1}       & \num{0.359 +- 0.044} & \num{0.045 +- 0.056} & \num{2.992 +- 0.670} \\
        & VFA\oom               & \num{0.714 +- 0.066} & \num{0.281 +- 0.216} & \num{0.821 +- 0.300} \\
        & Ours\gpus{1}         & \best{\num{0.895 +- 0.029}} & \best{\num{0.597 +- 0.204}} & \best{\num{0.216 +- 0.098}} \\
      \bottomrule
    \end{tabular}
    }
    \label{tab:exp1_allmethods_grouped}
  \end{subfigure}
  % \end{table*}
  % \end{wraptable}
\end{minipage}
\hfill
\begin{minipage}{0.37\linewidth}
  \input{figures/fauxoasis-efficiency.tex}
\end{minipage}

\caption{
% Evaluation across resolutions at native \SI{250}{\micro\meter}. 
% Mean~$\pm$~std across pairs. 
% Within each resolution block, the best is highlighted using 
% \colorbox{green!25}{\strut\ } (green) and the second best using 
% \colorbox{yellow!25}{\strut\ } (yellow). 
% $\uparrow$ higher is better; $\downarrow$ lower is better. 
% Patch-based inference is marked with \oom, non-patch-based inference is left unmarked, 
% and the number of GPUs used is indicated where applicable.
Registration performance on Faux-OASIS dataset at \SI{1}{\milli\meter}, \SI{500}{\micro\meter}, and \SI{250}{\micro\meter} (native \SI{250}{\micro\meter}); mean~$\pm$~std over pairs. 
$\uparrow$ higher is better; $\downarrow$ lower is better. 
HD90 values are reported using our cumulative definition (see Sec.~\ref{subsec:hd90}). 
% Full-context methods like FireANTs+{\methodname}, CLAIRE, and ITK-DReg show improvement in performance with resolution, while all deep learning methods degrade in performance.
\colorbox{green!25}{\strut\ }(Green)/\colorbox{yellow!25}{\strut\ }(Yellow) = best/second; \oom = patch-based %; GPU count is annotated.
% {\methodname} leads at \SI{500}{\micro\meter} and \SI{250}{\micro\meter} across Dice/InvDice/HD90, while VFA/TransMorph are strongest at \SI{1}{\milli\meter}. Green/yellow = best/second; 
}

% end
\end{figure}

%% file: figures/fauxoasis-efficiency.tex
% \begin{figure}[t!]
    \centering
    % --- Subfigure A: GPU Compute Cost ---
    \begin{subfigure}[t]{0.99\linewidth}
        \centering
        \includegraphics[width=\linewidth,height=0.42\textheight,keepaspectratio]{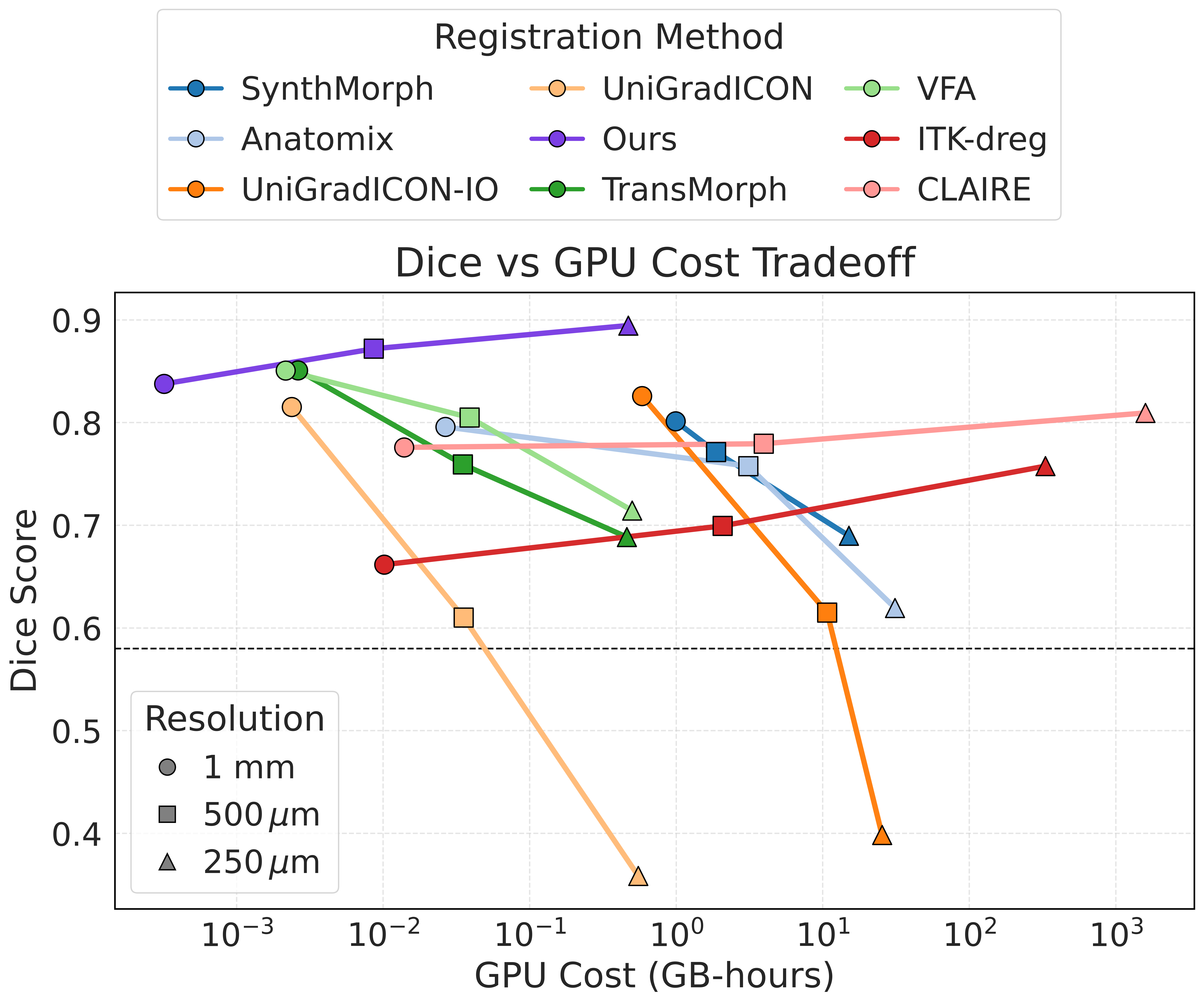}
        \caption{Accuracy vs. GPU Compute Cost.}
        \label{fig:acc_gbhours}
    \end{subfigure}
    
    \vspace{0.6em} % add vertical breathing space

    % --- Subfigure B: Wall-clock Time ---
    \begin{subfigure}[t]{0.99\linewidth}
        \centering
        \includegraphics[width=\linewidth,height=0.28\textheight,keepaspectratio]{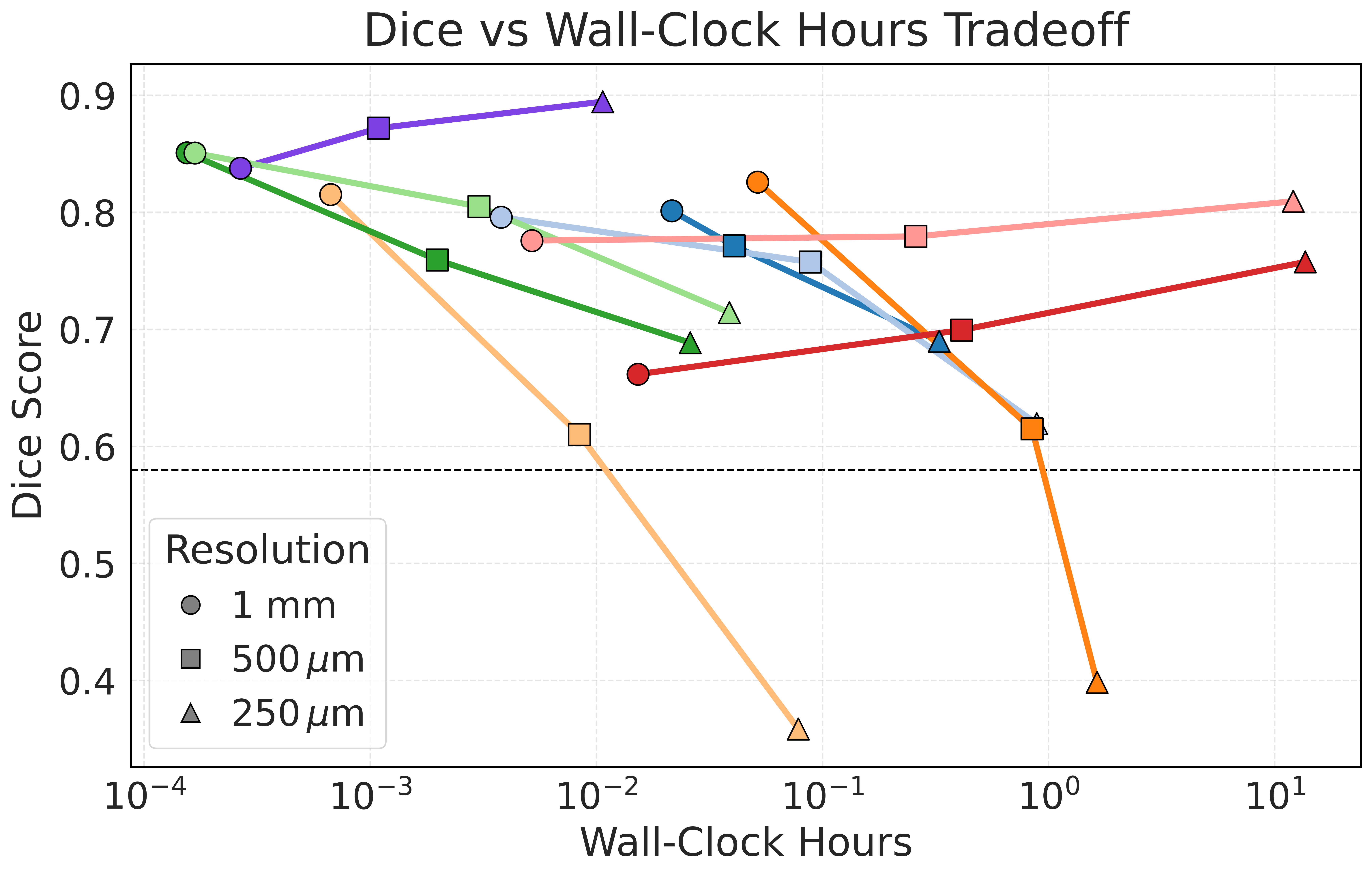}
        \caption{Accuracy vs. Wall-clock Time.}
        \label{fig:acc_time}
    \end{subfigure}
    \label{fig:acc_vs_cost}
% \end{figure}

%% file: sections/experiments.tex
\input{figures/qualitative_experiment}

\section{Experiments}
% We present experiments to demonstrate the efficacy of our method.
Our primary goals are to 
(a) accelerate both optimization and neural network based registration workflows, and
(b) solve significantly larger image registration problems.
% (c) scale the registration objective to arbitrary problem sizes by using a multi-GPU distributed framework.
% To this end, we compare our method against various state-of-the-art baselines on a simulated 250$\um$ ex-vivo brain MRI dataset, both in performance and computational efficiency.
% We demonstrate the speedups in both optimization and deep learning training pipelines with our fused kernels.
% We benchmark the individual components of our method against widely used implementations in the community, study the weak scaling properties of our distributed framework, and present qualitative results on various high-resolution applications.
% We ablate the correctness of our GridParallel implementation in the Appendix.
We show the efficacy of our method by accelerating existing registration workflows on standard clinical data. 
This is followed by optimizing a multimodal registration task with more than 11.8B optimizable parameters, an unprecedented result in large-scale registration.
% a first-of-its-kind demonstration to our knowledge.
We compare the performance and computational efficiency of our method with various state-of-the-art baselines on a simulated 250$\um$ ex-vivo brain MRI dataset, followed by ablations on various components of our framework.
% This is followed by a demonstration of our method on a first-of-its-kind large-scale registration problem: registering a 250$\um$ in-vivo MRI image to a 100 $\um$ \textit{ex-vivo} FLASH human brain volume.

\input{tables/transmorph-speedup}

\paragraph{Baselines.}
To accelerate existing registration workflows, we compare against TransMorph \citep{transmorph} and FireANTs \citep{fireants}, which are state-of-the-art deep learning and optimization based registration frameworks respectively.
In addition, we perform comparative evaluation with two methods explicitly designed for large-scale registration: ITK-DReg \citep{itkdreg} (CPU-based) and CLAIRE \citep{claire} (multi-GPU), and several SOTA learning-based approaches for clinical data - SynthMorph \citep{synthmorph}, Vector-Field Attention \citep{vfa}, unigradICON \citep{unigradicon} (with/without instance optimization), anatomix+ConvexAdam \citep{anatomix}.

\subsection{Accelerating existing registration workflows and ablations}
\label{sec:accelerating}

\input{figures/scaling}

% We evaluate the impact of our fused operations in both deep learning and classical optimization settings. 
For deep networks, we train \texttt{TransMorph-large} under three loss configurations: (a) LNCC+Dice, (b) MI+Dice, and (c) LNCC+scaling-and-squaring \citep{ashburner2007fast} +Dice. 
For classical optimization, we benchmark runtime and memory against multiple LNCC backends (FireANTs, VoxelMorph/TransMorph, Fast LNCC, \texttt{torch.compile}, and Ours) and MI backends (PyTorch and Ours with and without \texttt{torch.compile}).
\cref{tab:transmorph-speedup,tab:fireants-speedup,fig:transmorph-speedup} show that during network training our kernels converge $6.1\times$ faster with LNCC while using 16.5\% less memory, and reduce MI memory usage by 24.7\%. 
Despite being designed for very large images, the runtime and memory benefits are significant for clinical-scale data (i.e., 30MB for OASIS). 
Optimization frameworks see larger gains: FireANTs achieves up to 95.2\% memory savings and $2.6\times$ speedup with MI, and a $7.5\times$ speedup over FastLNCC \citep{xijiafastlncc} (and $2.9\times$ over FireANTs' LNCC backend which applies separable convolutions on FastLNCC), with 44-59\% lower memory usage overall.

\subsection{Registration to a 100 micron ex-vivo brain MRI volume}
\label{sec:edlow-250um}
To showcase the efficacy of our method on real large scale images, we register a 250$\um$ in-vivo MRI image \citep{lusebrink2017t1} to a 100 $\um$ \textit{ex-vivo} FLASH human brain volume \citep{edlow20197}.
This represents an inverse problem with more than 11.2B optimizable parameters (compared to $\sim$20M for clinical datasets), or 44.8GB of GPU memory. % -- barely fitting on a single A6000 GPU.
The entire problem does not fit on most GPUs, necessitating distributed multimodal registration.
We optimize a composite transform - affine followed by a diffeomorphic mapping; details can be found in \cref{app:edlow-250um}.
Multimodal deformable registration took $\sim$58 seconds on 8 NVIDIA A6000 GPUs, which is unprecedented at this resolution. % to our knowledge.
\cref{fig:qualtitative_registration_comparison} shows qualitative results, highlighting the ability to register highly detailed structures such as cerebellar white matter; these structures are not visible at macroscopic scales.
The resultant advantages of performing registration at this scale can allow researchers to characterize the neuroanatomy at microscopic resolutions and allow morphometric analysis of cortical layers and subcortical nuclei among other structures.

Registration accuracy in these studies is measured using privately annotated fiducial markers, hindering reproducibility and comparability of methodological advances.
Due to lack of scalable frameworks, most high-resolution studies simply run ANTs at a significantly downsampled resolution \citep{waxholm,dukeatlas,allen,devccf,drosophilatemplate,edlow20197} and upsample the warp field to the native resolution.
% To provide a thorough comparative analysis, we compare existing SOTA on a high-resolution simulated dataset.

% High-resolution life science studies lack comparative benchmarks, partly due to (a) the difficulty of handling and annotating data at scale, and (b) the absence of widely adopted evaluation protocols. Therefore, these studies simply run ANTs or Elastix at a downsampled resolution \citep{waxholm,dukeatlas,allen,devccf,drosophilatemplate,edlow20197}.
% Existing high-resolution studies on mice, \textit{Drosophila}, and {zebrafish} do not show comparative studies due to the lack of annotated data at that scale, and simply run ANTs or Elastix at a downsampled resolution.
% We tackle this problem by using a simulated dataset that mimics the anatomical distribution of the OASIS dataset at $250\um$ resolution.

% \subsection{Fitting larger problems}
\subsection{Comparative Analysis on a Simulated ex-vivo Brain MRI Dataset}

\textbf{The faux-OASIS dataset}
To compare registration performance at high resolutions and leverage existing methods as baselines, we synthesize the \textit{faux-OASIS} dataset, which mimics the anatomical distribution of an MRI dataset at $250\um$ isotropic resolution (more details in \cref{app:faux-oasis}).
% Owing to a surface-based generation pipeline, the dataset contains finer subcortical structures and GM-WM boundaries, making high-precision registration challenging. 
% This dataset generates finer subcortical structures and GM-WM boundaries, making high-precision registration challenging. 
At $250\um$, the deformation field has 1.32B degrees of freedom per image pair, compared to $\sim$20M for OASIS.
\input{figures/loss-ablations}

\textbf{Baselines and evaluation.}
% We evaluate against two methods explicitly designed for large-scale registration: ITK-DReg \citep{itkdreg} (CPU-based) and CLAIRE \citep{claire} (multi-GPU). 
% To broaden the comparison, we adapt several SOTA learning-based approaches commonly used in neuroimaging - SynthMorph \citep{synthmorph}, Vector-Field Attention \citep{vfa}, unigradICON \citep{unigradicon} (with/without instance optimization), TransMorph \citep{transmorph}, anatomix+ConvexAdam \citep{anatomix}.
All methods (including CLAIRE and FireANTs without {\methodname}) run out of memory at $250\um$ resolution.
We proposed two modifications to deep learning based methods to enable them to work on this dataset: (a) inspired by several high-resolution studies \citep{allen,dukeatlas,edlow20197}, we register the images at a downsampled resolution, and then upsample the deformation field (b) inspired by several histology registration methods \citep{deeperhistreg,patchbasedhistoreg,patchbasedhisto3}, we perform patchwise registration and mosaicing of the final deformation.
% 
% Inspired by several high-resolution studies \cite{}, we 
% Since all of these methods run out of memory at the $500\um$ resolution, we augment the methods to perform patchwise
% Methods are applied at three resolutions: 1mm, $500\um$, and $250\um$. 
We compare the methods at three resolutions: 1mm, $500\um$, and $250\um$. 
At 1mm, the full image fits within a patch, providing a baseline reference comparable to reported OASIS performance. 
At higher resolutions, patches are defined by each method's default input size with stride equal to 50\% of the patch size. 
FireANTs augmented with {\methodname} is denoted as \textit{Ours}.  
% We also include FireANTs with fused kernels at full resolution (denoted \textit{Ours}).  
% 
% \paragraph{Baselines.}
% % We compare with ITK-DReg, a CPU-based baseline that 
% We use several baselines for comparison. 
% We compare with two methods aimed at large-scale registration - ITK-DReg, a CPU library for the purpose of registering large-scale images out of memory, and CLAIRE, a GPU library for intramodal image registration on multiple GPUs.
% % 
% To compensate for the lack of deep and classical baselines that work at high-resolution, we take inspiration from the broad spectrum of histology registration methods that perform patchwise registration and mosaicing the final deformation for very large images.
% % 
% Specifically, we take SynthMorph, Vector-Field Attention, unigradICON (with and without instance optimization), TransMorph-large, anatomix+ConvexAdam, FireANTs, and modify the algorithms to register patches of images followed by mosaicing the result. 
% We do this at three scales of the image - 1mm, $500\um$ and $250\um$ isotropic.
% At 1mm isotropic, all methods operate on a single patch, i.e. the image and we get an estimate of the baseline performance and expect it to closely match the validation score on the OASIS dataset. 
% At higher resolutions, we select the patch size as configured by the baseline and select a stride of half the patch size dimensions. 
% We also run FireANTs at native-resolution using our fused kernels - denoted as Ours.
% % 
% % and extend the baselines to perform patchwise registration, followed by mosaicing the final deformation.
We report Dice, inverse-weighted Dice (InvDice; \cite{claire}), and average Haussdorf distance capped at 90 percentile (AvgHD90). 
To compare efficiency, we measure both wall-clock time and GPU-hours.

% \paragraph{Evaluation.} For each baseline, we compare the Dice scores, the inverse-weighted Dice scores \cite{claire} to penalize smaller structures, and HD90. 
% To compare the efficiency of these methods, we also include plots comparing the accuracy with both the Wall clock time, and the GPU-hours consumed.
% \input{figures/fauxoasis-efficiency.tex}

\textbf{Results.}
~\cref{tab:exp1_allmethods_grouped} summarizes performance metrics.
At 1mm, most methods achieve performance consistent with their reported performance on OASIS, including VFA and TransMorph which were trained on the OASIS dataset with label supervision.
At higher resolutions, nearly all methods degrade, especially for InvDice and HD90, which emphasize alignment of fine structures. 
In contrast, our method improves in accuracy: at $250\um$, we improve Dice 
by 18.1 points, InvDice by 31.6 points, and reduce AvgHD90 by 62.1\%. 
The correlation between resolution and performance is also observed in \citep{claire,claire2,highres}; in addition we verify that patch-based methods \textit{degrade} in performance at higher resolutions. %do not recover performance at higher resolutions - they do quite the opposite.

This degradation among patchwise methods is expected; histology-style pipelines typically register consecutive slides with small deformations after affine alignment. 
At high resolution, patching reduces anatomical context and the patches become progressively more out-of-distribution (see \cref{fig:patch_analysis}).
Patchwise or downsampling strategies are therefore insufficient for ultra-high resolution large-scale registration, and existing deep methods cannot be repurposed to work at higher resolutions efficiently.
% 
% Finally, our method is substantially more efficient, requiring up to $500\times$ fewer GPU-hours compared to alternatives, particularly at $250\um$.
Accuracy-efficiency tradeoffs in \cref{fig:acc_gbhours,fig:acc_time} show that our method is Pareto-efficient compared to all other methods (CPU, deep learning, and distributed GPU methods), requiring up to $500\times$ fewer GPU-hours compared to alternatives at $250\um$.

\subsection{Ablation Studies}
% \textbf{Ablation Studies}
We ablate on the efficiency of various workhorse operations used in image registration in \cref{fig:ablations,tab:cc-ablation-table}.
We compare our implementations to community-standard PyTorch implementation \citep{xijiafastlncc,transmorph} and \texttt{torch.compile} versions.
For grid sampler and MI kernels, our kernels have $O(1)$ extra HBM overhead instead of $O(N)$ in the PyTorch implementation.
For LNCC, our implementation achieves an average speedup in the forward pass by $5.22\times$ and $56.98\times$ in the backward pass. 
Our \texttt{grid\_sampler} also leads to an efficient scaling-and-squaring operation, commonly used in deep learning registration pipelines \citep{transmorph}, with a memory reduction of $50\%$ compared to the baseline implementation.
% 
% \subsection{Ablations on correctness of Grid Parallel}
% Eval on faux-OASIS with grid parallel on and off.
% Without grid parallel, the smoothing happens with incorrect boundary conditions leading to lower performance
% 
% \input{figures/qualitative_experiment}
% 
% \vspace{-10pt}
% \subsection{Scalability Analysis}
\textbf{Scalability Analysis.}
% We test the scaling of our proposed distributed framework and the effectiveness of the Ring Sampler 
We test the weak scaling of our distributed framework by registering synthetic images with increasing voxel sizes.
For $H$ GPUs, we instantiate an image pair of size $700\times700\times700H$ and shard the images, warp, and optimizer state across $H$ GPUs.
\cref{fig:ring-sampler-memory-consumption} shows weak scaling of {\methodname} with and without ring sampler.
Without the ring sampler, the \texttt{grid\_sample} operation requires storing the moving image of size $700\times700\times700H$ on each GPU, leading to peak HBM memory increasing linearly with $H$.
This implies the framework would not scale to arbitrarily large problem sizes, regardless of cluster size $H$.
Peak Memory consumption is independent of $H$ with the Ring Sampler, and scaling efficiency is only minimally affected. % showing its efficacy. %compared to without the ring sampler, showing the effectiveness of the ring sampler.
\textbf{Ablation on GP.}
We ablate the effect of GP by replacing it with DTensor sharding (no boundary sync).
\cref{fig:gpfmost-trailer,fig:gpfmost,fig:gpoasis} show that incorrect boundary synchronization leads to undesirable artifacts in the moved images, and reduces labelmap overlap.

% We test the scalability of our distributed framework 
% to verify the efficiency of our framework for arbitrarily large problem sizes.

% The non-GEMM and random access nature of computation in LNCC, MI, grid sampling limit the applicability of various optimizations proposed for large-scale transformer training.

% Figure \cref{fig} shows that our kernels scale at large problem sizes effectively.

\section{Conclusion}
We propose a novel distributed framework for arbitrarily large image registration problems.
Our work identifies and proposes IO-aware and distributed-friendly implementations of workhorse operations in image registration algorithms, enabling registration of images at arbitrarily large resolutions on a single GPU.
Our fused primitives demonstrate compelling results in both improving existing registration pipelines and scaling to arbitrarily large, multimodal problems pertinent in modern life science applications, that were previously infeasible without approximations.
{\methodname} shows unprecedented registration capabilities that will enable researchers to leverage and effectively work with large-scale image volumes and unearth new insights leveraging the large resolution images.

%% file: figures/qualitative_experiment.tex
\begin{figure}[t!]
\centering
\scalebox{0.706}{
\begin{tikzpicture}[x=6cm, y=6cm, spy using outlines={every spy on node/.append style={smallwindow}}]

% \begin{tikzpicture}[spy using outlines={every spy on node/.append style={smallwindow}}]
% Moving image
\node[anchor=south] (FigA) at (-0.75,0) {\includegraphics[height=3.2cm]{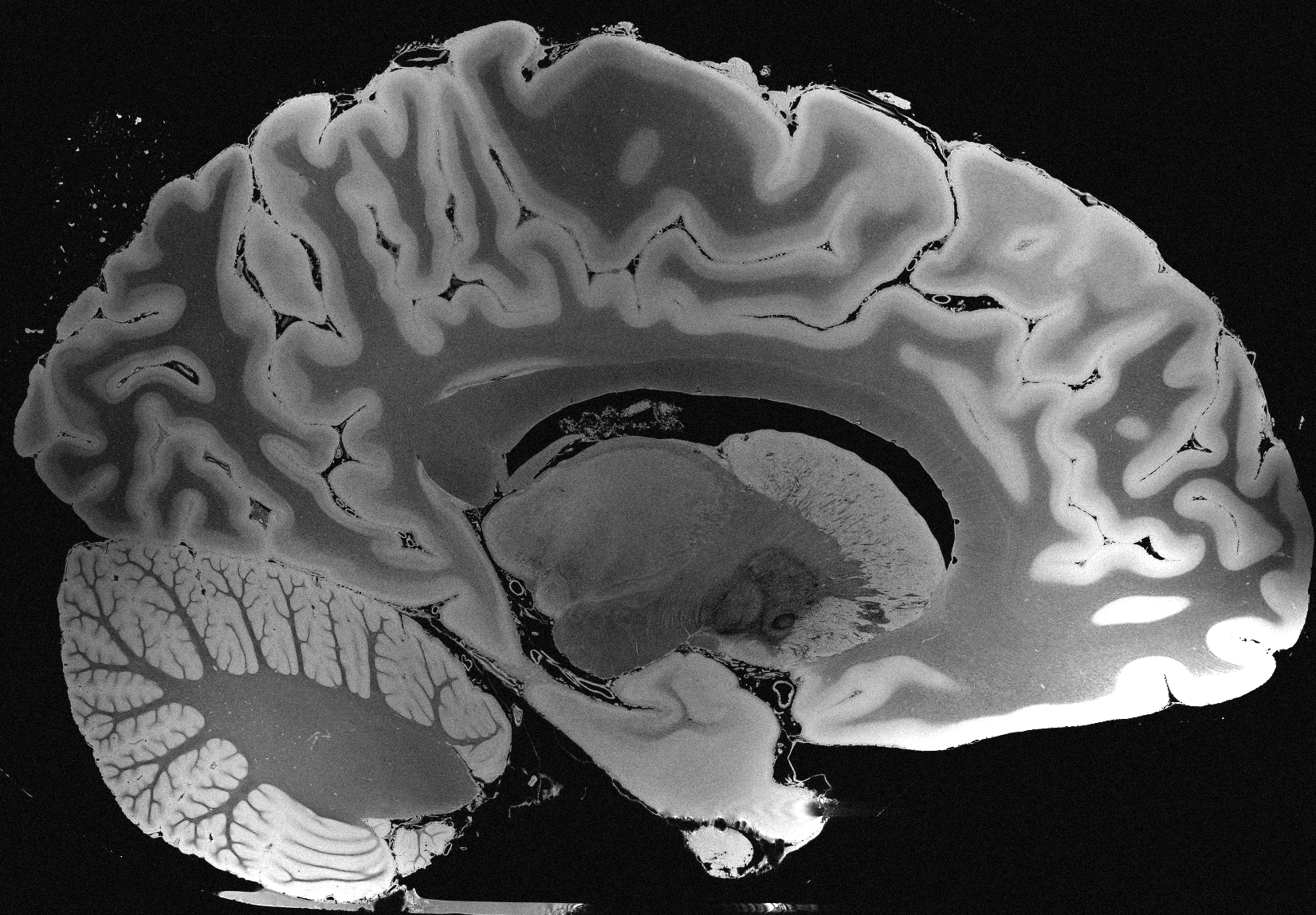}};
\spy [closeup,magnification=2,vibrantred,size=1.5cm] on ($(FigA)+(+0.09,+0.0)$)
    in node[upperwindow,anchor=north west] at ($(FigA.north east) - (0,0.02)$);
\spy [closeup,magnification=1.2,vibrantblue,size=1.5cm] on ($(FigA)+(-0.21,-0.12)$)
    in node[lowerwindow,anchor=south west] at ($(FigA.south east) + (0,0.02)$);
\node [anchor=north] at ($(FigA.south)$) {\sf Fixed (100$\um$ FLASH)};

\node[anchor=south] (FigB) at (0.35,0) {\includegraphics[height=3.2cm]{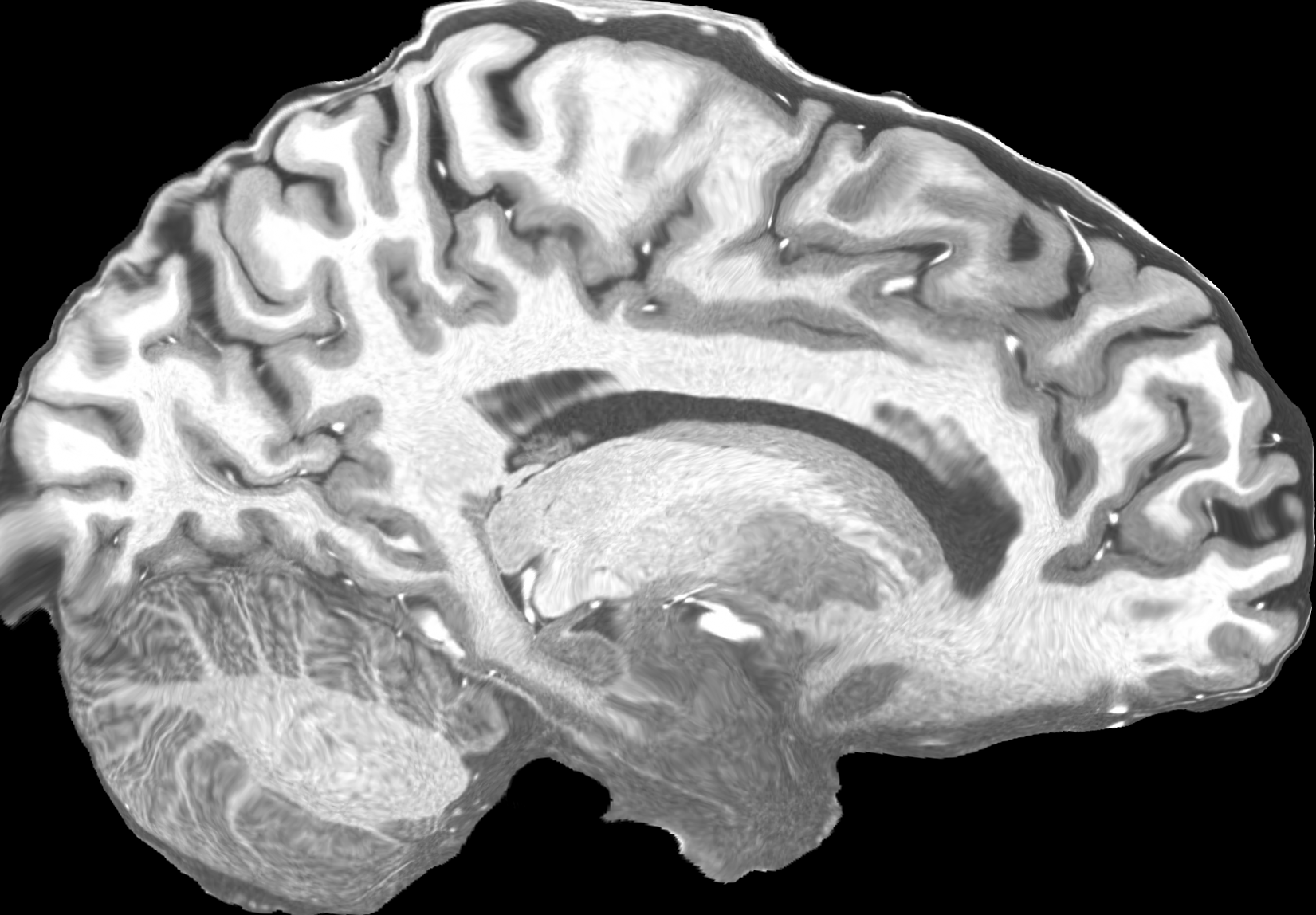}};
\spy [closeup,magnification=2,vibrantred,size=1.5cm] on ($(FigB)+(+0.09,+0.0)$)
    in node[upperwindow,anchor=north west] at ($(FigB.north east) - (0,0.02)$);
\spy [closeup,magnification=1.2,vibrantblue,size=1.5cm] on ($(FigB)+(-0.21,-0.12)$)
    in node[lowerwindow,anchor=south west] at ($(FigB.south east) + (0,0.02)$);
\node [anchor=north] at ($(FigB.south)$) {\sf Moved (250$\um \rightarrow$ 100$\um$)};

\node[anchor=south] (FigC) at (1.45,0) {\includegraphics[height=3.2cm]{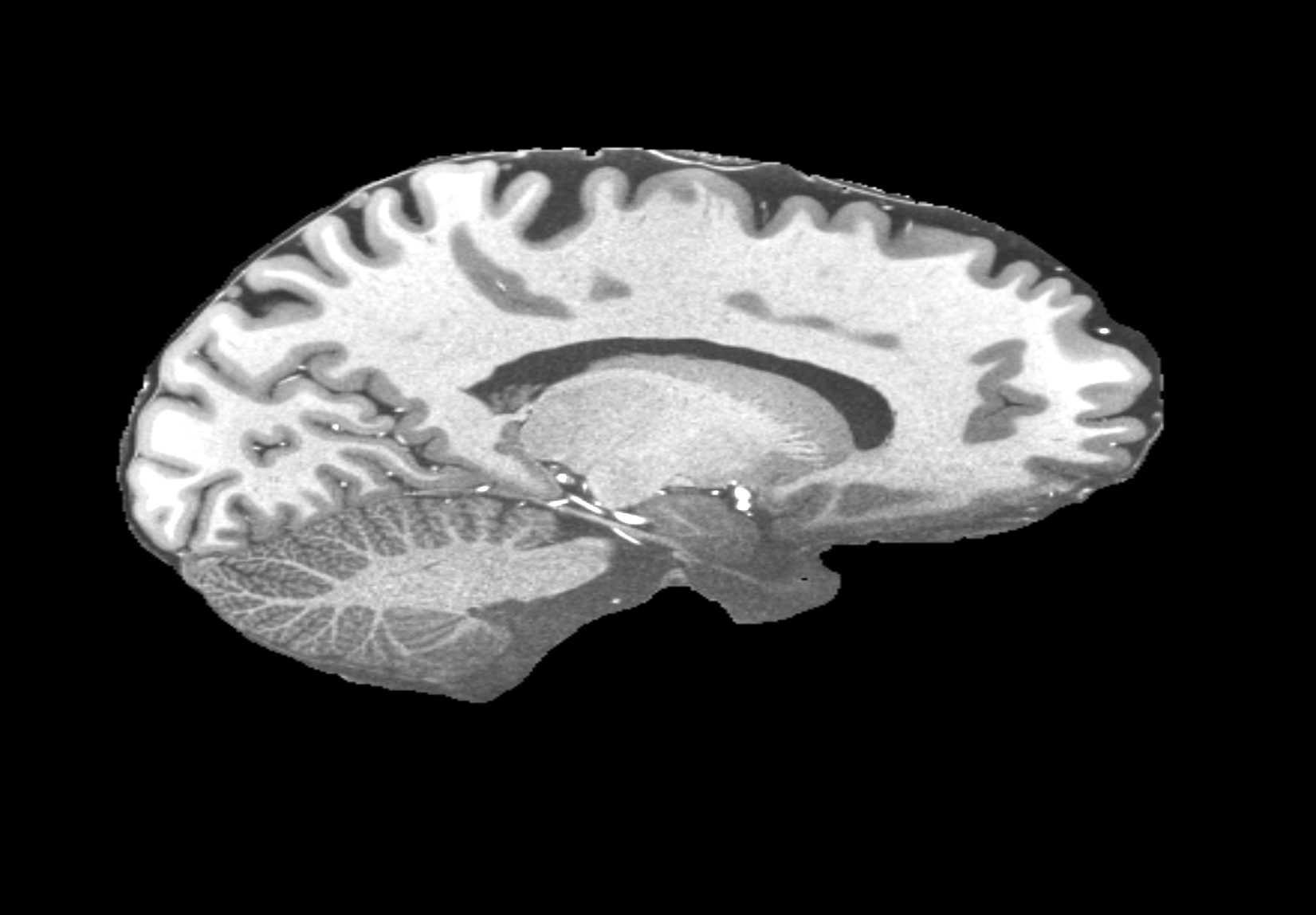}};
\spy [closeup,magnification=2,vibrantred,size=1.5cm] on ($(FigC)+(+0.05,+0.02)$)
    in node[upperwindow,anchor=north west] at ($(FigC.north east) - (0,0.02)$);
\spy [closeup,magnification=1.2,vibrantblue,size=1.5cm] on ($(FigC)+(-0.15,-0.08)$)
    in node[lowerwindow,anchor=south west] at ($(FigC.south east) + (0,0.02)$);
\node [anchor=north] at ($(FigC.south)$) {\sf Moving (250$\um$ T1)};

\end{tikzpicture}
}
\caption{Qualitative comparison on registration of 100$\um$ ex-vivo brain MRI (T1 $\rightarrow$ FLASH) image. 
Fine details like cerebellar white matter are not visible at macroscopic scales, but are aligned at 100$\um$. 
Fixed image is of size $1760\times1760\times1278$. Best viewed zoomed in. More results in \cref{fig:qualtitative_registration_comparison_app}.
}
\label{fig:qualtitative_registration_comparison}
\end{figure}

%% file: tables/transmorph-speedup.tex
\begin{table}[H]
% \begin{wraptable}{R}{0.7\linewidth} 
\centering
\caption{\small Accelerating TransMorph (\textbf{Top}) and FireANTs (\textbf{Bottom}) training with various computation backends.}
\label{tab:transmorph-speedup}
\renewcommand{\arraystretch}{1.1}
\resizebox{0.9\linewidth}{!}{%
\begin{tabular}{llcccc}
\hline
\textbf{Variant} & \textbf{Loss} & \textbf{Diffeomorphic} & \textbf{Training Time (h)} & \textbf{GPU Mem (GB)} & \textbf{Val DSC} \\
\hline
Baseline & LNCC & \no & 171.20 & 20.01& 86.74 \\
Ours & LNCC & \no & {27.84} & {16.95} & 87.23 \\
\hline
Baseline & LNCC & \yes & 171.42 & 21.28 & 86.55 \\
Ours & LNCC & \yes & {27.93} & {17.34} & 87.09 \\
\hline
Baseline & MI & \no & 26.09 & 22.34 & 86.74 \\
Ours & MI & \no & {24.94} & {16.80} & 86.80 \\
\hline
\end{tabular}
}
\vspace{1em}
\resizebox{\linewidth}{!}{
\begin{tabular}{llccr}
\hline
\textbf{Loss} & \textbf{Backend} & \textbf{Dice Score} $\uparrow$ & \textbf{Runtime (s)} $\downarrow$ & \textbf{Memory (MB)} $\downarrow$ \\
\hline
LNCC & FireANTs & \best{78.81} $\pm$ 3.87 & 1.44 $\pm$ 0.08 & 1044.5 $\pm$ 0.0 \\
LNCC & FastLNCC & 76.96 $\pm$ 3.60 & 3.76 $\pm$ 0.16 & 1026.3 $\pm$ 0.0 \\
LNCC & VXM/TM & 76.96 $\pm$ 3.60 & 57.08 $\pm$ 2.45 & 1418.5 $\pm$ 0.0 \\
LNCC & \texttt{torch.compile} & 69.35 $\pm$ 4.09 & \second{0.82 $\pm$ 0.04} & \second{860.7 $\pm$ 0.0} \\
LNCC & Ours & \second{78.67} $\pm$ 3.04 & \best{0.50} $\pm$ 0.01 & \best{577.5 $\pm$ 0.0} \\
\hline
MI & PyTorch & \second{75.88 $\pm$ 3.45} & 7.51 $\pm$ 0.37 & 12206.3 $\pm$ 0.0 \\
MI & \texttt{torch.compile} & \second{75.88 $\pm$ 3.45} & \best{1.05 $\pm$ 0.05} & 3865.5 $\pm$ 0.0 \\
MI & Ours & \second{75.88 $\pm$ 3.44} & \second{2.90 $\pm$ 0.16} & \best{577.5 $\pm$ 0.0} \\
MI & \texttt{torch.compile}+Ours & \best{75.93} $\pm$ 3.47 & 2.95 $\pm$ 0.16 & \second{657.3 $\pm$ 0.0} \\
\hline
\end{tabular}
}
% \end{wraptable}
\end{table}

%% file: figures/scaling.tex
% \begin{figure}[h]
\begin{figure}
    \centering
    \begin{minipage}{\linewidth}
        \includegraphics[width=0.49\linewidth]{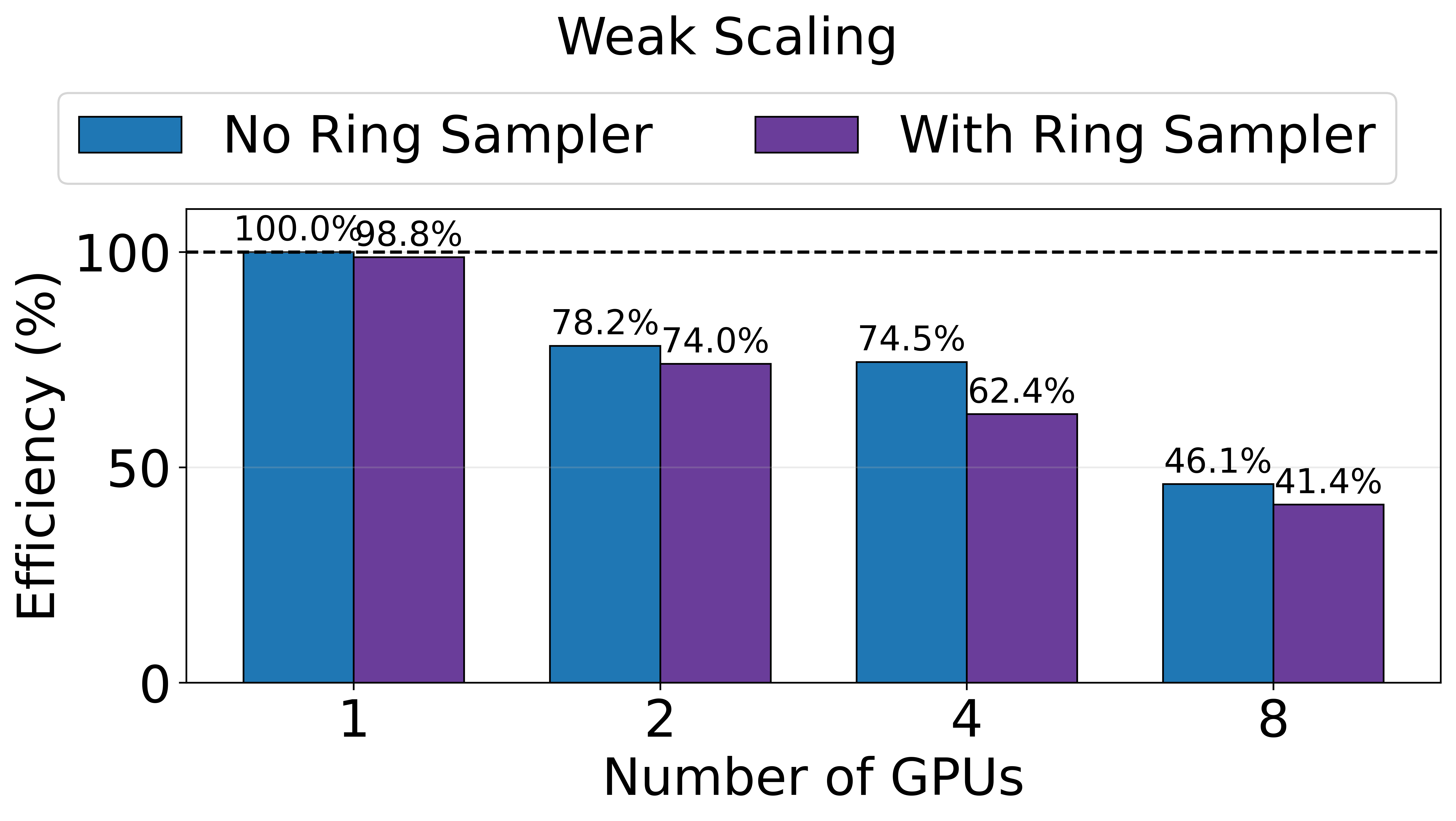}
        \includegraphics[width=0.49\linewidth]{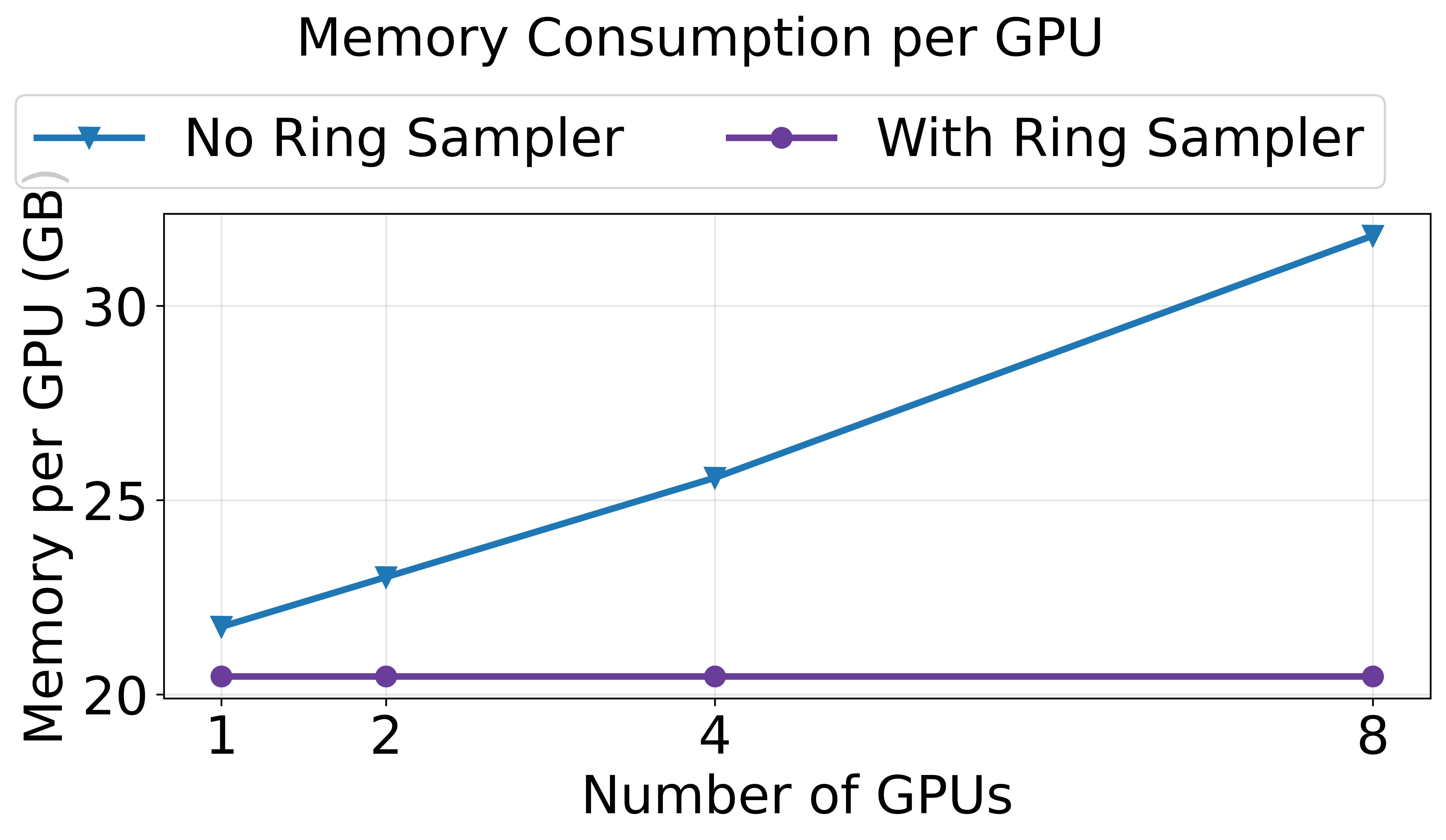}
        \label{fig:ring-sampler-memory-consumption}
    \end{minipage}
    \caption{Weak scaling and Per-GPU memory consumption of {\methodname}.}
    \label{fig:ringsampler-mem-ablation}
\end{figure}

\begin{wrapfigure}{L}{0.37\textwidth} % r = right, l = left
    \centering
    \begin{minipage}{\linewidth}
        \includegraphics[width=\linewidth]{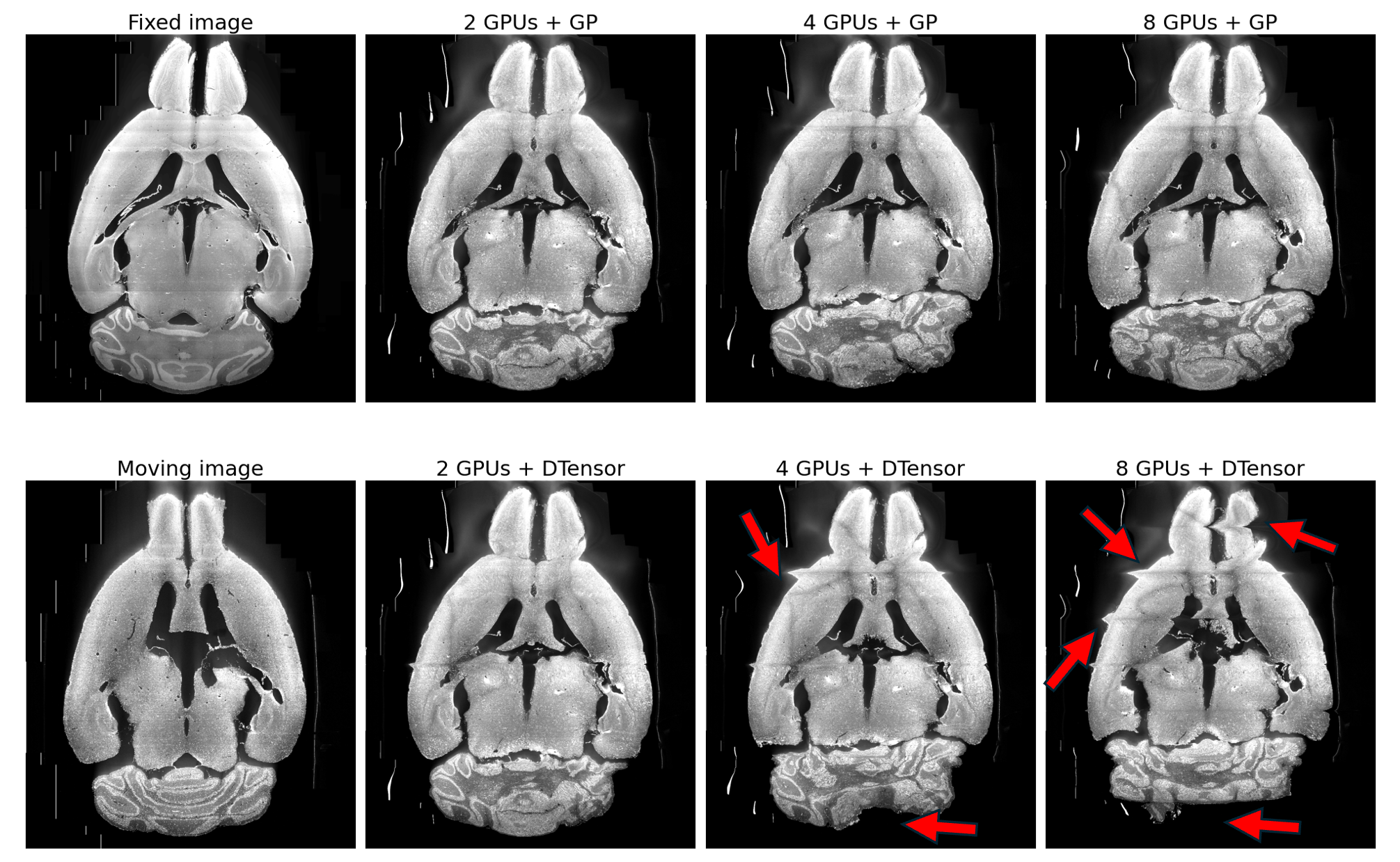}
        \subcaption{Qualitative ablation of GP synchronization in {\methodname} on the fMOST mouse brain dataset \citep{tustison2024antsx}. \textcolor{red}{Red} arrows highlight regions affected by incorrect boundary effects due to no GP. See \cref{fig:gpfmost} for more examples.}
        \label{fig:gpfmost-trailer}
    \end{minipage}
    \caption{Scaling and GP ablations.}
    \label{fig:scaling}
% \end{figure}
\end{wrapfigure}

%% file: figures/loss-ablations.tex
\begin{figure}[t!]
    \centering
    \begin{minipage}{0.49\linewidth}
        \includegraphics[width=\linewidth]{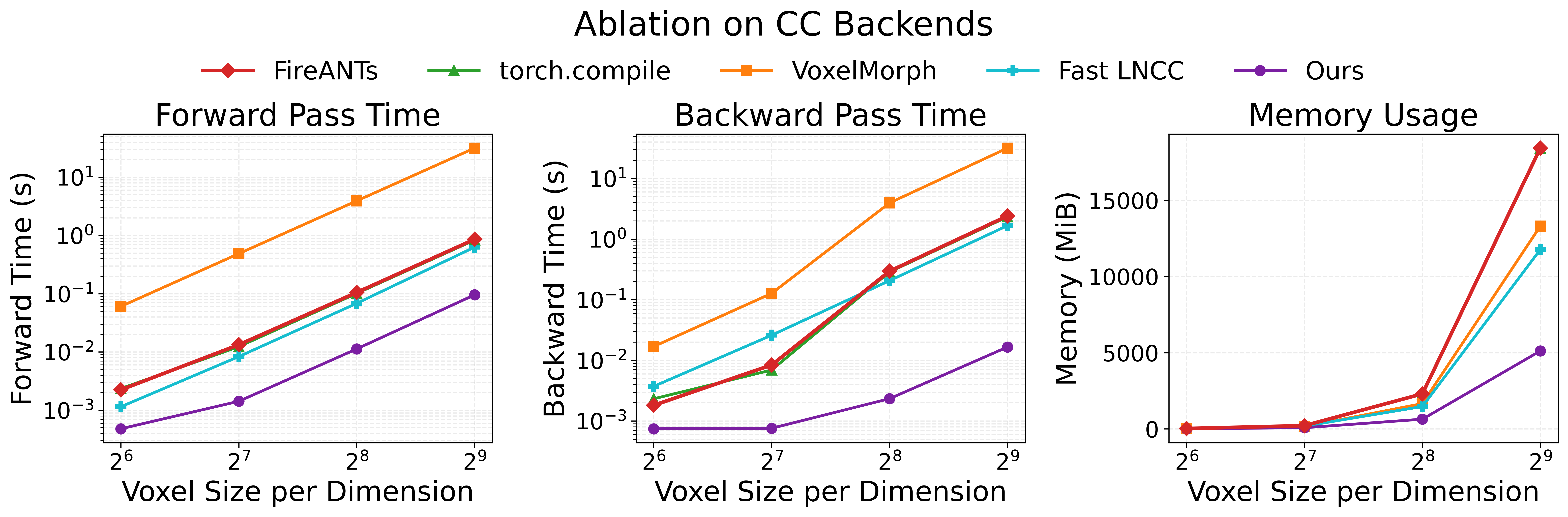} 
    \end{minipage}
    \hfill 
    \begin{minipage}{0.49\linewidth}
        \includegraphics[width=\linewidth]{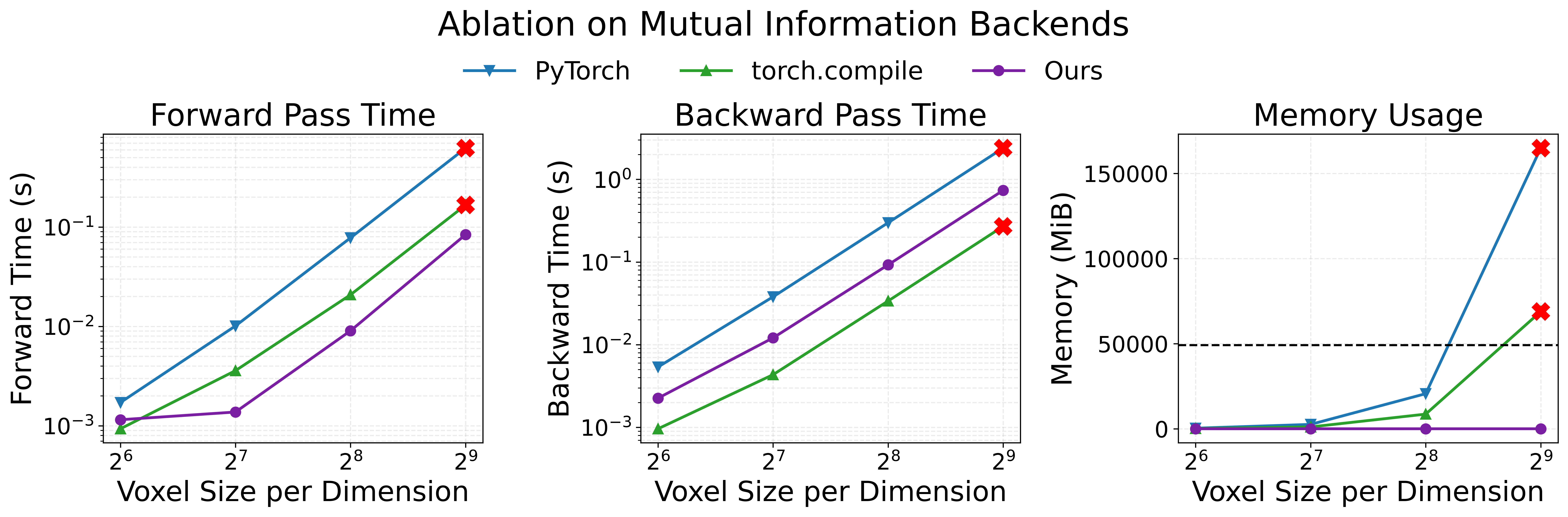} 
    \end{minipage}
    \begin{minipage}{0.49\linewidth}
        \includegraphics[width=\linewidth]{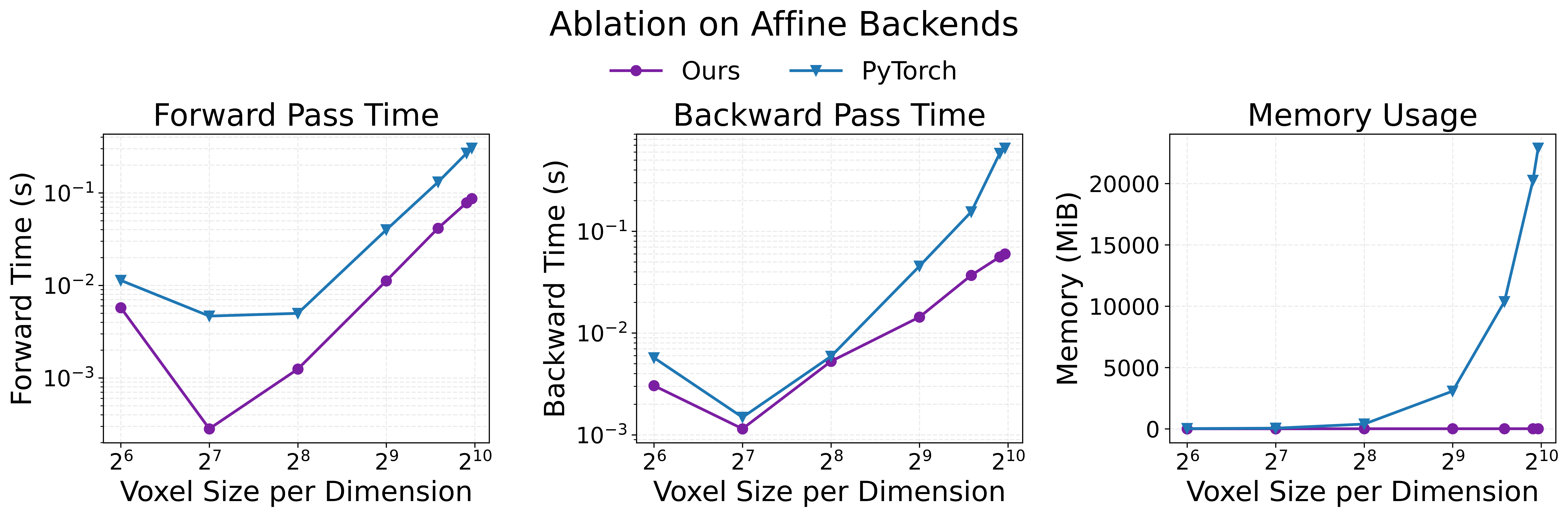} 
    \end{minipage}
    \hfill 
    \begin{minipage}{0.49\linewidth}
        \includegraphics[width=\linewidth]{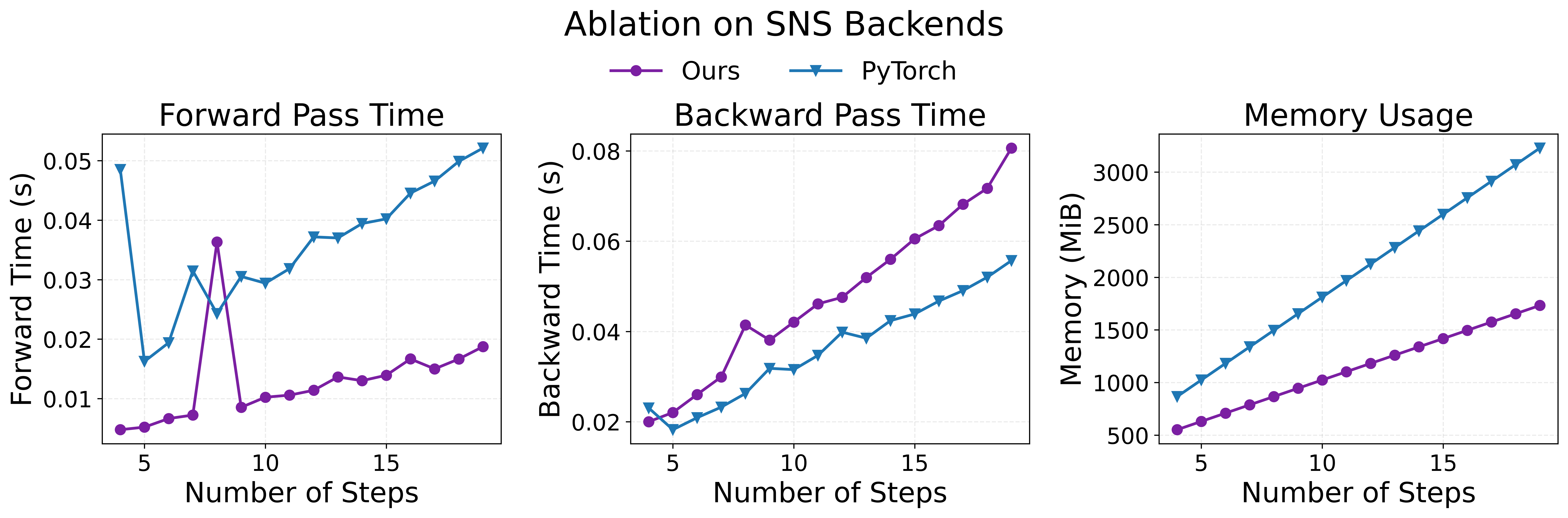} 
    \end{minipage}
    \caption{Ablations on key workhorse operations: LNCC, MI, \texttt{grid\_sampler}, and {scaling-and-squaring} operations. Our fused kernels consume significantly less HBM and runtime.}
    \label{fig:ablations}
\end{figure}

%% file: sections/appendix.tex
%%%%%%%%%%%%%%%%%%%%%%%%%%%%%%%%%%%%%%%%%%%%%%%%%%%%%%%%%%%%
\input{sections/relatedworks}

\section{Limitations and Future Work}
One of the limitations of the proposed framework is the relatively poor weak scaling of the method in
the distributed setting (41\% on 8 GPUs without NVLink or Infiniband). 
Even so, for most life science applications feasibility is the first step towards scalable, distributed, multimodal registration, and future work will focus on improving the weak scaling of the method. 
Another active avenue for future work is to enable Virtual GridParallel (VGP) to use fewer GPUs by sequentially offloading and onloading consecutive shards from CPU onto a single GPU. 
Deformable Registration of the 100 $\um$ volume in \cref{sec:edlow-250um} took only \textit{one minute} on 8 A6000 GPUs, but equivalently it would take around 15-20 minutes to register this pair on a single A6000 GPU with VGP, accounting for repeated CPU offloading. 
This is an acceptable timeframe for large-scale studies, allowing researchers to prototype and iterate on large-scale image volumes as well with a single GPU. 
Other avenues for future work include collecting labeled data at high-resolution for various real-world life science applications and performing comparative studies on these datasets.

\section{LLM Usage}
We use an LLM (minimally) to polish the manuscript and improve clarity of ideas and organization.
All LLM-generated text is thoroughly reviewed, proofread, and revised by the first author of the paper.

% results
\input{sections/app/gridparallel}
\input{figures/qual-images}

\input{sections/app/transmorph-speedup}
\input{sections/app/qualitative-improvements}
%  methods
\input{sections/app/fusedcc}
\input{sections/app/fusedmi}
\input{sections/app/gridsampler}
\input{sections/app/ringsampler}

\section{Correctness of Implementation}
All code is checked for numerical correctness by comparing the results with PyTorch implementations using unit and integration tests.
Code and generated data will be made available to the community.

\subsection{Ablation on FireANTs speedup}
We run FireANTs with different backends for LNCC and MI loss functions on the OASIS validation set \citep{oasis,hering2022learn2reg}.
We measure the end-to-end runtime, peak memory usage (except the fixed and moving images), and Dice score.
We ablate on both Greedy and SyN algorithms; in the case of SyN, additional gradients may be required.
Results in \cref{tab:fireants-speedup} show that our implementation achieves a significant speedup over the baseline implementations.
Although the \texttt{torch.compile} version of LNCC is faster than other variants, it leads to brittle performance.

\input{tables/fireants-speedup}
\input{tables/faux-oasis-efficiency}

% results
\input{sections/app/faux-oasis}

%%%%%%%%%%%%%%%%%%%%%%%%%%%%%%%%%%%%%%%%%%%%%%%%%%%%%%%%%%%

\input{figures/synthetic-patches}
\input{figures/flamegraph}

%% file: sections/relatedworks.tex
\section{Related Works}
\label{sec:related}

\subsection{Memory Efficient and Large Scale Optimization}
Recent advances in large scale transformer-based model training has amassed significant attention and efforts to alleviate key bottlenecks in both memory and compute efficiency.
Activation memory forms a key bottleneck in many deep learning training pipelines, and recent advances propose fused operations \citep{flashattn,flashattn2,flashattn3,pytorch:fusing_conv_bn,casestudyflashattn,flexattention} to significantly reduce HBM usage without approximations.
Other techniques propose sub-quadratic approximations to the quadratic complexity of the attention operation and propose highly efficient and IO-aware fused kernels \citep{nsa,flexattention,flashmask}.
However, as these models and their inputs get increasingly larger in size, they do not fit on a single GPU.
Various distributed techniques like Tensor Parallel \citep{megatronlm}, Sequence Parallel \citep{sequenceparallel,sequenceparallel2,distflashattn}, pipeline parallel \citep{zerobubblepipeline,breadthfirstpipeline,deepseekv3}, fully-sharded data parallel (FSDP2) \citep{torch2,fsdp2scaling,zero} have been proposed that distribute (shard) the model and its inputs across multiple GPUs for transformer-like models. 
Another research area approaches the problem of scaling large models by building compilers and intermediate representations to enable writing optimized kernels at runtime \cite{triton,torch2,thunderkittens,tvm,tensorflow}.
To our knowledge, most of these techniques are tailored to transformer-specific architectures and GEMM-like operations (self attention, feedforward, batchnorm, etc.) only, and a Tensor/Model Parallel variant for convolution-aware sharding is not available.
However, other disciples including biomedical and clinical imaging, life sciences, climate modeling, drug discovery, genomics, geosciences, robotics leverage other key components that do not fit in the transformer-specific framework, or are GEMM-like in nature.
We focus on the image registration problem, that is a key component in a variety of biomedical and life science applications.
% We optimize the key bottlenecks in a fundamental algorithm - deformable image registration, that is used ubiquitously in a variety of disciplines.

\subsection{Modern applications in life sciences and biomedical imaging}
\label{subsec:modern_apps_bioimaging}

Over the past decade, imaging across the life sciences and biomedical domains has progressed from mesoscale surveys to organ- and organism-wide acquisitions at cellular or even subcellular resolution. 
These span transparent organisms and small animal models (e.g., \textit{C.\ elegans}, zebrafish, adult \textit{Drosophila}) \citep{varol2020statistical,venkatachalam2016pan,zebrafishreg,zebrafishreg2,peng2011brainaligner,brezovec2024mapping}, whole-rodent brains imaged at micron or submicron sampling \citep{fmostimaging,wang_allen_2020}, and non-human primate (NHP) and human ex vivo MRI at hundreds of microns \citep{marmoset,nhpopenresource,edlow20197,lusebrink2017t1}. 
Such modalities routinely generate giga- to teravoxel volumes \citep{clarity2016,highres}. 
Their scientific utility, however, hinges 
% on what we term \emph{registration performed at the native resolution of acquisition}: 
on the ability to perform registration at the native resolution of acquisition, i.e. 
aligning specimens (or modalities) in a common coordinate system without sacrificing the fine-scale morphologies-cell bodies, layers, axon bundles, synaptic neighborhoods--
that motivate high-resolution acquisition in the first place \citep{highres,goubran2013image}.

\paragraph{Cellular-resolution atlases in model organisms.}
% In small transparent organisms, atlases demand cellular precision. 
In \textit{C.\ elegans}, statistical atlases of neuron positions require aligning whole-animal volumes to preserve the fidelity of closely apposed cells \citep{varol2020statistical,venkatachalam2016pan}. 
In zebrafish, deformable registration with cellular-level precision and minimal perturbation of tissue morphology enables pooling of gene expression, single-neuron morphologies, and brain-wide activity \citep{zebrafishreg,zebrafishreg2}. 
In adult \textit{Drosophila}, whole-brain registration underpins large-scale databases and enables structure--function integration (for example, aligning two-photon functional volumes to EM-derived connectomes) \citep{peng2011brainaligner,brezovec2024mapping}.

\paragraph{Whole-brain rodent imaging and fMOST-scale data.}
In rodents, fMOST pipelines yield whole-brain images at micron sampling (e.g., 0.32$\um$ voxels generating $>$10TB datasets) for tracing long-range axons and quantifying cytoarchitecture \citep{fmostimaging}. 
Constructing stereotaxic spaces such as the Allen CCFv3 and Waxholm rat atlas requires deformable registration that preserves layers and boundaries \citep{wang_allen_2020,waxholm,devccf}. 
Registration must also be deformation-tolerant and contrast-aware to handle tissue-clearing distortions \citep{clarity2016}.
Tools like ANTs have been successfully applied to highly downsampled versions of these images to generate the atlases, but atlas reconstruction at higher resolutions remains challenging due to computational bottlenecks.
Empirical benchmarks show that conventional tools like ANTs, developed for MRI-scale inputs, fail or require impractical runtimes at single-cell whole-brain resolution (multi-GB to TB per sample) \citep{highres}. 
Deep learning methods are fast at clinical scales, but run out of memory at higher resolutions.

\paragraph{Non-human primates (NHP), human ex vivo MRI, and biomedical imaging.}  
At larger scales, NHP and human ex vivo MRI achieve hundreds of microns resolution \citep{nhpopenresource,marmoset,edlow20197,lusebrink2017t1}. 
Registration is essential for fusing these volumes with histology or in vivo MRI, enabling alignment of cytoarchitectural detail and correction of distortions \citep{goubran2013image}. 
Without such alignment to a stereotaxic space, the cellular and laminar motifs motivating ultra-high-resolution imaging cannot be meaningfully compared or aligned across specimens or modalities.

In biomedical contexts more broadly, native-resolution registration plays a similarly critical role. 
For example, ex vivo MRI combined with serial histology has been used to build 3D probabilistic atlases of the medial temporal lobe and map tau pathology across layers \citep{Ravikumar2021,Zhu2025}. 
Meanwhile, in digital pathology, registration of gigapixel whole-slide images is challenged by deformations, staining variability, and massive data scale, yet remains a prerequisite for integrating multi-stain, serial-section, and spatially resolved analyses \citep{Gatenbee2023}.  

% Across these diverse domains, the unifying requirement demands fundamentally new approaches to \todo{distributed, scalable, morphology-preserving registration} - a challenge we address in this work.
Across these diverse domains, the unifying requirement demands access to scalable multimodal registration algorithms - a challenge we address in this work.

\subsection{Deformable Image Registration}
Given two images $F: \Omega \rightarrow \reals^d$ and $M: \Omega \rightarrow \reals^d$ defined on domain $\Omega$ (usually a compact subset of $\reals^d$), Deformable Image Registration (DIR) refers to an inverse problem to  find a transformation $\varphi: \Omega \rightarrow \Omega$ that warps the moving image $M$ to the fixed image $F$.
Prior to deep learning, the inverse problem was solved using iterative solvers \citep{klein2009elastix,tustison2013explicit,andersson2007non,ashburner2012spm,avants_lagrangian_2006}, and has been made significantly more scalable by recent advances in GPU-based libraries \citep{claire,claire2,fireants}.
Meanwhile, earliest deep learning for image registation (DLIR) methods \citep{cao2017deformable,krebs2017robust,rohe2017svf,sokooti2017nonrigid} used supervised learning to predict a transformation field using pseudo ground truth transformations.
However, since the inverse problem is generally ill-posed, unsupervised and weakly supervised learning methods \citep{balakrishnan_voxelmorph_2019,zhao2019unsupervised,Zhao_2019_ICCV,joshi_diffeomorphic_nodate,de2019deep,mok_large_2020,zhang2021cascaded,qiu2021learning,lebrat_corticalflow_2021,lku,mok2022affine} became dominant.
However, these methods perform virtually identically to iterative solvers in the unsupervised setting, and show relatively brittle performance under domain shift\citep{magicormirage,jian2024mamba,dio}.
Other recent work has shown increased reliability under domain shift \citep{vfa,transmorph}.
Another line of work combines neural priors with iterative solvers to leverage learnable features with strong convergence properties and robustness of solvers \citep{wu2024neural,pirate,nodeo,Zhao_2019_ICCV,zhao2019unsupervised}.
However, almost all deep learning-based methods typically work reliably only at a macroscopic resolution, with most methods working only at a standard resolution of 1mm or $192\times 160\times 224$ voxels, and running out of memory on larger images, even on other macroscopic problems like lung or full body CT unless they are significantly downsampled \citep{lung250m}.
This is a significant limitation given modern real life applications including the ultra-high-resolution image acquisition techniques used for \textit{ex-vivo} neuroanatomical and developmental biology studies, spanning subcellular structures and connectomes in species like C.elegans \citep{varol2020statistical,venkatachalam2016pan}, zebrafish \citep{zebrafishreg,zebrafishreg2}, adult Drosophila \citep{drosophilatemplate,brezovec2024mapping,peng2011brainaligner}, rodents \citep{allen,dukeatlas,waxholm,devccf}, and non-human primates \citep{marmoset,apereg,nhpalz}.
The scale of these problems is often two to three orders of magnitude larger than the scale of existing deep learning methods.
A simple extrapolation shows that existing deep learning methods will require $\approx 1.87$ TB of GPU memory to train a model on a 250$\um$ ex-vivo brain dataset, making them impractical for training on larger problems.
\citep{claire,claire2} propose a distributed framework for registering arbitrarily large images, but is limited to MSE loss function and a one-parameter subgroup of diffeomorphic transforms (stationary velocity field), which is less flexible than the entire space of diffeomorphic transforms \citep{claire,fireants}.
Moreover, they show results on upto 256 GPUs which indicates room for improvement in terms of scaling efficiency.
In our work, we propose a distributed framework that is upto an order of magnitude more efficient than \citep{claire} on large problems.

%% file: sections/app/gridparallel.tex
\section{Correctness of Grid Parallel Implementation}
\label{app:gridparallel}

\begin{wrapfigure}{R}{0.5\linewidth}
    \centering
    \includegraphics[width=\linewidth]{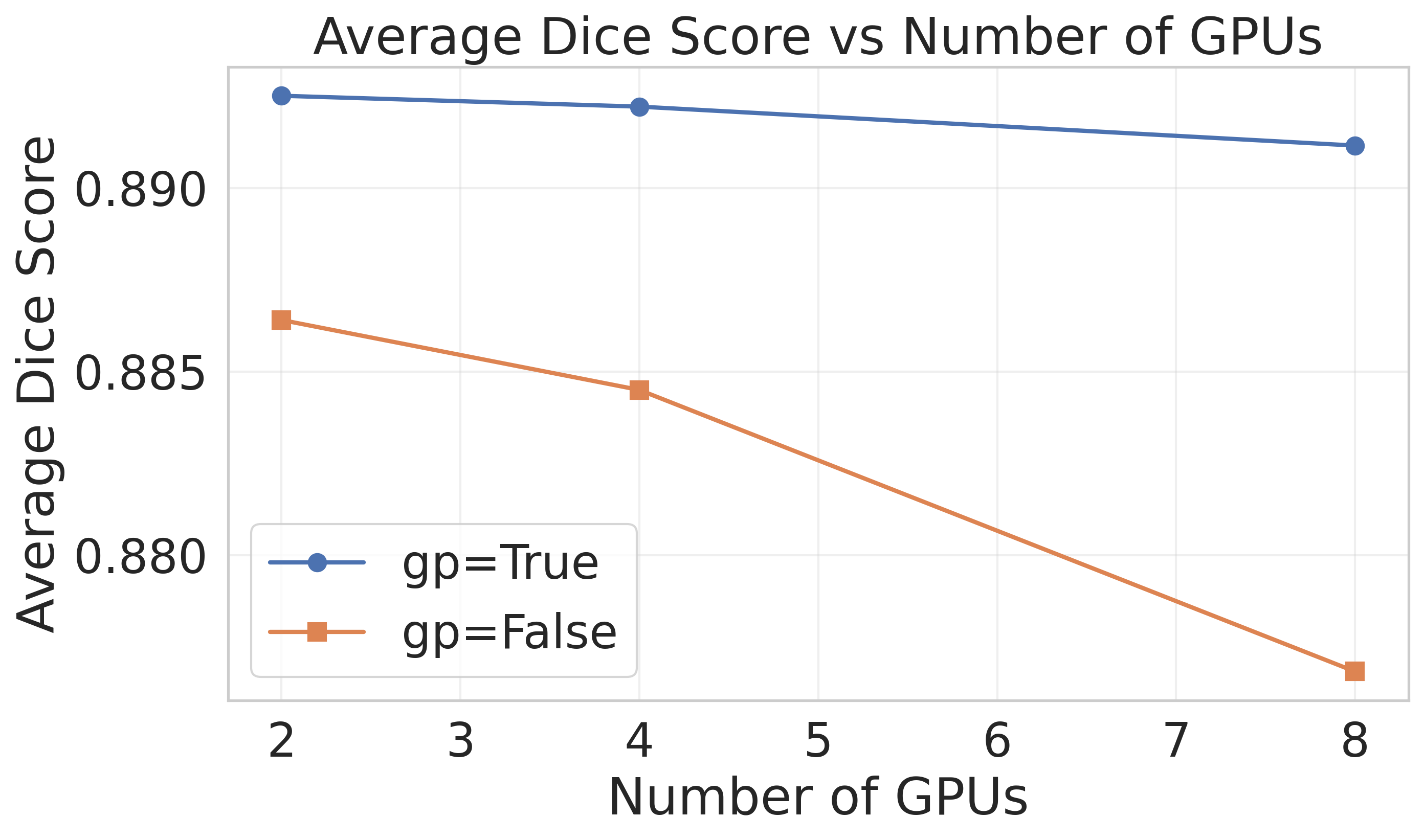}
    \caption{Quantitative ablation of GP synchronization on the faux-OASIS dataset.}
    \label{fig:gpoasis}
\end{wrapfigure}

The GridParallel framework aims to add additional synchronization primitives for performing mathematically correct convolutions across image or grid shards.
These convolutions are required to compute the LNCC loss, 
%smoothing regularization of the warp field, and the Sobolev preconditioning of the warp field.
and applying Sobolev preconditioning of the warp field and its gradient.
Without GP synchronization, the implementation is equivalent to a DTensor sharding \citep{torch2}. 
To our knowledge, existing Model Parallel and FSDP techniques are exclusively built for model weights and activations for linear and self.-attention layers and do not support this functionality.
The pseudocode for convolution with GP synchronization is provided in \cref{alg:gridparallel-conv}.

To ablate the effect of GridParallel synchronization, we register images at 500$\um$ resolution from the faux-OASIS dataset with and without the GridParallel synchronization to measure the effect on performance.
Results in \cref{fig:gpoasis} show only a minimal drop in performance with DTensor sharding.
We posit that this is because the faux OASIS dataset does not contain real-world noise and other artifacts that can degrade performance with incorrect boundary synchronization.
To study the effect of GP synchronization on a more challenging dataset, we register images at 10$\um$ resolution from the fluorescence micro-optical sectioning tomography (fMOST) mouse brain dataset \citep{tustison2024antsx} with and without the GridParallel synchronization to measure the effect on performance.
This dataset contains image volumes of size $1202 \times 1078 \times 627$ voxels, or a displacement field of 9.74GB.
The data contains a myriad of complex artifacts, namely stripe artifacts, boundary halo effects, and speckle noise from image stitching and reconstruction.
We run FireANTs with multi-scale optimization at scales 16, 8, 4, 2, 1$\times$ downsampling for 200, 200, 200, 100, 50 iterations. 
We use our Fused LNCC implementation with a window size of $7$, and a learning rate of 0.5.
Smoothing regularizations are set to $\sigma_{\text{grad}} = 1.0$ and $\sigma_{\text{warp}} = 0.5$.
We ablate on 2, 4 and 8 GPUs.

\begin{figure}[h!]
    \centering
    \includegraphics[width=\linewidth]{images/mousebrain-qual1-v2.png}
    \includegraphics[width=\linewidth]{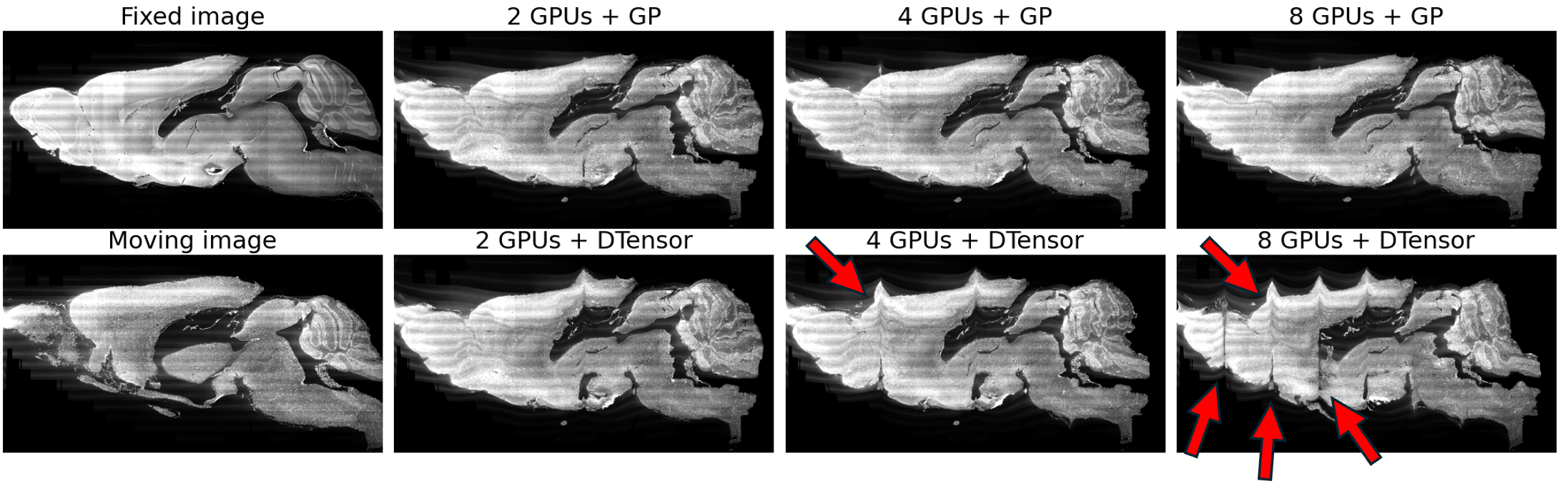}
    \includegraphics[width=\linewidth]{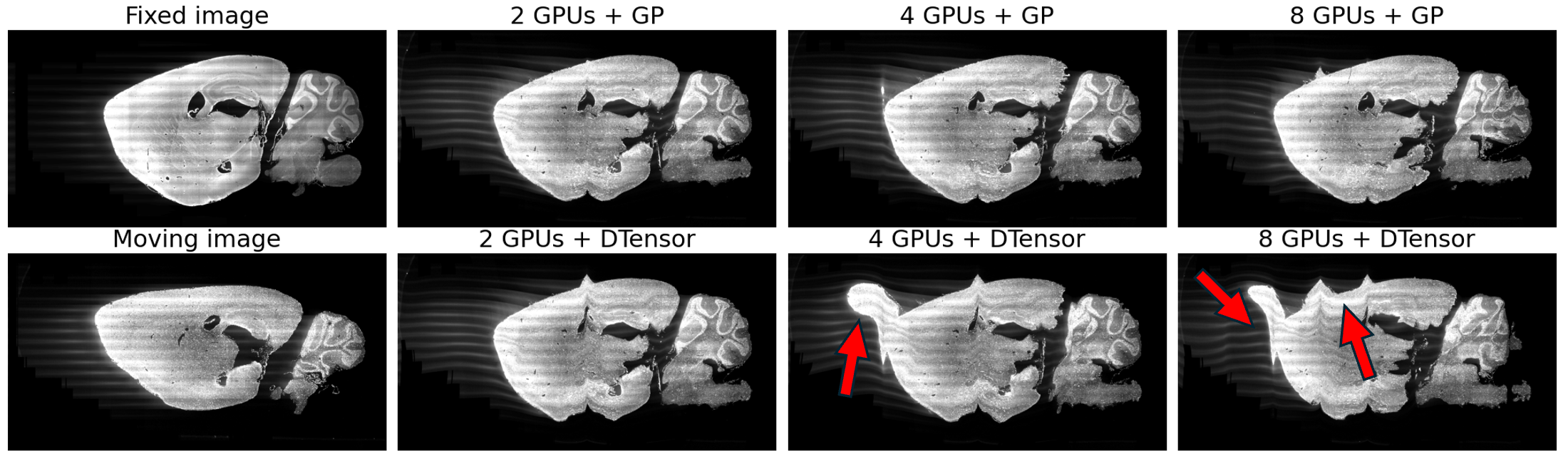}
    \caption{Qualitative ablation of GP synchronization in {\methodname} on the fMOST mouse brain dataset. \textcolor{red}{Red} arrows highlight regions affected by incorrect boundary effects due to no synchronization.}
    \label{fig:gpfmost}
\end{figure}

% \cref{fig:gpoasis} shows that the performance without GP synchronization is only minimally affected as a function of GPUs.
% We posit that this is because the dataset is `easy' enough for `approximate' loss functions and regularizations to work as well.
% To test this hypothesis, we run {\methodname} with and without GP synchronization on a high-resolution fMOST mouse brain dataset at $10\um$ resolution.
Since we do not have ground truth annotations for this dataset, we only make qualitative observations.
Unlike the faux-OASIS dataset, the fMOST dataset is more challenging with high levels of image heterogeneity and complex anatomical structures.
\cref{fig:gpfmost} shows that the performance without GP synchronization is significantly affected as a function of GPUs.
Specifically, the boundaries introduce undesirable artifacts due to mathematically incorrect smoothing and LNCC losses computed across shard.
GP synchronization produces qualitatively better results regardless of the number of GPUs used to shard the problem.

\begin{algorithm}
\caption{Convolution with GP synchronization}
\label{alg:gridparallel-conv}
\begin{algorithmic}[1]
\Require $T$ (tensor), $r$ (rank), $k$ kernel size, $W$ kernel filter, sharding index $sh$, GP size $gp\_size$
\State $\text{pad} \leftarrow (k-1)/2$
\State $\text{bl} \leftarrow \texttt{None}$
\State $\text{br} \leftarrow \texttt{None}$
\If{$r > 0$}
    \State $\text{bl} \leftarrow \texttt{get\_boundary}(r-1, \text{pad})$
\EndIf
\If{$r < \text{gp\_size}$}
    \State $\text{br} \leftarrow \texttt{get\_boundary}(r+1, \text{pad})$
\EndIf
\State $T_{\text{pad}} \leftarrow \texttt{concat}([\text{bl}, T, \text{br}], \text{dim}=\text{sh})$
\State $\text{out} \leftarrow \texttt{conv}(T_{\text{pad}}, W)$
\State $\text{crop\_from\_left} \leftarrow 0$
\State $\text{crop\_from\_right} \leftarrow 0$
\If{$r > 0$}
    \State $\text{crop\_from\_left} \leftarrow \text{pad}$
\EndIf
\If{$r < \text{gp\_size}$}
    \State $\text{crop\_from\_right} \leftarrow \text{pad}$
\EndIf
\State $\text{out} \leftarrow \texttt{crop}(\text{out}, (\text{crop\_from\_left}, \text{crop\_from\_right}), \text{dim}=\text{sh})$
\State \Return $\text{out}$
\end{algorithmic}
\end{algorithm}

%% file: figures/qual-images.tex
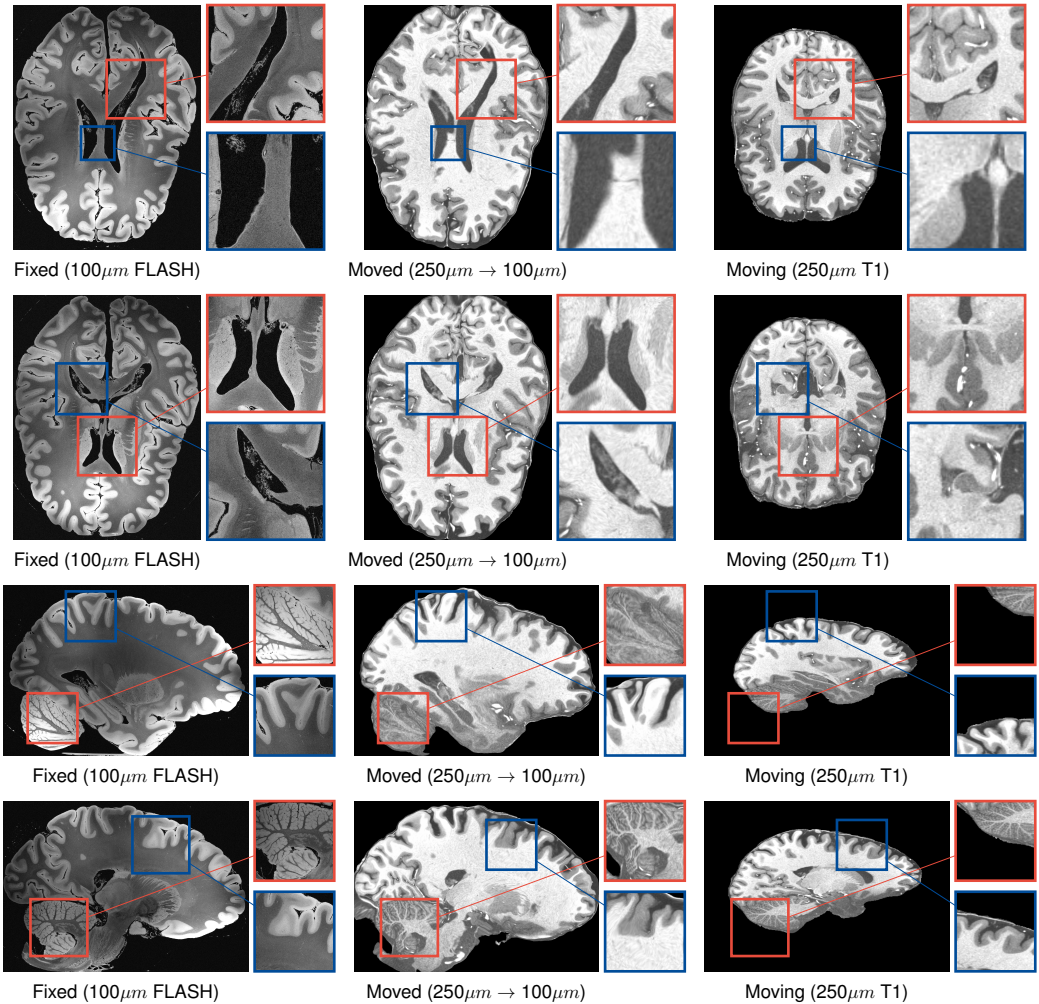
\begin{figure}[t!]
    \centering
    \scalebox{0.706}{

    %%% Picture 1
    \begin{tikzpicture}[x=6cm, y=6cm, spy using outlines={every spy on node/.append style={smallwindow}}]
    % \begin{tikzpicture}[spy using outlines={every spy on node/.append style={smallwindow}}]
    % Moving image
    \newcommand{\figheight}{4.6cm}
    \node[anchor=south] (FigA) at (-0.75,0) {\includegraphics[height=\figheight]{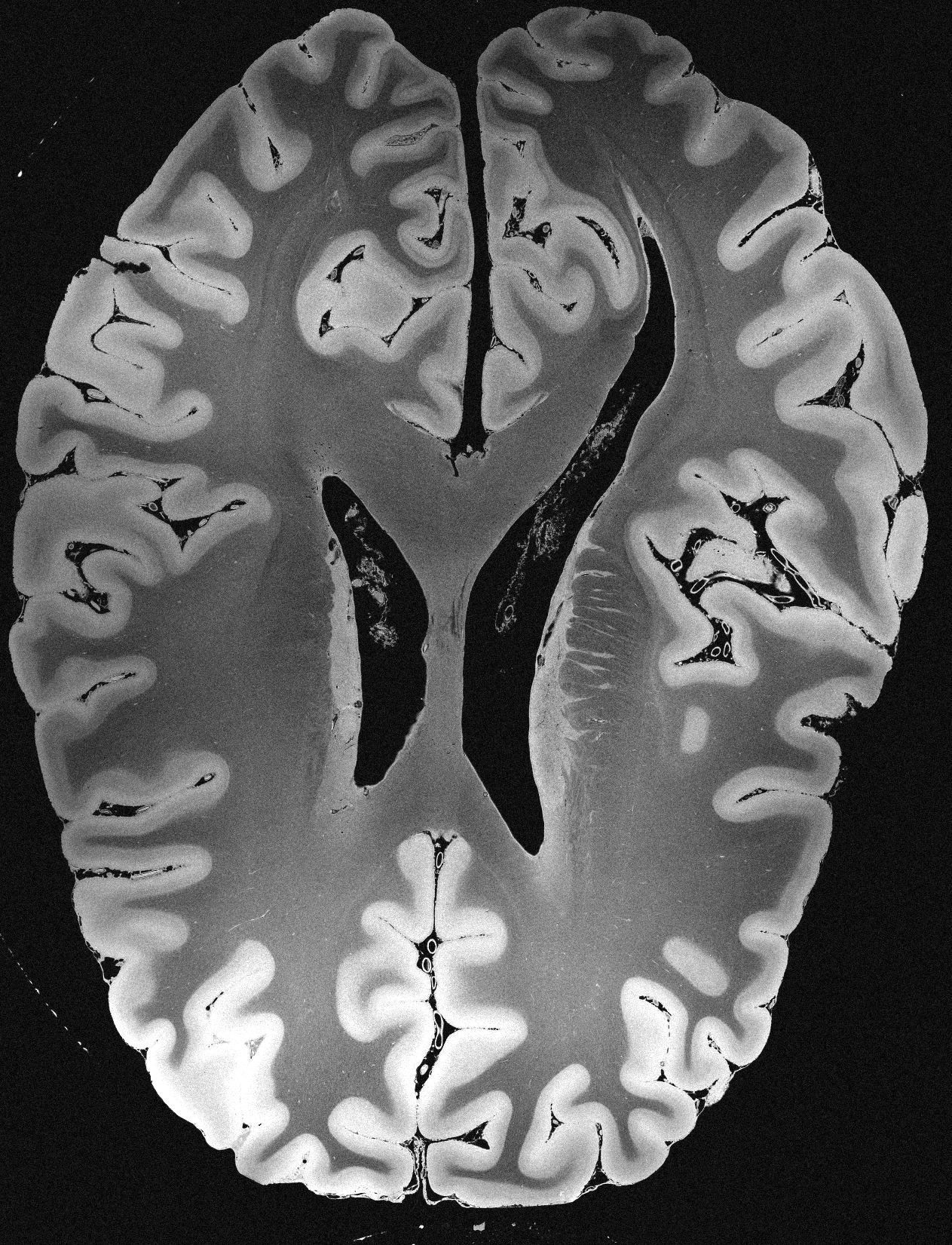}};
    \spy [closeup,magnification=2,vibrantred,size=2.2cm] on ($(FigA)+(+0.09,+0.12)$)
        in node[upperwindow,anchor=north west] at ($(FigA.north east) - (0,0.02)$);
    \spy [closeup,magnification=3.5,vibrantblue,size=2.2cm] on ($(FigA)+(-0.03,-0.05)$)
        in node[lowerwindow,anchor=south west] at ($(FigA.south east) + (0,0.02)$);
    \node [anchor=north] at ($(FigA.south)$) {\sf Fixed (100$\um$ FLASH)};
    
    \node[anchor=south] (FigB) at (0.35,0) {\includegraphics[height=\figheight]{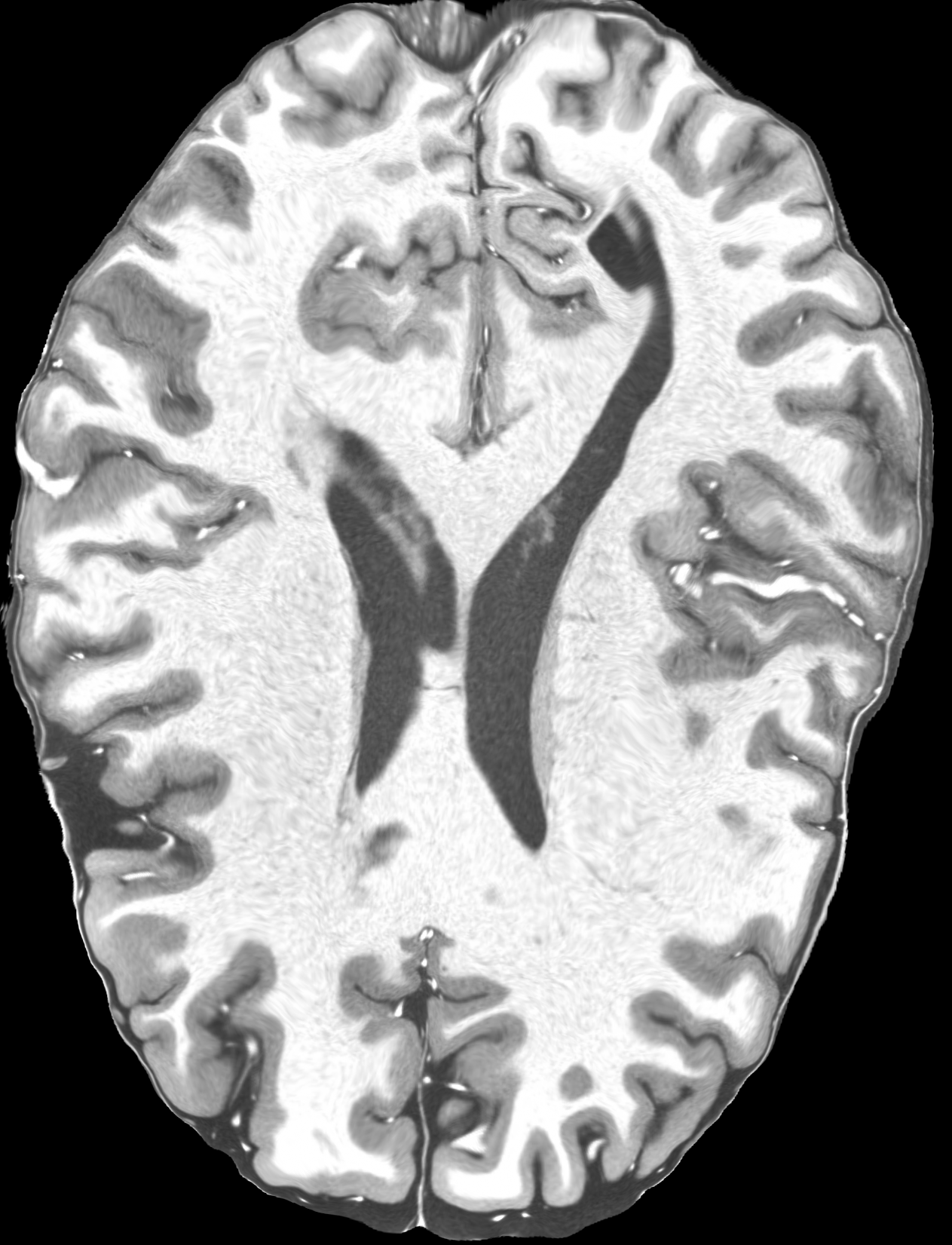}};
    \spy [closeup,magnification=2,vibrantred,size=2.2cm] on ($(FigB)+(+0.09,+0.12)$)
        in node[upperwindow,anchor=north west] at ($(FigB.north east) - (0,0.02)$);
    \spy [closeup,magnification=3.5,vibrantblue,size=2.2cm] on ($(FigB)+(-0.03,-0.05)$)
        in node[lowerwindow,anchor=south west] at ($(FigB.south east) + (0,0.02)$);
    \node [anchor=north] at ($(FigB.south)$) {\sf Moved (250$\um \rightarrow$ 100$\um$)};
    
    \node[anchor=south] (FigC) at (1.45,0) {\includegraphics[height=\figheight]{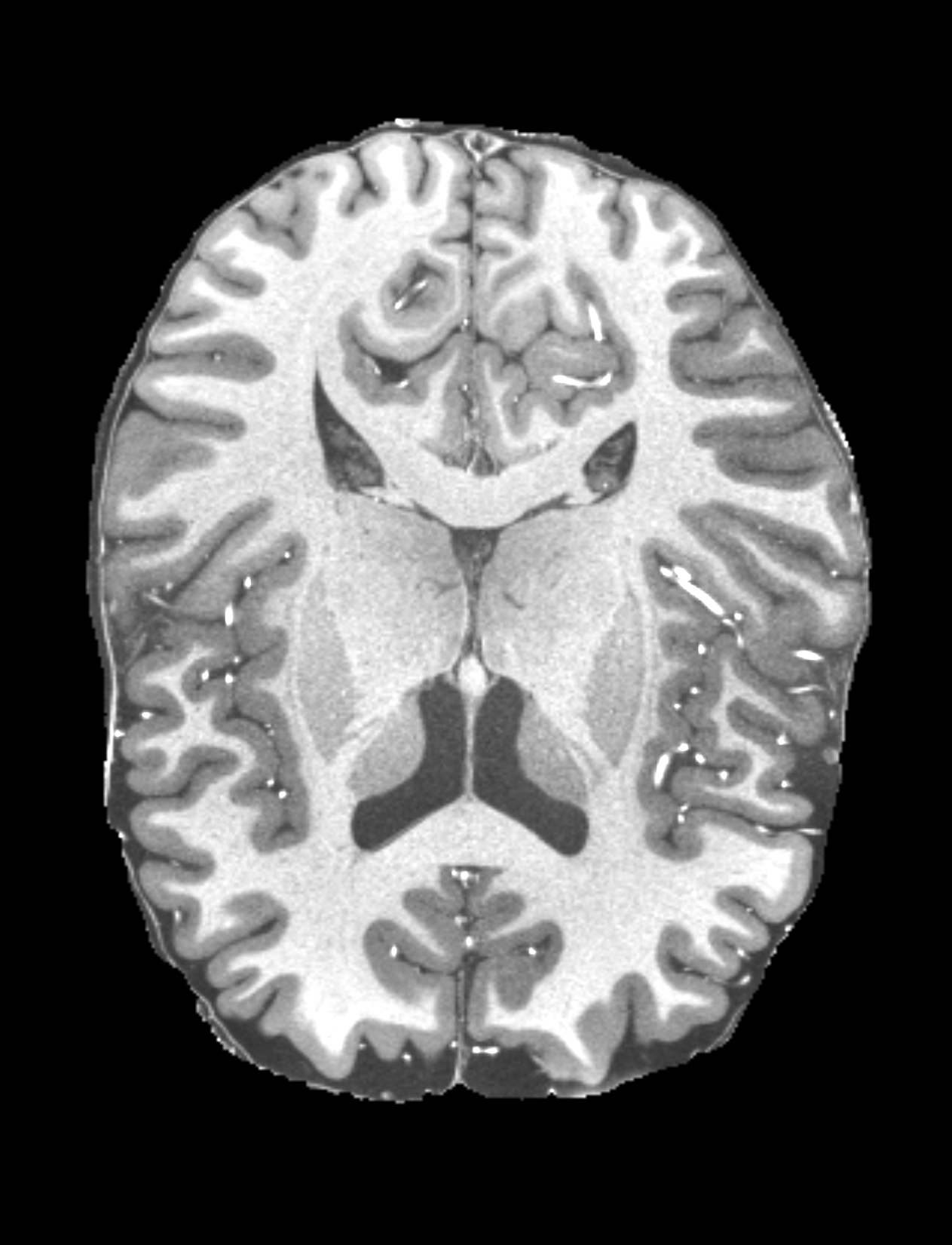}};
    \spy [closeup,magnification=2,vibrantred,size=2.2cm] on ($(FigC)+(+0.05,+0.12)$)
        in node[upperwindow,anchor=north west] at ($(FigC.north east) - (0,0.02)$);
    \spy [closeup,magnification=3.5,vibrantblue,size=2.2cm] on ($(FigC)+(-0.03,-0.05)$)
        in node[lowerwindow,anchor=south west] at ($(FigC.south east) + (0,0.02)$);
    \node [anchor=north] at ($(FigC.south)$) {\sf Moving (250$\um$ T1)};
    \end{tikzpicture}

    }

    \scalebox{0.706}{
    %%% Picture 2
        \begin{tikzpicture}[x=6cm, y=6cm, spy using outlines={every spy on node/.append style={smallwindow}}]
        
        % \begin{tikzpicture}[spy using outlines={every spy on node/.append style={smallwindow}}]
        % Moving image
        \newcommand{\figheight}{4.6cm}
        \node[anchor=south] (FigA) at (-0.75,0) {\includegraphics[height=\figheight]{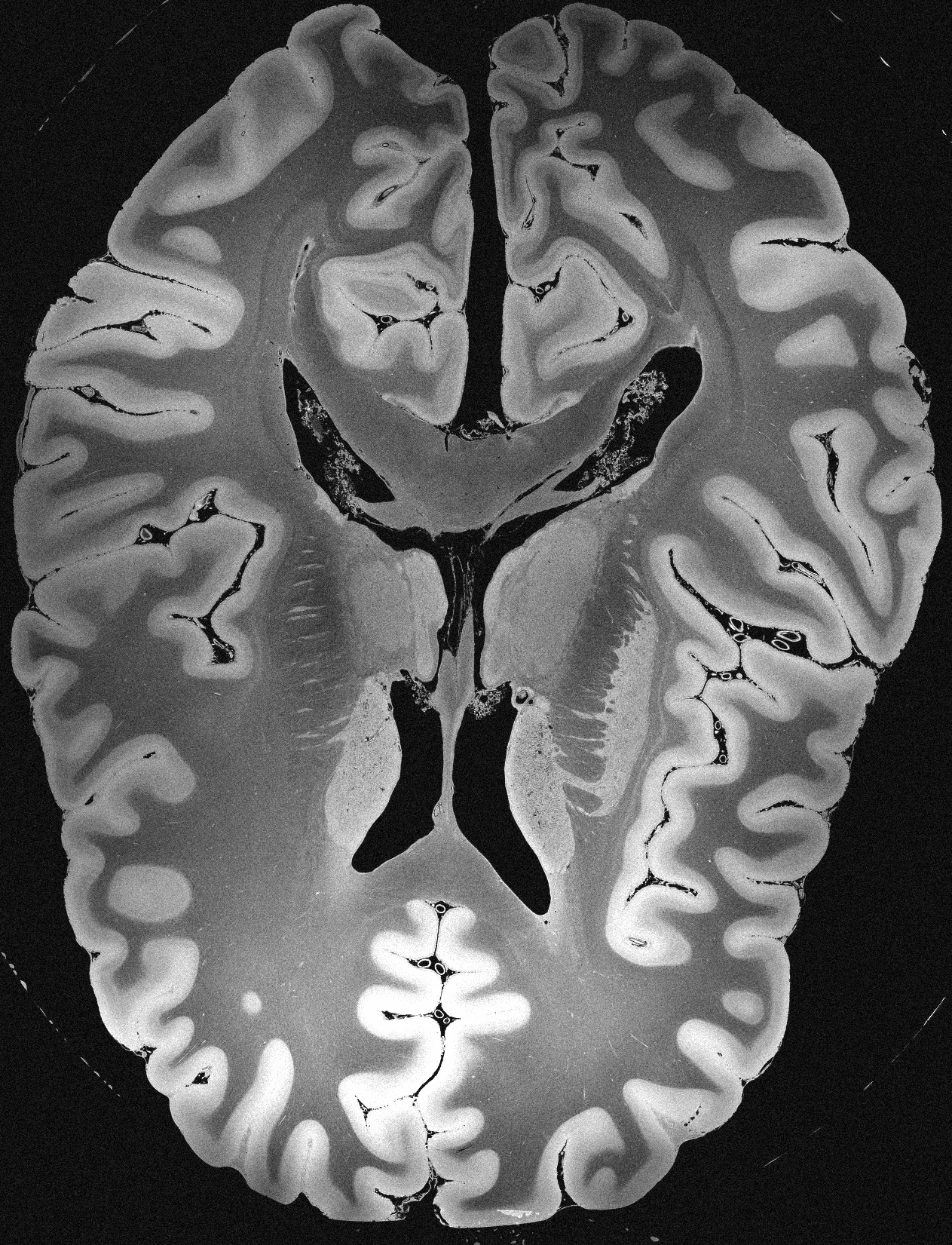}};
        \spy [closeup,magnification=2,vibrantred,size=2.2cm] on ($(FigA)+(+0.0,-0.09)$)
            in node[upperwindow,anchor=north west] at ($(FigA.north east) - (0,0.02)$);
        \spy [closeup,magnification=2.3,vibrantblue,size=2.2cm] on ($(FigA)+(-0.08,+0.09)$)
            in node[lowerwindow,anchor=south west] at ($(FigA.south east) + (0,0.02)$);
        \node [anchor=north] at ($(FigA.south)$) {\sf Fixed (100$\um$ FLASH)};
        
        \node[anchor=south] (FigB) at (0.35,0) {\includegraphics[height=\figheight]{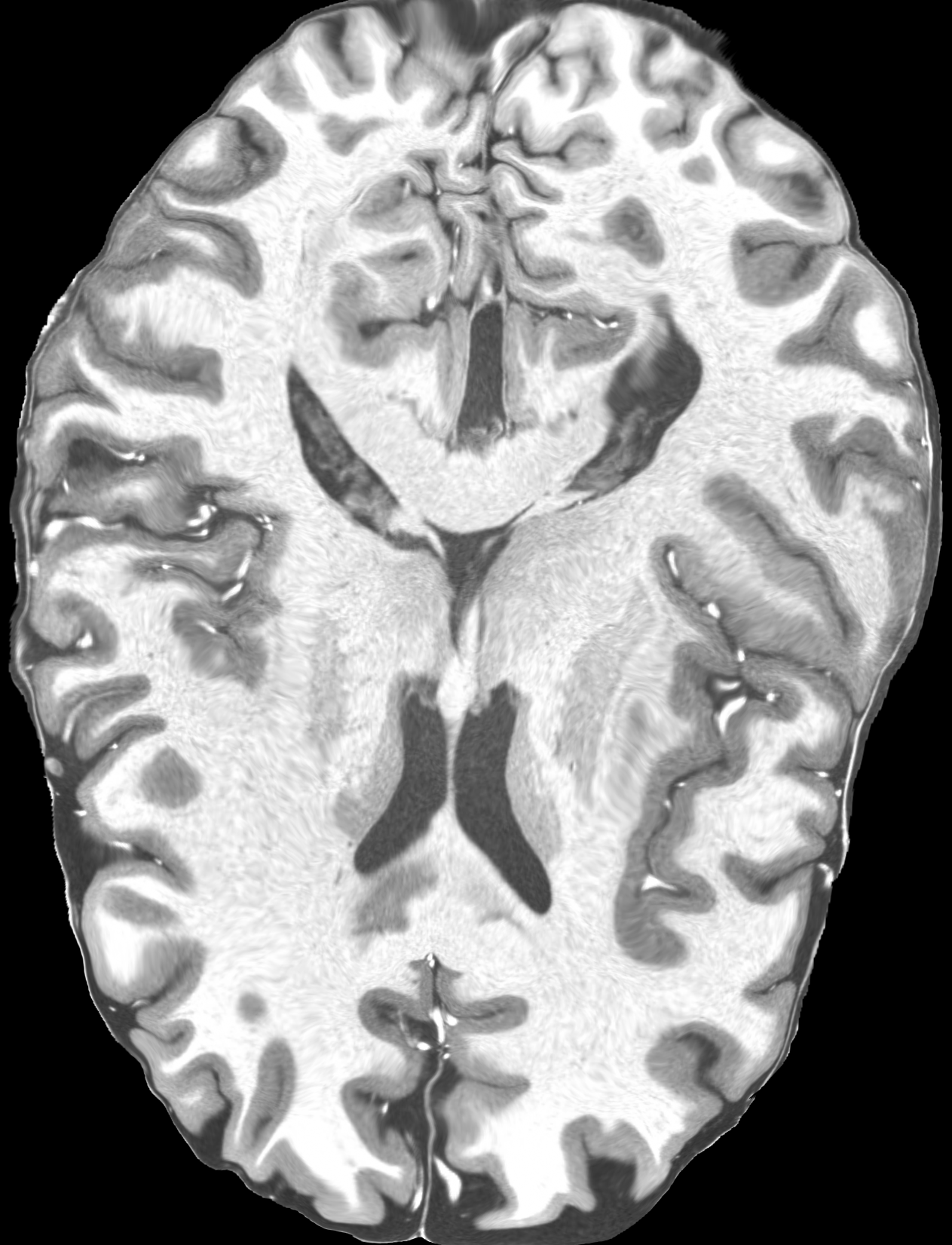}};
        \spy [closeup,magnification=2,vibrantred,size=2.2cm] on ($(FigB)+(+0.0,-0.09)$)
            in node[upperwindow,anchor=north west] at ($(FigB.north east) - (0,0.02)$);
        \spy [closeup,magnification=2.3,vibrantblue,size=2.2cm] on ($(FigB)+(-0.08,+0.09)$)
            in node[lowerwindow,anchor=south west] at ($(FigB.south east) + (0,0.02)$);
        \node [anchor=north] at ($(FigB.south)$) {\sf Moved (250$\um \rightarrow$ 100$\um$)};
        
        \node[anchor=south] (FigC) at (1.45,0) {\includegraphics[height=\figheight]{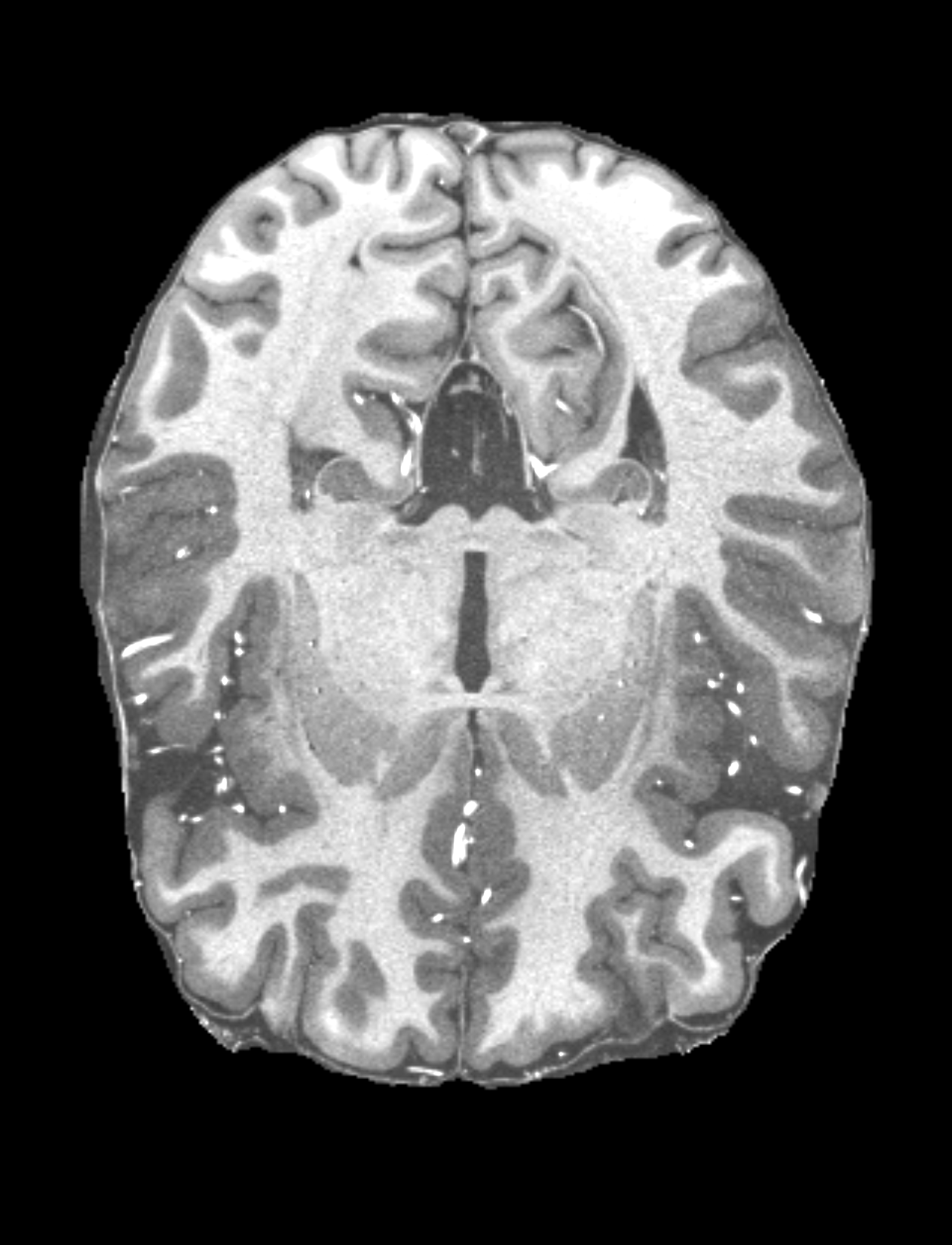}};
        \spy [closeup,magnification=2,vibrantred,size=2.2cm] on ($(FigC)+(+0.0,-0.09)$)
            in node[upperwindow,anchor=north west] at ($(FigC.north east) - (0,0.02)$);
        \spy [closeup,magnification=2.3,vibrantblue,size=2.2cm] on ($(FigC)+(-0.08,+0.09)$)
            in node[lowerwindow,anchor=south west] at ($(FigC.south east) + (0,0.02)$);
        \node [anchor=north] at ($(FigC.south)$) {\sf Moving (250$\um$ T1)};
        \end{tikzpicture}
    }

    \scalebox{0.706}{
    %%% Picture 3
    \begin{tikzpicture}[x=6cm, y=6cm, spy using outlines={every spy on node/.append style={smallwindow}}]
        
        % \begin{tikzpicture}[spy using outlines={every spy on node/.append style={smallwindow}}]
        % Moving image
        \newcommand{\figheight}{3.2cm}
        \node[anchor=south] (FigA) at (-0.75,0) {\includegraphics[height=\figheight]{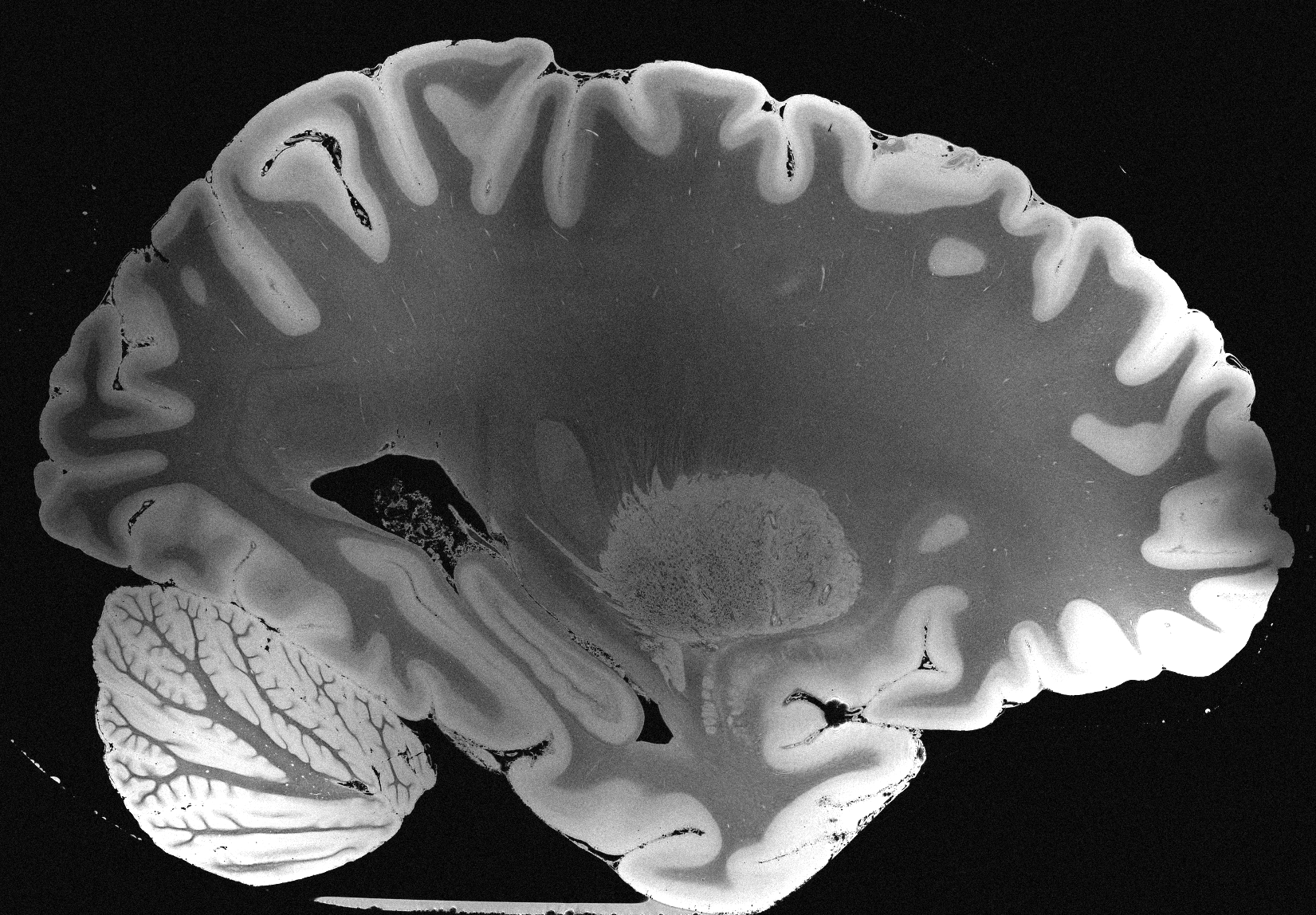}};
        \spy [closeup,magnification=1.6,vibrantred,size=1.5cm] on ($(FigA)+(-0.23,-0.15)$)
            in node[upperwindow,anchor=north west] at ($(FigA.north east) - (0,0.02)$);
        \spy [closeup,magnification=1.6,vibrantblue,size=1.5cm] on ($(FigA)+(-0.11,+0.17)$)
            in node[lowerwindow,anchor=south west] at ($(FigA.south east) + (0,0.02)$);
        \node [anchor=north] at ($(FigA.south)$) {\sf Fixed (100$\um$ FLASH)};
        
        \node[anchor=south] (FigB) at (0.35,0) {\includegraphics[height=\figheight]{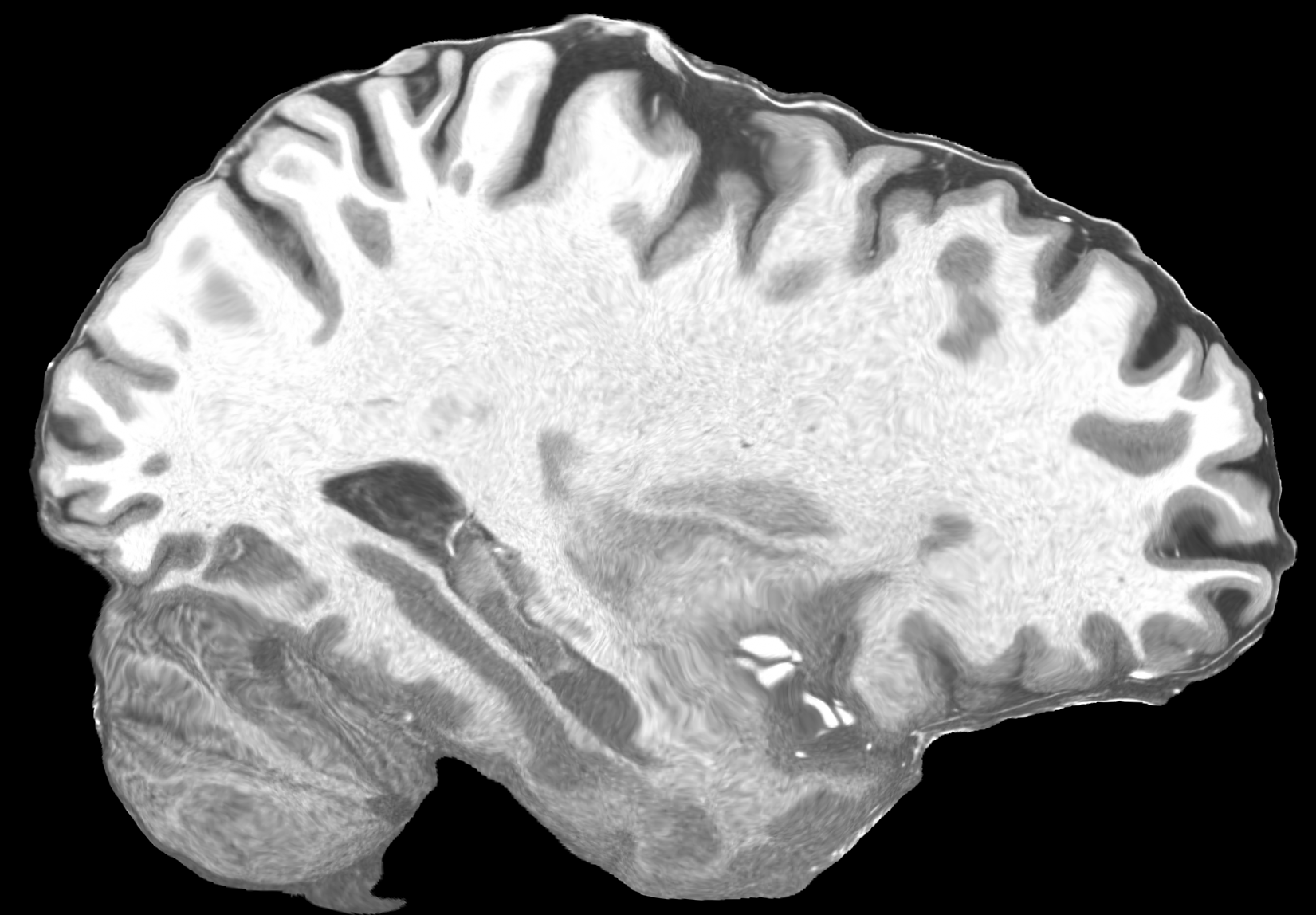}};
        \spy [closeup,magnification=1.6,vibrantred,size=1.5cm] on ($(FigB)+(-0.23,-0.15)$)
            in node[upperwindow,anchor=north west] at ($(FigB.north east) - (0,0.02)$);
        \spy [closeup,magnification=1.6,vibrantblue,size=1.5cm] on ($(FigB)+(-0.11,+0.17)$)
            in node[lowerwindow,anchor=south west] at ($(FigB.south east) + (0,0.02)$);
        \node [anchor=north] at ($(FigB.south)$) {\sf Moved (250$\um \rightarrow$ 100$\um$)};
        
        \node[anchor=south] (FigC) at (1.45,0) {\includegraphics[height=\figheight]{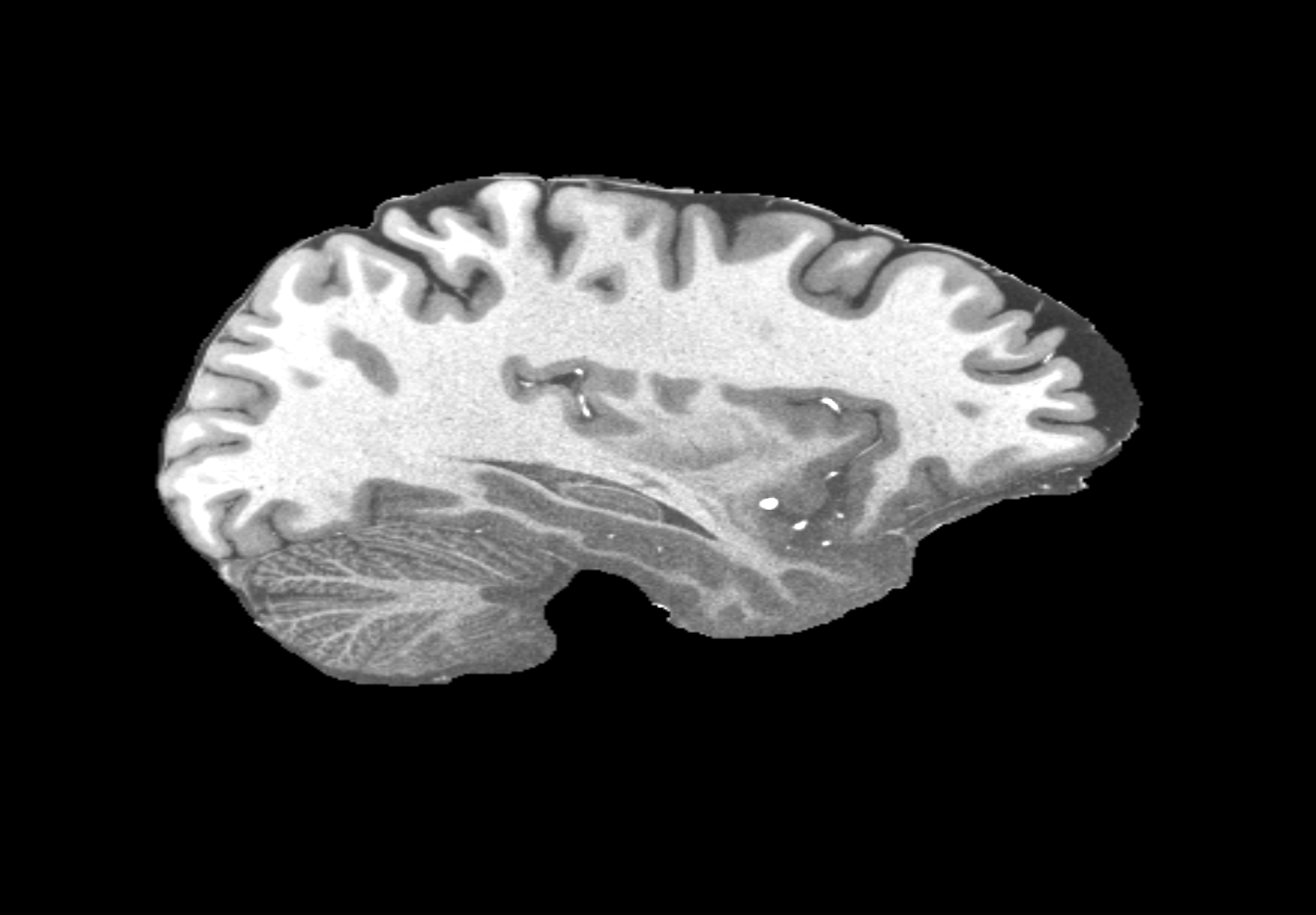}};
        \spy [closeup,magnification=1.6,vibrantred,size=1.5cm] on ($(FigC)+(-0.23,-0.15)$)
            in node[upperwindow,anchor=north west] at ($(FigC.north east) - (0,0.02)$);
        \spy [closeup,magnification=1.6,vibrantblue,size=1.5cm] on ($(FigC)+(-0.11,+0.17)$)
            in node[lowerwindow,anchor=south west] at ($(FigC.south east) + (0,0.02)$);
        \node [anchor=north] at ($(FigC.south)$) {\sf Moving (250$\um$ T1)};
        \end{tikzpicture}
    }

    \scalebox{0.706}{
    %%% Picture 3
    \begin{tikzpicture}[x=6cm, y=6cm, spy using outlines={every spy on node/.append style={smallwindow}}]
        % \begin{tikzpicture}[spy using outlines={every spy on node/.append style={smallwindow}}]
        % Moving image
        \newcommand{\figheight}{3.2cm}
        \node[anchor=south] (FigA) at (-0.75,0) {\includegraphics[height=\figheight]{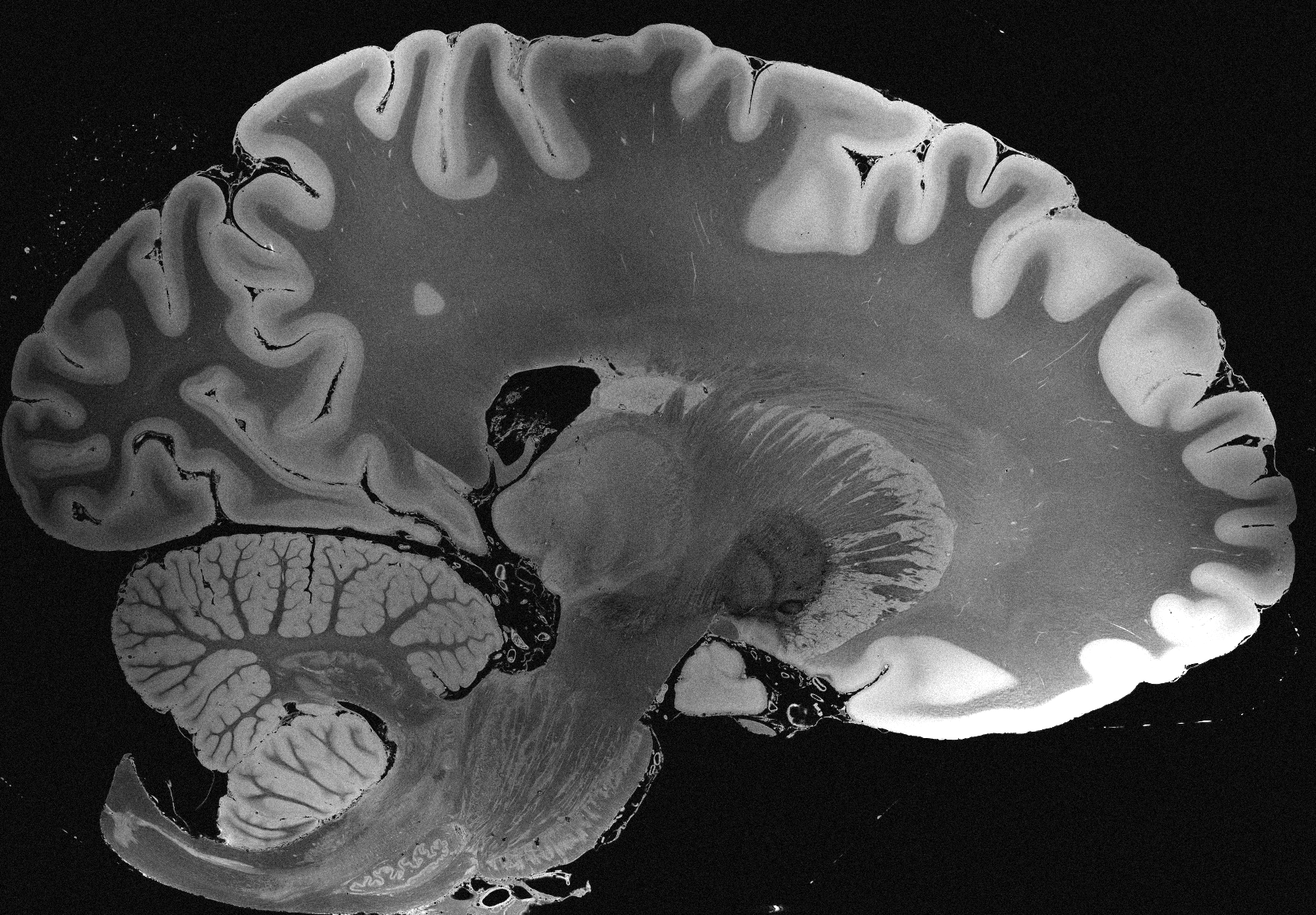}};
        \spy [closeup,magnification=1.4,vibrantred,size=1.5cm] on ($(FigA)+(-0.21,-0.13)$)
            in node[upperwindow,anchor=north west] at ($(FigA.north east) - (0,0.02)$);
        \spy [closeup,magnification=1.4,vibrantblue,size=1.5cm] on ($(FigA)+(0.11,+0.13)$)
            in node[lowerwindow,anchor=south west] at ($(FigA.south east) + (0,0.02)$);
        \node [anchor=north] at ($(FigA.south)$) {\sf Fixed (100$\um$ FLASH)};
        
        \node[anchor=south] (FigB) at (0.35,0) {\includegraphics[height=\figheight]{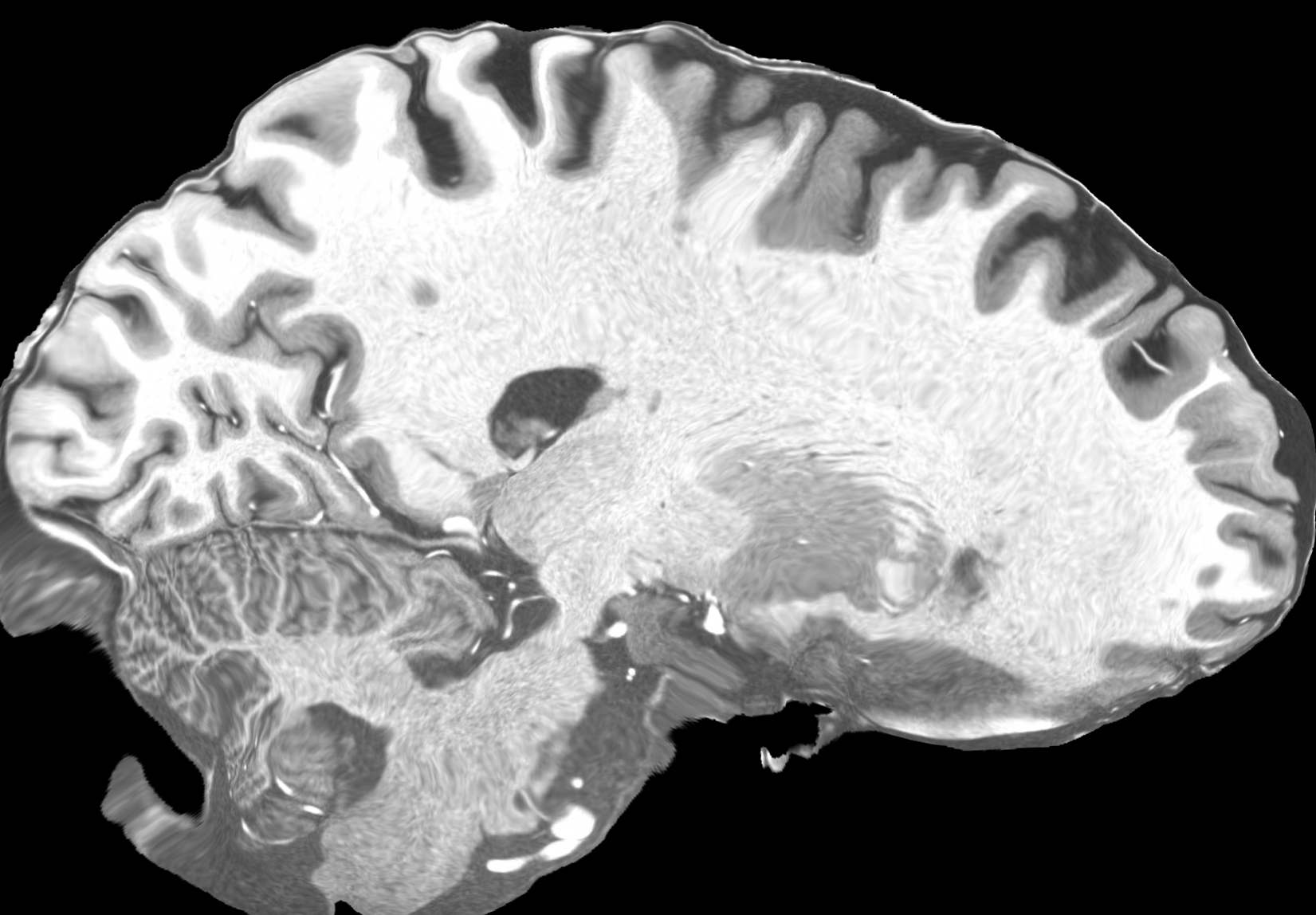}};
        \spy [closeup,magnification=1.4,vibrantred,size=1.5cm] on ($(FigB)+(-0.21,-0.13)$)
            in node[upperwindow,anchor=north west] at ($(FigB.north east) - (0,0.02)$);
        \spy [closeup,magnification=1.6,vibrantblue,size=1.5cm] on ($(FigB)+(0.11,+0.13)$)
            in node[lowerwindow,anchor=south west] at ($(FigB.south east) + (0,0.02)$);
        \node [anchor=north] at ($(FigB.south)$) {\sf Moved (250$\um \rightarrow$ 100$\um$)};
        
        \node[anchor=south] (FigC) at (1.45,0) {\includegraphics[height=\figheight]{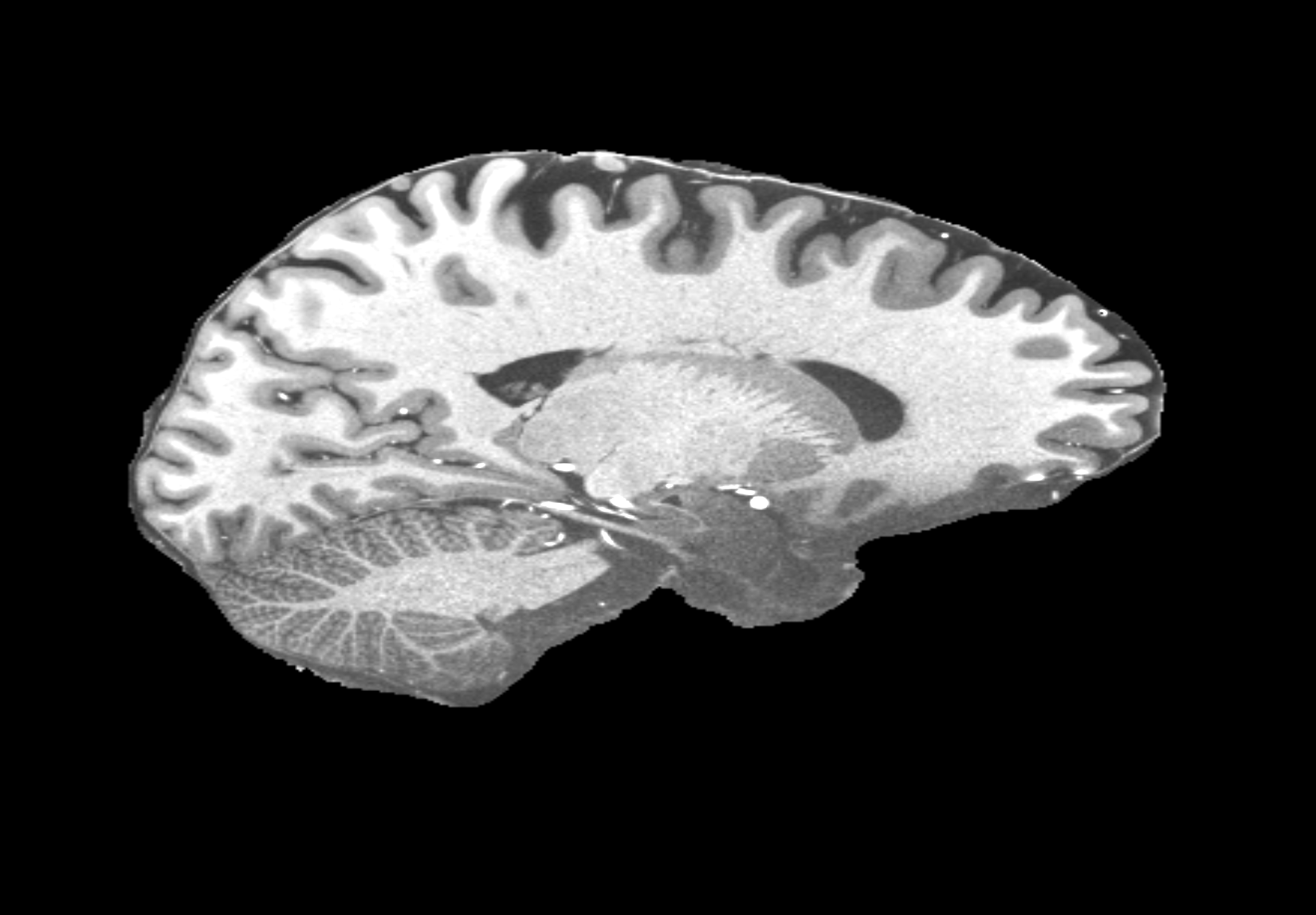}};
        \spy [closeup,magnification=1.4,vibrantred,size=1.5cm] on ($(FigC)+(-0.21,-0.13)$)
            in node[upperwindow,anchor=north west] at ($(FigC.north east) - (0,0.02)$);
        \spy [closeup,magnification=1.6,vibrantblue,size=1.5cm] on ($(FigC)+(0.11,+0.13)$)
            in node[lowerwindow,anchor=south west] at ($(FigC.south east) + (0,0.02)$);
        \node [anchor=north] at ($(FigC.south)$) {\sf Moving (250$\um$ T1)};
        \end{tikzpicture}
    }
    \caption{Qualitative comparison of registration results from 250$\um$ T1 \citep{lusebrink2017t1} to 100$\um$ ex-vivo FLASH \citep{edlow20197}. Intricate structures like cerebellar white matter and GM-WM interfaces are not very discernable at 1mm, but can be aligned at 100$\um$ with our method.}
    \label{fig:qualtitative_registration_comparison_app}
\end{figure}

%% file: sections/app/transmorph-speedup.tex
\section{Accelerating TransMorph training}
\label{sec:transmorph-speedup}

\begin{figure*}[t!]
    \centering
    \includegraphics[width=0.32\linewidth]{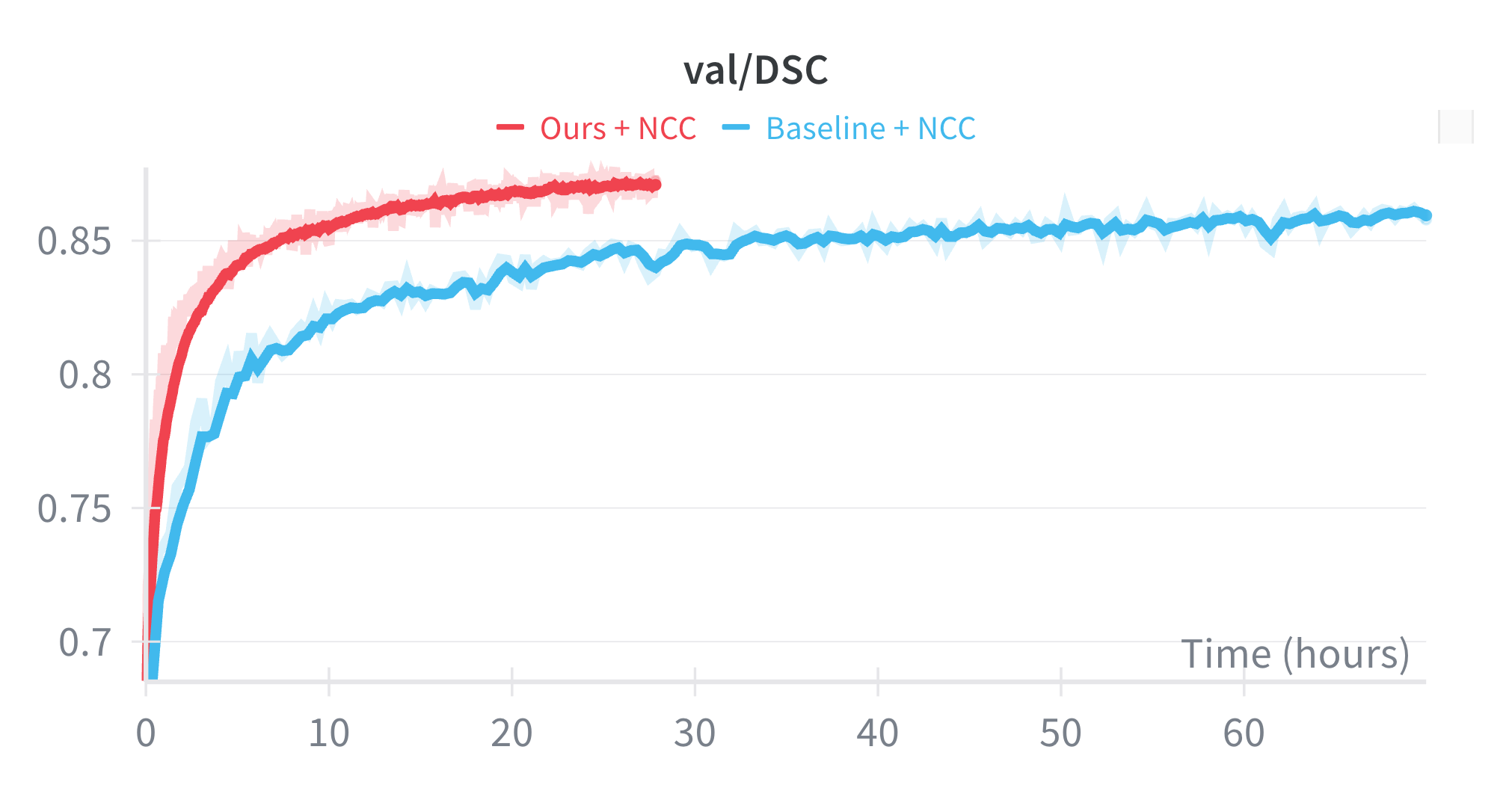}
    \includegraphics[width=0.32\linewidth]{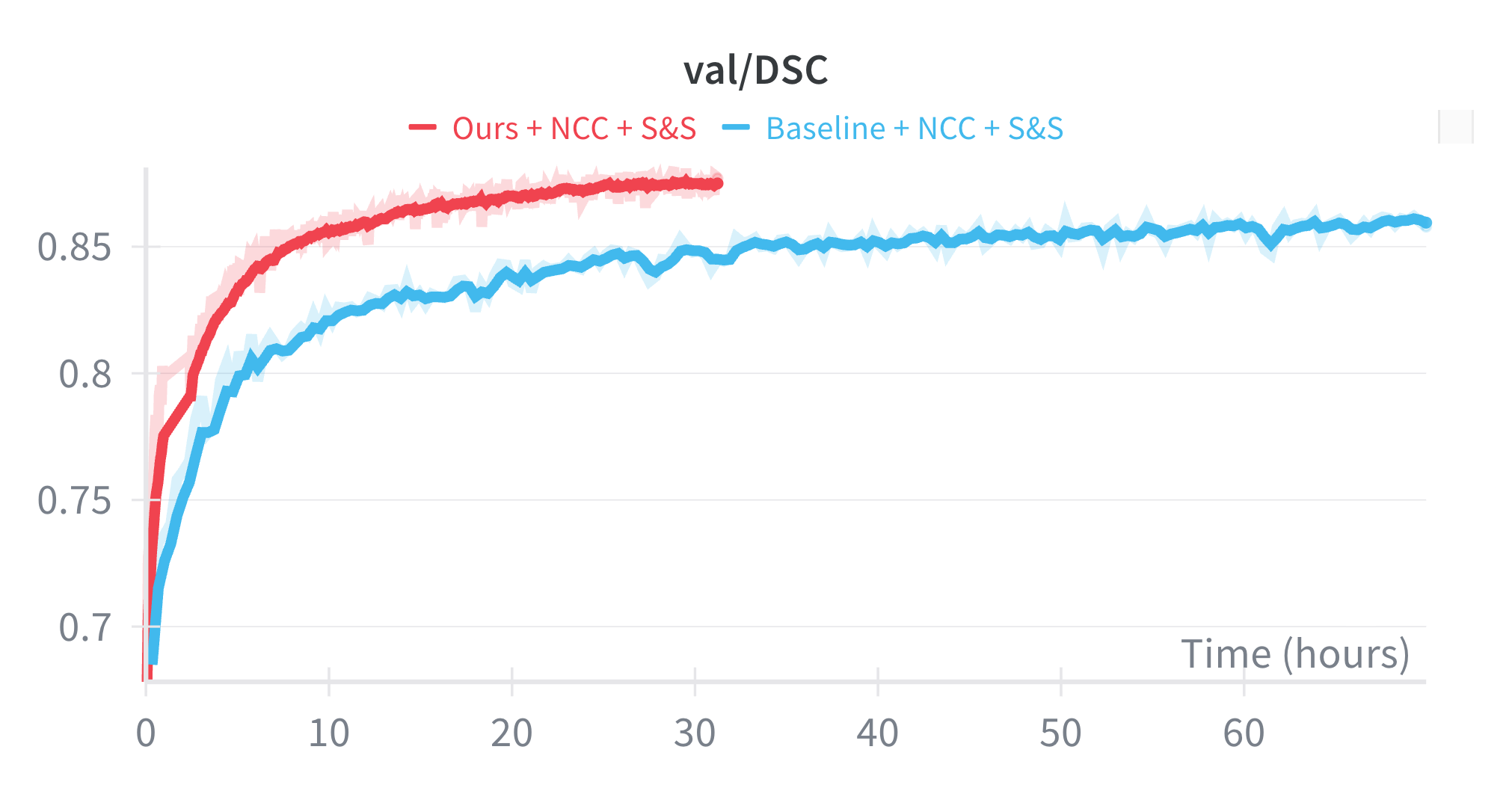}
    \includegraphics[width=0.32\linewidth]{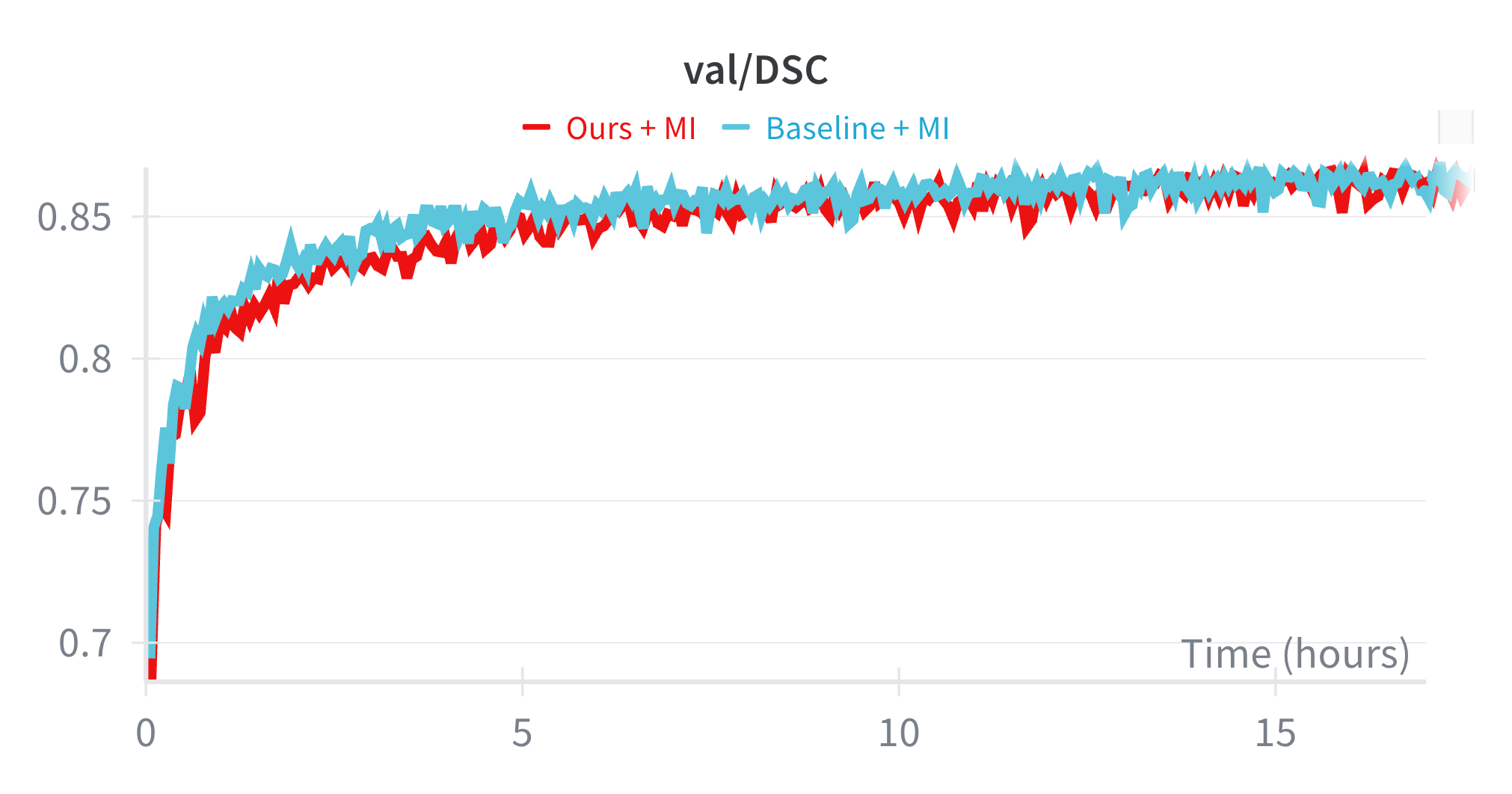}
    \caption{Ablation on TransMorph training runtime with and without our fused operations. For LNCC, our method converges in about 30 hours, while the baseline converges in about a week.}
    \label{fig:transmorph-speedup}
\end{figure*}

In this section, we plot the performance of TransMorph training with and without our fused operations.
\cref{tab:transmorph-speedup} summarizes the performance of TransMorph training with and without our fused operations for three commonly used configurations.
We further plot the validation performance across these settings with respect to Wall clock time in \cref{fig:transmorph-speedup}.
Our fused operations demonstrate efficiency with fast convergence while reducing memory usage.

%% file: sections/app/qualitative-improvements.tex
\subsection{Registration to a 100 micron ex-vivo brain MRI volume}
\label{app:edlow-250um}

In this section, we describe the parameters used for the registration of a 250$\um$ in-vivo T1-weighted MRI volume described in \cite{lusebrink2017t1} to the 100$\um$ ex-vivo brain FLASH volume described in \cite{edlow20197}.
First, we perform an multi-scale affine registration at 3mm, 2mm, 1mm, 500$\um$ resolutions for 500, 250, 100, 100 iterations respectively, using the Fused Mutual Information loss.
This step takes about 12 seconds to run on a single NVIDIA A6000 GPU. 
The second step was to run multi-scale deformable optimization with scales 3.2mm, 1.6mm, 0.8mm, 0.4mm, 0.2mm, 0.1mm (scale factors of 32, 16,8, 4, 2, 1) for 250, 100, 100, 100, 50, 20 iterations respectively, using the fused LNCC loss.
This step took about 58 seconds on 8 NVIDIA A6000 GPUs.
Qualitative results are shown in \cref{fig:qualtitative_registration_comparison_app}.

\begin{table}[t!]
    \centering
        \caption{\textbf{Qualitative Comparison of Methods}. We compare the methods on qualitative features such as GPU support, multimodal capabilities, ability to run for unequal sizes of fixed and moving images, non-standard image sizes, whether the model can work with full context for larger images, whether the model supports multi-GPU training, and whether the model supports arbitrary loss functions.
    Deep learning methods support multimodal registration only if they are trained on multiple modalities.
    CLAIRE requires the image sizes to be divisible by the number of GPUs, and does not support arbitrary loss functions.
    Our method supports all of the above features, leading to a seamless experience for users with minimal data preprocessing overhead.
    }
    \resizebox{\linewidth}{!}{%
    \begin{tabular}{lccccccr}
    \hline
    \textbf{Method} & {\textbf{GPU}} & \textbf{Multimodal} & \textbf{$N\ne M$} & \textbf{Non-std sizes} & \textbf{Full} & \textbf{Multi-} & \textbf{Supported} \\
     & \textbf{support} & & & &\textbf{Context}  & \textbf{GPU} & \textbf{Similarty functions}\\
     \hline
    Deep learning & \yes & \sometimes & \no & \no & \no & \no & Fixed at training \\
    ITK-DReg & \no & \yes & \yes & \yes & \no & \no & ITK-filters \\
    CLAIRE & \yes & \no & \no & \sometimes & \yes & \yes & MSE only \\
    Ours & \yes & \yes & \yes & \yes & \yes & \yes & Any  \\
    \hline
    \end{tabular}
    }
    \label{tab:qualitative-improvements}
\end{table}

%% file: sections/app/fusedcc.tex
\section{An efficient Fused Local Normalized Cross Correlation loss}
\label{app:fusedcc}
Local Normalized Cross Correlation (LNCC) loss is ubiquitously used throughout the image registration literature\citep{vfa,avants_symmetric_2008-1,ants,antsgithub,fireants,wu2024neural,pirate,nodeo,Zhao_2019_ICCV,zhao2019unsupervised}, owing to its robust behavior to unimodal and multimodal images alike.
This operation is a key memory-bound bottleneck in image registration pipelines.
Few approaches have been proposed to provide improved implementations \citep{xijiafastlncc,transmorph}, but we note that these implementations are still memory intensive and thus not scalable.
We address this bottleneck by analytically deriving a fused implementation that is memory efficient and scalable.

\paragraph{Definition of LNCC loss.}
\label{app:fusedcc-definition}
Given two images $F$ and $M$, and a radially symmetric averaging convolution filter $W$ such that $\sum_k w_k = 1$ , we define the Local Normalized Cross Correlation (LNCC) loss as:

\begin{align}
\mathcal{L} &= \frac{1}{N}\sum_i n_i \quad , \quad n_i = \frac{A_i^2}{B_iC_i + \epsilon}
\label{eq:lncc-loss}
\end{align}
where 
\begin{align}
    \mu^F_i, \mu^M_i &= \sum_k w_{ik} F_k , \sum_k w_{ik} M_k \\
    \label{eq:lncc-a}
    A_i &= \sum_k w_{ik} (F_k - \mu^F_i)(M_k - \mu^M_i) \\
    \label{eq:lncc-b}
    B_i &= \sum_k w_{ik}(F_k - \mu^F_i)^2 \\
    C_i &= \sum_k w_{ik} (M_k - \mu^M_i)^2
    \label{eq:lncc-c}
\end{align} 
Here, we use overloaded notation $w_{ik} = w_{(i-k)} = w_{(k-i)} = w_{ki}$ due to radial symmetry of $w$.
We can expand \cref{eq:lncc-a,eq:lncc-b,eq:lncc-c} as follows:
\begin{align}
    A_i &= \left(\sum_k w_{ik} F_k M_k\right) - \mu^F_i\mu^M_i  = \mu^{FM}_i - \mu^F_i\mu^M_i  \label{eq:lncc-a-expanded} \\
    B_i &= \left(\sum_k w_{ik}F_k^2\right) - (\mu^F_i)^2 = \mu^{F^2}_i - (\mu^F_i)^2  \label{eq:lncc-b-expanded} \\
    C_i &= \left(\sum_k w_{ik}M_k^2\right) - (\mu^M_i)^2 = \mu^{M^2}_i - (\mu^M_i)^2  \label{eq:lncc-c-expanded}
\end{align}

\input{algs/lncc}

\begin{figure*}[t!]
    \centering
    \includegraphics[width=\linewidth]{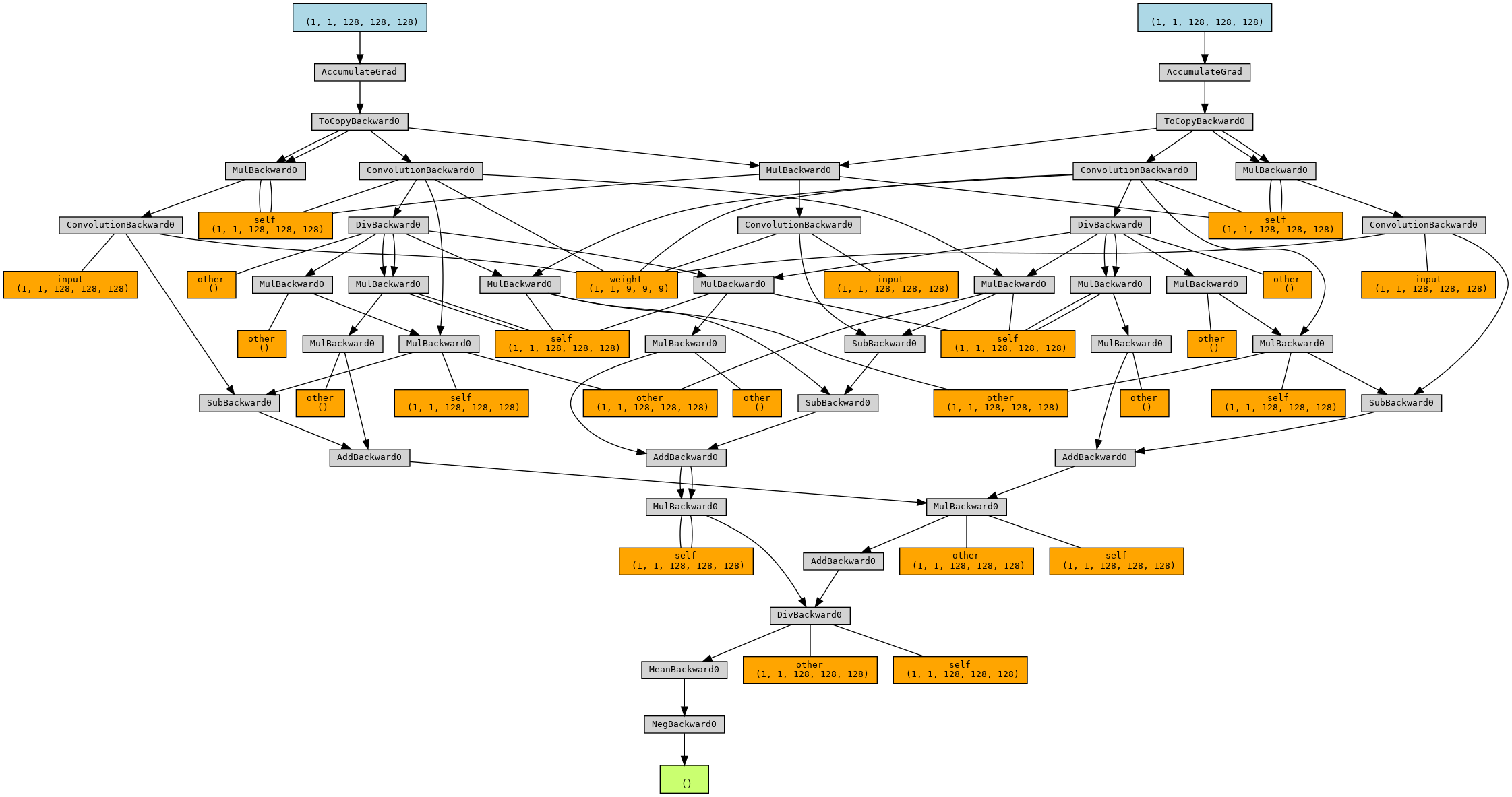}
    \caption{Computational graph of the vanilla PyTorch implementation of the LNCC loss function. \textcolor{blue}{Blue} nodes denote the input images, \textcolor{orange}{Orange} nodes denote intermediate tensors that are stored in HBM, \textcolor{gray}{Gray} nodes denote operations on the computational graph, and \textcolor{ForestGreen}{Green} node denotes the final loss.
    \textcolor{orange}{Orange} nodes are the primary memory bottleneck.
    }
    \label{fig:lncc-computegraph}
\end{figure*}

\cref{alg:lncc} outlines a vanilla PyTorch implementation of the LNCC loss function.
The computational overhead of the algorithm arises due to many intermediates stored in high-bandwidth memory (HBM).
Specifically, the quantities $W * \texttt{state}, I^2, J^2, IJ, \sigma_I^2, \sigma_J^2, \sigma_{IJ}, \mu_I^2, \mu_J^2, \mu_I\mu_J, \sigma_I^2\sigma_J^2, (\sigma_I^2\sigma_J^2 + \epsilon), \sigma_{IJ}^2, \sigma_{IJ}^2/(\sigma_I^2\sigma_J^2 + \epsilon)$ are all stored as intermediate tensors, each of size $N$, totalling a $16N$ memory overhead in addition to storing $\texttt{state}$.
The computational graph of the vanilla PyTorch implementation is shown in \cref{fig:lncc-computegraph}.
During the backward pass, the backprop algorithm computes the gradient with respect to each of these variables costing an additional $16N$ memory overhead.
A \texttt{torch.compile} implementation fuses some of the arithmetic, but leaves a lot of room for improvement (see \cref{fig:ablations}).
We present an algorithm that only requires an additional intermediate variable \texttt{state} of size $5N$, saving upto $27N$ memory.

\subsection{An efficient fused LNCC implementation}
During the forward pass, we initialize a \texttt{state} variable of size $5N$.
To minimize HBM reads from $F$ and $M$, we write a fused kernel to initialize the \texttt{state} variable using only one HBM read from $F$ and $M$.
The code in Line 4-9 are elementwise operations, and can be fused into another kernel.
The forward pass therefore consumes only $5N$ additional memory.
The pseudocode for the efficient fused LNCC implementation is shown in \cref{alg:fusedlncc}.

\paragraph{Efficient Backward Pass}
In a vanilla PyTorch implementation, the gradients are computed for each intermediate variables in the reverse order in the computational DAG shown in \cref{fig:lncc-computegraph}.
Typically, our implementation would also require defining the backward pass by computing the gradients with respect to the intermediate variables, and then propagating them to the input images.
However, we derive the backpropagation with respect to $I$ and $J$, given the gradient $g_i = \frac{\partial L}{\partial n_i}$ to avoid calculating intermediate gradients.
Using the chain rule, we have:
\begin{align}
    \pder{L}{F_k} &= \sum_i \pder{L}{n_i} \pder{n_i}{F_k} \\ 
    &= \sum_i g_i \left( \frac{2A_i}{B_iC_i}\pder{A_i}{F_k} - \frac{A_i}{B_i^2C_i}\pder{B_i}{F_k} \right)\label{eq:gradientcc-first} \\
\end{align}
which can be simplified to:
\begin{align}
   \pder{\mu_i^F}{F_k} = \pder{\mu_i^M}{M_k} = w_{ik}
\end{align}

\begin{align}
    \pder{A_i}{F_k} &= \pder{\left(\sum_k w_{ik}F_kM_k - \mu_i^F\mu_i^M\right)}{F_k} = w_{ik}\left(M_k - \mu_i^M\right)
\end{align}
and 
\begin{align}
    \pder{B_i}{F_k} &= \pder{\left(\sum_k w_{ik} F_k^2 - (\mu^F_i)^2\right)}{F_k} = 2w_{ik}(F_k - \mu_i^F)
\end{align}
Substituting these results to \cref{eq:gradientcc-first} we have:
\begin{align}
    &= \sum_i g_i \left( \frac{2A_i}{B_iC_i}\left(w_{ik}\left(M_k - \mu_i^M\right)\right)- \frac{A_i^2}{B_i^2C_i}2w_{ik}(F_k - \mu_i^F) \right)\label{eq:gradientcc-second} \\
    &= \sum_i \frac{2g_iA_i}{B_iC_i} w_{ik} \left[M_k - \frac{F_k A_i}{B_i} + \mu_i^F\frac{A_i}{B_i} - \mu_i^M\right]
\end{align}

Using the property $w_{ik} = w_{ki}$, and letting $\gamma_i = \frac{2g_iA_i}{B_iC_i}$, we rewrite the previous equation as:
\begin{align}
    &= M_k\cdot \left(\sum_i w_{ki} \gamma_i\right) - F_k\cdot\left(\sum_i w_{ki}\frac{\gamma_iA_i}{B_i}\right) + \sum_i w_{ki}\gamma_i\left(\frac{\mu_i^FA_i}{B_i} - \mu_i^M\right) \\
    &= M_k \cdot (w * \gamma)_k - F_k \cdot (w * \gamma_{AB})_k + (w * \gamma_{FM})_k \label{eq:finalccgradF}
\end{align}
where $\gamma_{AB} = \gamma_i \frac{\mu_i^FA_i}{B_i}$, $\gamma_{FM} = \gamma_i\cdot\left(\frac{\mu_i^FA_i}{B_i} - \mu_i^M\right)$ - and $*$ is the convolution operation.
Similarly, the gradient with respect to the moving image $M_k$ is:
\begin{align}
    \pder{L}{M_k} &= F_k \left(\sum_i w_{ki} \gamma_i\right) - M_k\left(\sum_i w_{ki}\frac{\gamma_iA_i}{C_i}\right) + \sum_i w_{ki}\gamma_i\left(\frac{\mu_i^MA_i}{C_i} - \mu_i^F\right) \\
    &= F_k \cdot (w * \gamma)_k - M_k \cdot (w * \gamma_{AC})_k + (w * \gamma_{MF})_k \label{eq:finalccgradM}
\end{align}
where $\gamma_{AC} = \gamma_i \frac{\mu_i^MA_i}{C_i}$, $\gamma_{MF} = \gamma_i\cdot\left(\frac{\mu_i^MA_i}{C_i} - \mu_i^F\right)$.
To compute the gradients with respect to $F$ and $M$, we need to compute five tensors of the $\gamma$ family, namely $\gamma, \gamma_{AB}, \gamma_{AC}, \gamma_{FM},$ and $\gamma_{MF}$.
This is followed by performing a convolution with all the tensors, and computing elementwise operations given by \cref{eq:finalccgradF} and \cref{eq:finalccgradM}.
The $\gamma$ family of tensors are simple elementwise operations on the \texttt{state} variable, and therefore can be computed by modifying the \texttt{state} variable \textit{inplace} to avoid initializing additional HBM memory.

\paragraph{ANTs gradient approximation.}
In the ANTs implementation, the gradient computation skips performing the convolution of the $\gamma$ family of tensors.
We implement this as an additional flag that the user can toggle as an option for faster backward passes.
All our experiments use this approximation.
% Our ablation on ANTs gradient in \cref{todo} shows no regression in performance.

\input{algs/fusedcc}

\subsection{Performance}
We compare the performance of our fused implementation to various backend implementations.
\cref{fig:ablations} shows the speedup and memory usage over different image sizes; we tabulate the results here.
For this experiment, we initialize two random images of size $N_v \times N_v \times N_v$ and compute the runtime and memory usage for the forward and backward passes.
Results are in \cref{tab:cc-ablation-table}.
% For this experiment, we initialize two random images of size $256 \times 256 \times 256$ and compute the LNCC loss and its gradient using various backends.
Our implementation consistently achieves upto $6\times$ forward time speedup and $\sim 98\times$ backward time speedup compared to \citep{xijiafastlncc} and consumes upto 76\% less memory than a compiled PyTorch implementation and 61.9\% less than a groupwise convolution implementation \citep{xijiafastlncc}.

\begin{table}[t!]
\centering
\caption{Speedup and memory usage of different LNCC backends}
\label{tab:cc-ablation-table}
\resizebox{0.98\linewidth}{!}{%
%%% table
\begin{tabular}{llrrrrrr}
\hline
 \textbf{N}                    & \textbf{Method}                 &   \textbf{Forward} & \textbf{Forward}   &   \textbf{Backward} & \textbf{Backward}   &   \textbf{Memory (MB)} &   \textbf{Memory} \\
 & &   \textbf{Time (s)} & \textbf{Speedup}   &   \textbf{Time (s)} & \textbf{Speedup}   &   &   \textbf{Reduction (\%)} \\
\hline
 \multirow{5}{*}{64}  & Fast LNCC              &              \second{0.001} & 2.95            &               0.003 & 4.86              &          21   &                   61.9 \\
                      & FireANTs               &              0.003 & 7.18             &               \second{0.002} & 3.07              &          25   &                   68   \\
                      & VoxelMorph             &              0.06  & 158.76           &               0.016 & 24.10             &          \second{17}   &                   52.9 \\
                      & \texttt{torch.compile} &              0.003 & 6.83             &               \second{0.002} & 2.30              &          24   &                   66.7 \\
                      & Ours                   &              \best{$<0.001$}     & 1.00             &               \best{0.001} & 1.00              &           \best{8}   &                    0   \\ \hline
 \multirow{5}{*}{128} & Fast LNCC              &              \second{0.008} & 5.88             &               0.026 & 34.09             &         168   &                   61.9 \\
                      & FireANTs               &              0.013 & 9.04             &               0.008 & 10.73             &         200   &                   68   \\
                      & VoxelMorph             &              0.482 & 341.65           &               0.126 & 168.33            &         \second{136}   &                   52.9 \\
                      & \texttt{torch.compile} &              0.012 & 8.67             &               \second{0.007} & 8.95              &         192   &                   66.7 \\
                      & Ours                   &              \best{0.001} & 1.00             &               \best{0.001} & 1.00              &          \best{64}   &                    0   \\ \hline
 \multirow{5}{*}{256} & Fast LNCC              &              \second{0.069} & 6.19             &               \second{0.204} & 82.52             &        \second{1344}   &                   61.9 \\
                      & FireANTs               &              0.103 & 9.25             &               0.294 & 118.80            &        2176   &                   76.5 \\
                      & VoxelMorph             &              3.905 & 351.54           &               3.903 & 1577.37            &        1536.2 &                   66.7 \\
                      & \texttt{torch.compile} &              0.1   & 9.02             &               0.284 & 114.74            &        2176   &                   76.5 \\
                      & Ours                   &             \best{0.011} & 1.00x             &               \best{0.002} & 1.00x              &        \best{512}   &                    0   \\ \hline
 \multirow{5}{*}{512} & Fast LNCC              &              \second{0.627} & 6.56             &               \second{1.657} & 98.75             &       \second{10752}   &                   61.9 \\
                      & FireANTs               &              0.856 & 8.95             &               2.396 & 142.77            &       17408   &                   76.5 \\
                      & VoxelMorph             &             31.335 & 327.71           &              31.665 & 1887.14            &       12288.2 &                   66.7 \\
                      & \texttt{torch.compile} &              0.829 & 8.67             &               2.312 & 137.80            &       17408   &                   76.5 \\
                      & Ours                   &              \best{0.096} & 1.00             &               \best{0.017} & 1.00              &        \best{4096}   &                    0   \\
\hline
\end{tabular}
}
\end{table}

%% file: algs/lncc.tex
\begin{algorithm}
\begin{algorithmic}[1]
\caption{Vanilla PyTorch LNCC implementation}
\label{alg:lncc} % <--- label for entire algorithm
\Require $F$ (input image), $M$ (reference image), $w$ (window size), $reduction$ (reduction type)
% \State Define a convolutional filter $W$ of size $w \times w \times w$ with all elements equal to $\frac{1}{w^3}$
\State Define a radially symmetric convolution filter $W$ of size $w \times w \times w$ with $\sum W[i] = 1$ 
\State Define $\texttt{state} = (F, M, F^2, M^2, FM)$ \hfill \(\triangleright\) Elementwise operations
\State \textbf{Compute} $\texttt{state} = W * \texttt{state}$ \hfill \textcolor{black}{\(\triangleright\) Convolution}
\State Get $\mu_F = \texttt{state}[0]$ \hfill \(\triangleright\) Local mean of $F$ \label{line:lnccstart}
\State Get $\mu_M = \texttt{state}[1]$ \hfill \(\triangleright\) Local mean of $M$
\State Compute $\sigma_F^2 = \texttt{state}[2] - \mu_F^2$ \hfill \(\triangleright\) Local variance of $F$
\State Compute $\sigma_M^2 = \texttt{state}[3] - \mu_M^2$ \hfill \(\triangleright\) Local variance of $M$
\State Compute $\sigma_{FM} = \texttt{state}[4] - \mu_F \cdot \mu_M$ \hfill \(\triangleright\) Local covariance of $F$ and $M$
\State Compute $\text{LNCC} = \frac{\sigma_{FM}^2}{{\sigma_F^2 \sigma_M^2} + \epsilon}$ \hfill \(\triangleright\) Add small $\epsilon$ to avoid division by zero
\If{\texttt{reduction} == \texttt{NONE}}
    \State \Return LNCC
\Else
    \State Compute loss: $\text{Loss} = 1 - \text{mean}(\text{LNCC})$
    \State \Return Loss
\EndIf
\end{algorithmic}
\end{algorithm}

%% file: algs/fusedcc.tex
\begin{algorithm}
\caption{Fused LNCC Implementation}
\label{alg:fusedlncc}
\begin{algorithmic}[1]

\Require $F$ (fixed image), $M$ (moving image), $w$ (window size), $\epsilon$ (smoothing term)

\Function{Forward}{$F$, $M$, $w$, $\epsilon$}
    \State Define convolution filter $W$ of size $w \times w \times w$ with $\sum W[i] = 1$
    \State $\texttt{state} \leftarrow \texttt{fused\_create\_interm}(F, M)$ \hfill $\triangleright$ Single HBM read: $(F, M, F^2, M^2, FM)$
    \State $\texttt{state} \leftarrow W * \texttt{state}$ \hfill $\triangleright$ Convolution on all channels
    \State $\text{LNCC} \leftarrow \texttt{fusedcc\_kernel}(\texttt{state}, \epsilon)$ \hfill $\triangleright$  Computes \cref{eq:lncc-a-expanded,eq:lncc-b-expanded,eq:lncc-c-expanded} followed by \cref{eq:lncc-loss}
    \State \Return LNCC
\EndFunction \\

\Function{Backward}{$g = \pder{\mathcal{L}}{n}$, $\texttt{state}$, $F$, $M$, $W$, use\_ants\_approximation}
    % \State Extract $\mu^F, \mu^M, \mu^{F^2}, \mu^{M^2}, \mu^{FM}$ from $\texttt{state}$
    % \State Compute $A \leftarrow \mu^{FM} - \mu^F \odot \mu^M$ \vedant{[replace with $\ast$ ? confusing when its used for multiplication]}
    % \State Compute $B \leftarrow \mu^{F^2} - (\mu^F)^2$, $C \leftarrow \mu^{M^2} - (\mu^M)^2$
    % \Comment $\Gamma \gets [\gamma, \gamma_{AB}, \gamma_{AC}, \gamma_{FM}, \gamma_{MF}]$
    \State $\texttt{state} \leftarrow \texttt{fused\_compute\_gamma}(g, \texttt{state})$ \hfill $\triangleright$  Computes $\gamma$ family of tensors \textit{inplace}
    \If{use\_ants\_approximation}
        \State \texttt{no-op} \Comment{ANTs approximation: skip convolutions}
    \Else
        \State $\texttt{state} \leftarrow W * \texttt{state}$ \hfill $\triangleright$ Convolution on all intermediates
    \EndIf
    \State $\frac{\partial L}{\partial F} \leftarrow M \odot \gamma - F \odot \gamma_{AB} + \gamma_{FM}$ \hfill $\triangleright$ \cref{eq:finalccgradF} computed in fused kernel
    \State $\frac{\partial L}{\partial M} \leftarrow F \odot \gamma - M \odot \gamma_{AC} + \gamma_{MF}$ \hfill $\triangleright$ \cref{eq:finalccgradM} computed in fused kernel
    \State \Return $\frac{\partial L}{\partial F}$, $\frac{\partial L}{\partial M}$
\EndFunction

\end{algorithmic}
\end{algorithm}

%% file: sections/app/fusedmi.tex
\section{A highly efficient Mutual Information implementation}
\label{app:fusedmi}
% The mutual information is a commonly used operation in multimodal image registration, \todo{more generic statement here}.
Mutual Information (MI) is one of the most commonly used loss functions for \textit{multimodal} image matching \citep{transmorph,ants,mattesmi}.
Beyond multimodal image matching, MI is a cornerstone operation in computer vision \citep{miboundary,miregion}, contrastive learning \citep{micontrastive}, remote sensing \citep{miremotesensing}, graph learning \citep{migraph}, ecological and social community interactions \citep{mi_network,mi_network_corso}, and cosmological dynamics \citep{mi_galaxy}.
% The Mattes Mutual Information (MI) \citep{mattesmi} is ubiquitous in various engineering and scientific applications spanning across computer vision\citep{}, self-supervised learning, remote sensing and hyperspectral imaging, representation learning on graphs, community detection and ecological interactions, cosmological dynamics and computational linguistics.
In biomedical imaging and life sciences, MI is used for multimodal image alignment using the assumption that pixels in multimodal images codify some nonlinear function of the underlying tissue type.

\paragraph{Vanilla MI implementation}
Given images $I$ and $J$, Mattes MI considers the intensities from the images as samples from probability distributions $p_I$ and $p_J$ that encode some imaging physics. 
The intensity pairs $(I_k, J_k)$ are considered to be samples from the joint distribution $p_{IJ}$.
If the images are aligned, then $I_k$ and $J_k$ are highly `predictable' from each other, implying a low conditional entropy $H(I|J)$, or equivalently a large distance from the distribution $p_Ip_J$ which models the joint distribution if samples from $I$ and $J$ were independent.
This is precisely the mutual information criteria.
Since the samples $I_k, J_k$ follow some unknown distributions, we use a kernel density estimator using kernel $\kappa$ to estimate the empirical distributions of the joint and marginal distributions.
% Furthermore, these density estimates are in the continuous space, with no closed form analytical integrals. 
To compute empirical MI, the continuous kernel density estimates are discretized into a probability mass function (PMF) with a finite number of bins.
%  we discretize the density functions using a parameter $B$ specifying the number of bins to discretize the domains of $I$ and $J$ into.
The number of bins $B$ is a hyperparameter that is used to define bin centers $b_i \in [0,1]$ for $i = \{1, \ldots, B\}$, assuming that the intensities are scaled to the range $[0,1]$.

To compute the discrete PMF with autodifferentiation, we compute a Parzen Block $\Psi_I\in \reals^{B\times N} , \text{s.t.} \Psi_I(i, k) = \kappa(b_i - I_k)$. 
This forms the memory bottleneck in computing the Mattes MI similarity criteria. In the following, we provide a fused implementation that avoids the $O(NB)$ cost of the Parzen Block, making our implementation only $O(1)$ additional HBM overhead.

\subsection{Implicit MI implementation}
We implement custom forward and backward passes to compute the joint and marginal histograms $p_{IJ}, p_I, p_J$ from $I$ and $J$ directly, avoiding the $O(NB)$ cost of the Parzen Block. 
We derive the backward pass first, followed by the forward pass followed by an efficient approximate estimator of the histograms leading to a faster forward pass.

\subsubsection{Backward pass}
We are interested in computing the gradients $\pder{L}{I}, \pder{L}{J}$ given $\pder{L}{p_{IJ}}, \pder{L}{p_I}, \pder{L}{p_J}$.
We denote $\omega(b_i - I_k) = \pder{\kappa(b_i - I_k)}{I_k}$.

\begin{align}
    \pder{L}{I_k} &= \sum_{m,n}\pder{L}{p_{IJ}[m,n]} \pder{p_{IJ}[m,n]}{I_k} + \sum_n \pder{L}{p_I[n]}\pder{p_I[n]}{I_k} \\
    &= \sum_{m,n} g_{IJ}[m,n]\left(\omega(b_m - I_k)\kappa(b_n - J_k)\right) + \sum_n g_{I}[n] \left(\omega(b_n - I_k)\right)  \\
    &= \sum_n \left[ \textcolor{red}{g_{I}[n]\omega(b_n - I_k)} + \textcolor{ForestGreen}{\sum_m g_{IJ}[m, n]\omega(b_m - I_k)} \right] \label{eq:gradientmi-threadn} = \sum_n \textcolor{red}{\zeta_1[n]} + \textcolor{ForestGreen}{\zeta_2[n]}
\end{align}
where $\zeta_1[n] = g_{I}[n]\omega(b_n - I_k)$ and $\zeta_2[n] = \sum_m g_{IJ}[m, n]\omega(b_m - I_k)$.

To compute this backward pass efficiently, we launch $\ceil{N/B}$ threadblocks and partition each threadblock in groups of $B$ threads, and compute the partial gradients $\zeta_1[n], \zeta_2[n]$ on each thread.
Each group loads the values of $I_k, J_k$ into register memory.
we first compute the quantities $\kappa(b_n - I_k), \kappa(b_n - J_k), \omega(b_n - I_k), \omega(b_n - J_k)$ on thread $n$ and use four shared memory arrays to store them.
On thread $n$, we compute the partial gradient $\zeta_1[n] = g_I[n]\omega(b_n - I_k)$ and $\zeta_2[n] = \sum_m g_{IJ}[m, n]\omega(b_m - I_k)$ using a for-loop over the index $m \in \{1, \ldots, B\}$.
Finally, on each thread we store the value $\zeta_1[n] + \zeta_2[n]$ on shared memory indexed at $n$, followed by a $O(\log(n))$ parallel sum over partitioned threads to compute the gradient $\pder{L}{I_k} = \sum_n \zeta_1[n] + \zeta_2[n]$.
A similar argument is used to compute the gradient over $\pder{L}{J_k}$.
This leads to a faster backward pass than the vanilla PyTorch implementation using no additional HBM overhead ~\cref{fig:ablations}(b).

\paragraph{Generalization to novel kernels}
Note that unlike the vanilla implementation, where some choices of $\kappa$ are more memory intensive than others (for example, the BSpline kernel has $k_P = 14$ versus $k_P = 4$ for the Gaussian kernel), the memory overhead of our implementation does not depend on the analytical form of $\kappa$.
To generalize the Implicit MI implementation to novel kernels, the user can specify the form of $\kappa$ and its derivative $\omega$ in the forward and backward passes without any additional considerations. 

\subsubsection{Forward Pass}
The forward pass is computed similarly.
Note that the individual contributions from $I_k, J_k$ to the joint histogram $p_{IJ}[m, n]$ are $p_{IJ}[m, n] = \kappa(b_m - I_k)\kappa(b_n - J_k)$ for all $m, n \in \{1, \ldots, B\}$.
The marginal histograms $p_I[n], p_J[n]$ are computed as $p_I[n] = \kappa(b_n - I_k)$ and $p_J[n] = \kappa(b_n - J_k)$ for all $n \in \{1, \ldots, B\}$.
Similar to the backward pass, we launch $\ceil{N/B}$ threadblocks and partition the threadblock in groups of $B$ threads.
Each group of $B$ threads loads the values of $I_k, J_k$ into register memory.
On thread $n$, we compute the quantities $\kappa(b_n - I_k), \kappa(b_n - J_k)$ and store them in shared memory.
Thread $n$ can add these quantities into the HBM for histogram entries $p_I[n], p_J[n]$ directly.
For computing the joint histogram $p_{IJ}[m, n]$, thread $n$ loops over $m \in \{1, \ldots, B\}$ and adds the quantities $\kappa(b_m - I_k)\kappa(b_n - J_k)$ into the HBM for histogram entries $p_{IJ}[m, n]$.
Since all values of $\kappa(b_m - I_k), \kappa(b_n - J_k)$ are stored in shared memory, this operation is not bottlenecked by slow HBM reads.
To avoid HBM write contentions, we write these values into intermediate histogram buffers of sizes $C\times B \times B, C\times B$ (where $C$ is a constant of choice), and sum along the $C$ dimension.
However, this is still a relatively slow operation due to computation of $\kappa(b_m - I_k), \kappa(b_n - J_k)$ and making $NB^2$ HBM writes.
We propose an efficient approximate forward pass that launches only $N$ instead of $NB$ threads, and makes only $3N$ HBM writes.

\paragraph{An approximate histogram estimator}
Given a kernel $\kappa$, we can write $\kappa(b_m - I_k) = \int_t \delta(b_m - I_k - t) \kappa(t) dt = \delta(b_m - I_k) * \kappa$, where $\delta$ is the Dirac delta function with the property $\int_{x = -\infty}^{\infty} \delta(x) f(x) dx = f(0)$ for any function $f$.
Using the principle of superposition, we can write $p_{I}[m] = \frac{1}{N}\sum_k \kappa(b_m - I_k) = \frac{1}{N}\sum_k \kappa * \delta(b_m - I_k) = \kappa * \left(\frac{1}{N}\sum_k \delta(b_m - I_k) \right)$.

In the continuous case, $p_{I}$ can be obtained \textit{exactly} by calculating the Dirac delta distribution $p_{I}^\delta(b) = \frac{1}{N}\sum_k \delta(b - I_k)$ and convolving it with the kernel $\kappa$.
However, in the discrete case, this value is inexact.
To see this, consider a value $I_k$ that is in bin $m$, i.e. $\|I_k - b_m\| < \frac{1}{2B}$.
The exact value of the PMF due to this sample is $\kappa(b_m - I_k)$.
However, the approximate value of the PMF is $\kappa(0)$ since $\delta(b_m - I_k) = 1$ for all $I_k: \|I_k - b_m\| < \frac{1}{2B}$ due to binning, and convolving with $\kappa$ returns $\kappa(0)$.
Since $\|I_k - b_m\| < \frac{1}{2B}$, we can assume that $\|\kappa(0) - \kappa(b_m - I_k)\|$ is small.

To implement this histogram computation efficiently, we launch $N$ threads and in each thread $k$, compute the bin indices $m^* = \lfloor I_k B\rfloor, n^* = \lfloor J_k B\rfloor$ for each thread, avoiding computation of \textit{soft entries} $\kappa(b_m - I_k), \kappa(b_n - J_k)$ altogether.
We simply add 1 to the histogram entries $p_{IJ}[m^*, n^*], p_I[m^*], p_J[n^*]$ in the aggregated histogram buffers, avoiding writing into HBM entries for all $(m, n) \in \{1, \ldots, B\}^2$.
This reduces the number of HBM writes from $NB^2 + 2NB$ to $3N$.
For $B=32$, this represents $362\times$ less HBM writes.
After performing the average, we convolve this histogram with the kernel $\kappa$ to get the approximate PMF.
Since the convolution is done on a $B$ and $B\times B$ sized histograms, this operation is cheap.
This implementation leads to faster runtime, consistent performance for both TransMorph and FireANTs (see \cref{tab:transmorph-speedup}).

%% file: sections/app/gridsampler.tex
\section{Composite Implicit Grid Sampler}
\label{app:gridsampler}

\input{algs/gridsampler}

%% file: algs/gridsampler.tex
\begin{algorithm}[t!]
\caption{Grid Sampler Implementation}
\label{alg:gridsampler}
\begin{algorithmic}[1]
\Require $J_h$ (moving image shard), $\gridu_j$ (warp field shard), $A_h$ (rescaled affine), $t_h$ (rescaled translation), $S_h$ (diag. scale)
\Function{Forward}{$J_h,\ A_h,\ t_h,\ S_h,\ \gridu_j$}
    \State $\texttt{out} \leftarrow \texttt{zeros\_like}(\gridu_j[0])$
    \ForAll{target voxels $(z,y,x)$ \textbf{in parallel (one thread per voxel)}}
        \State $X \leftarrow (x,y,z)$
        \State $X_{\text{aff}} \leftarrow A_h X + t_h$ \hfill $\triangleright$ \textit{affine transform only}
        \State $X_{\text{disp}} \leftarrow S_h\,\gridu_j[:,z,y,x]$ \hfill $\triangleright$ \textit{add scaled displacement}
        \State $X_{\text{src}} \leftarrow X_{\text{aff}} + X_{\text{disp}}$
        \State $\texttt{out}[z,y,x] \leftarrow \texttt{trilinear\_interpolate}(J_h,\ X_{\text{src}})$ \hfill \textit{zero padding at bounds}
    \EndFor
    \State \Return \texttt{out}
\EndFunction \\

\Function{Backward}{$g = \pder{\mathcal{L}}{\texttt{out}}$, $J_h,\ A_h,\ t_h,\ S_h,\ \gridu_j$}
    \State Initialize $g_{J_h}=0,\ g_{\gridu_j}=0,\ g_{A_h}=0,\ g_{t_h}=0$
    \ForAll{target voxels $(z,y,x)$ \textbf{in parallel (one thread per voxel)}}
        \State Recompute $X,\ X_{\text{aff}},\ X_{\text{disp}},\ X_{\text{src}}$
        \State Compute tri-linear weights $w_{b_x b_y b_z}$ and $\frac{\partial v}{\partial X_{\text{src}}}$
        \State Accumulate $g_{J_h}$ into 8 neighbors using $w_{***}\cdot g[z,y,x]$ \textit{(bounds-checked, zero-padded)}
        \State $g_{\gridu_j}[:,z,y,x] \mathrel{+}= S_h\ \frac{\partial v}{\partial X_{\text{src}}}\ g[z,y,x]$
        \State $g_{A_h} \mathrel{+}= \left(\frac{\partial v}{\partial X_{\text{src}}}\ g[z,y,x]\right) X^\top$
        \State $g_{t_h} \mathrel{+}= \frac{\partial v}{\partial X_{\text{src}}}\ g[z,y,x]$
    \EndFor
    \State \Return $g_{J_h},\ g_{\gridu_j},\ g_{A_h},\ g_{t_h}$
\EndFunction
\end{algorithmic}
\end{algorithm}

%% file: sections/app/ringsampler.tex
\section{Ring Sampler for scalable distributed interpolation}
\label{app:ringsampler}
The random-access nature of deformable interpolation making scaling a difficult challenge for arbitrarily large problem sizes.
Given a configuration of sharded images and warp fields across $H$ hosts, neighboring voxels in the sharded warp field can point to pixels in arbitrary regions in the image, illustrated in \cref{fig:ring-sampler}(a).
Moreover, the control points for interpolation can be irregularly distributed across different hosts, illustrated in \cref{fig:ring-sampler}(b).
This makes computation of the interpolated image challenging for displacements that point to pixels between boundaries of different hosts.
One approach to avoid this problem is to store the entire moving image on each GPU to compute the interpolated image.
However, this approach is impractical once the image size exceeds the memory per GPU.
To achieve weak scaling, the HBM overhead per GPU must be proportional to $N/H$.
To alleviate this problem, we propose a ring sampler that avoids the need to store the entire moving image on each GPU by decomposing linear interpolation into partial sums.
This produces mathematically correct interpolated images regardless of the nature of the warp field, without storing the entire moving image on each GPU.

\subsection{Derivation}
Consider a $d$-linear interpolation of an image $I$ defined on $\Omega$ using warp coordinates $\gridu_\Omega$ defined on $\Omega$.
\begin{align}
    I = \sum_{b\in\{0,1\}^n}
\left(\prod_{k=1}^n (1-\alpha_k)^{1-b_k}\,\alpha_k^{b_k}\right)
\;I\big[i_1+b_1,\ i_2+b_2,\ \dots,\ i_n+b_n\big]
\label{eq:interp}
\end{align}
where $i_k = \lfloor \varphi(x)_k\rfloor,\qquad \alpha_k = \varphi(x)_k - i_k,\qquad \text{for }k=1,\dots,d$.
Let the individual pixels $I\big[i_1+b_1,\ i_2+b_2,\ \dots,\ i_n+b_n\big]$ be partitioned across $H$ hosts.
Since each pixel belongs to exactly one host, we can write $\sum_{h=1}^{H} \mathbb{I}(\mathbf{i+b} \in [x]_h) = 1$ and multiply with $I[\mathbf{i + b}]$ to get:

\begin{align}
    I &= \sum_{b\in\{0,1\}^n}
\left(\prod_{k=1}^n (1-\alpha_k)^{1-b_k}\,\alpha_k^{b_k}\right)
\;\left(I\big[\mathbf{i + b}] * \left(\sum_{h=1}^{H} \mathbb{I}(\mathbf{i+b} \in [x]_h)\right)\right) \\
&= \sum_{h=1}^{H} \sum_{b\in\{0,1\}^n}
\left(\prod_{k=1}^n (1-\alpha_k)^{1-b_k}\,\alpha_k^{b_k}\right)
\;\left(I\big[\mathbf{i + b}] * \mathbb{I}(\mathbf{i+b} \in [x]_h)\right)  \label{eq:interp-h} \\
&= \sum_{h=1}^{H} I_h
\end{align}
where 
\begin{align}
    I_h &= \sum_{b\in\{0,1\}^n} \left(\prod_{k=1}^n (1-\alpha_k)^{1-b_k}\,\alpha_k^{b_k}\right) \;I\big[\mathbf{i + b}] * \mathbb{I}(\mathbf{i + b}\in \left[x\right]_h) \\
    &= \sum_{b\in\{0,1\}^n} \left(\prod_{k=1}^n (1-\alpha_k)^{1-b_k}\,\alpha_k^{b_k}\right) \;J_h\big[\mathbf{i + b}] \label{eq:interp-honly}
\end{align}
where $J_h\big[\mathbf{x}] = I[\mathbf{x}]$ if $\mathbf{x} \in \gridx_h$ else 0.
Image $J_h$ is therefore \textit{identical} to the sharded image $I$ on host $h$.
\cref{eq:interp-honly} refers to performing trilinear interpolation on the shard $I_h$ (with zero padding) since the sum is only over coordinates that reside in $[x]_h$.
This means the warped image in \cref{eq:interp} can be obtained by performing interpolation over the shards individually and adding the warped images together.
This is illustrated in \cref{fig:ring-sampler}(c).
Coordinates residing between multiple shards will accumulate partial sums from each sharded image, and no additional consideration is needed for boundary conditions. 
% This algorithm will fetching an image shard from  GPUs, compute the partial warped shard and add it to the result.  
The communication protocol in this algorithm is similar to Ring Attention \citep{ringattention}, where image shards are passed across hosts, and partial results are accumulated into the final result. 
Our algorithm requires a memory overhead of only $N/H$ to store the sharded image from host $j \ne i$.
Our pseudocode is provided in \cref{alg:ringsampler}.

\input{algs/ringsampler}

\subsection{Implementation Considerations}

\paragraph{Rescaling the warp function to sample sharded images}
\label{sec:ringsampler-rescale}
Interpolating from sharded images requires one additional consideration. 
% When an image is interpolated using grid sampling operator, it is assumed that the bounds of the image are $x_{\min}, x_{\max} = \{-1\}^d, \{1\}^d$.
% However, the sharded image $J_h$ has bounds $x_{\min}^h, x_{\max}^h$ stored as part of its metadata.
% Therefore, when we compute the partial sum $I_h = J_h(\varphi)$, we must rescale $\varphi$ such that the scaled transform corresponds to coordinates in $J_h$ and not $I$.
The grid sampler interpolates an image $I$ defined on $\Omega$ using warp coordinates $\gridu_\Omega$ defined on $\Omega$.
However, the sharded image $J_h$ is defined on the domain $\oh$, and therefore any warped coordinate $\varphi(x) \in \Omega$ must be rescaled to the corresponding coordinates in $\varphi_h(x) \in \oh$.
From the implementation standpoint, the leftmost coordinate of $J_h$ is $x_{\min}^h$ when the entire image $I$ is passed to \texttt{grid\_sampler}.
However, when $J_h$ is provided as input to \texttt{grid\_sampler}, the leftmost pixel of $J_h$ is located at $[-1, -1, \dots, -1]$ according to PyTorch convention.
Since our optimization variables $A, t, \gridu$ refer to locations on $\Omega$, and not $\oh$, we need to rescale these variables appropriately when sampling from $J_h$.
% 
% Therefore, we need to rescale the warp function to sample from $J_h$ instead of $I$ correctly.

% The rescaling corresponds 
% 
% This corresponds 
The rescaling corresponds to a diagonal scaling matrix $S_h$ and translation $t_h$ such that $S_h x_{\min}^h + t_h = x_{\min}^\Omega$ and $S_h x_{\max}^h + t_h = x_{\max}^\Omega$.
The resampled warp function to sample from $J_h$ becomes $\varphi_h(x) = S_h(Ax + t + u(x)) + t_h = (A'_hx + t'_h) + S_hu(x)$.
where 
% The first two terms correspond to the affine matrix 
$A'_h, t'_h = S_hA, (S_ht + t_h)$.
Therefore, we must sample $J_h$ using the transform $A'_h \gridx_\oh + t_h' + S_h \gridu_\oh$.
In the vanilla grid sampler implementation, the intermediate grid $S_h\gridu_\oh$ and its gradient consume another $6N/H$ memory.
Combined with the $N/H$ overhead for storing the received image shard, we add a total of $7N/H$ memory overhead, which is less than $N$ for $H \ge 8$, making the algorithm impractical for fewer GPUs (say $H = 4$). 

To prevent this $6N/H$ additional overhead, we extend the generalized grid sampler as mentioned \cref{sec:gridsampler} to sample from a transform of the form $A[x] + t + S[u]$ directly. 
This computes the value $Su(x)$ directly inside the CUDA kernel, and the backward pass also computes and accumulates the gradient w.r.t. $u(x)$ directly, avoiding the $6N/H$ overhead.

\begin{figure}[h!]
    \centering
    \includegraphics[width=0.8\textwidth]{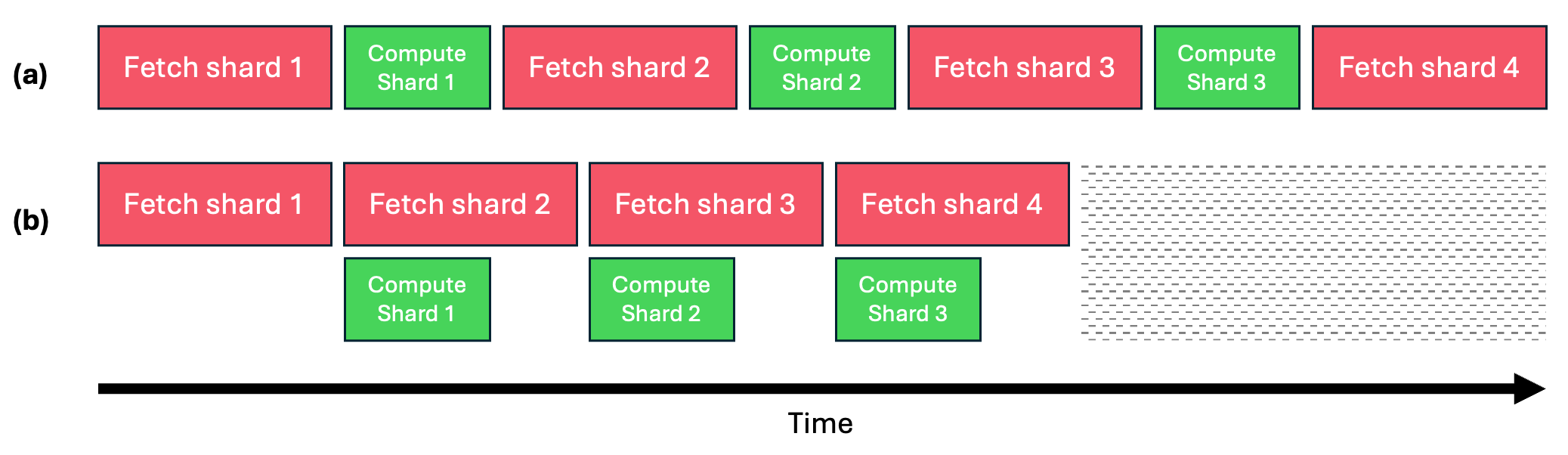}
    \caption{Interleaved communication (\textcolor{red}{red}) and computation (\textcolor{ForestGreen}{green}) in the ring sampler. {\textcolor{gray}{gray}} denotes time saved by interleaving communication and computation.}
    \label{fig:ringsampler-scheduling}
\end{figure}

\paragraph{Interleaved communication}
An important implementation detail is the interleaving of communication and computation in the ring sampler.
While we compute the partial moved image aggregate, the next image shard can be fetched asynchronously in the background.
This is illustrated in \cref{fig:ringsampler-scheduling}.

\subsection{Alternative Designs}

A naive approach can be to route the coordinate $\varphi(x_i) \in \gridx_j$ to GPU $j$ and retrieve the image coordinate, similar to routing tokens using expert parallelism (EP) used for Mixture-of-Experts (MoEs) \citep{moe2,moe1}.
However, this approach has two major drawbacks in our setting. 
First, due to the deformable nature of $\varphi$, the partitioning of coordinates across hosts is generally uneven. 
In the worst case, a single GPU can receive all $3N$ coordinates leading to an indirect \texttt{allgather} operation resulting in OOMs or uneven GPU utilization across hosts. 
Second, coordinates that point to regions between two multiple image boundaries need to be sent to variable number of hosts, which is non-trivial to implement.
These two factors make both the forward and backward pass implementations cumbersome.
Inspired by \citep{ringattention}, we propose a distributed ring sampler that decomposes the computation into partial sums, leading to a simple implementation without degraded scaling performance \cref{fig:scaling}.

%% file: algs/ringsampler.tex
\begin{algorithm}[t]

\caption{Ring Sampler Implementation}
\label{alg:ringsampler}
\begin{algorithmic}[1]

\Require $M_j$ (moving image shard), $\gridu_j$ (warp field shard), $(A, t)$ (affine transform)
\Function{Forward}{$M_j$, $\gridu_j$, $(A, t)$}
    \State Define $\texttt{moved}_j = 0$
    \For{$h = 1$ {\bf to} $H$}
         \State $J_h \leftarrow \texttt{send\_and\_recv}(M_j, h)$ \hfill $\triangleright$ Send and receive the image shard from offset $h$
         \State Compute diagonal $S_h, t_h$ such that $S_h x_{\min}^h + t_h = x_{\min}^\Omega$ and $S_h x_{\max}^h + t_h = x_{\max}^\Omega$
         \State Rescale affine transform $A_h \leftarrow S_h A, t_h \leftarrow S_h t + t_h$
         \State $\texttt{moved}_j \leftarrow \texttt{moved}_j + \texttt{grid\_sampler}(J_h; A_h, t_h, S_h, \gridu_j)$ \hfill $\triangleright$ Avoid computing $S_h \gridu_j$ explicitly
    \EndFor
    \State \Return $\texttt{moved}_j$
\EndFunction \\ 

\Function{Backward}{$g = \pder{\mathcal{L}}{\texttt{moved}_j}$, $\texttt{moved}_j$, $M_j$, $\gridu_j$, $(A, t)$}
\State Define ${g}_{\gridu_j} = 0, {g}_A = 0, {g}_t = 0, {g}_{M_j} = 0$
\For{$h = 1$ {\bf to} $H$}
    \State $J_h \leftarrow \texttt{send\_and\_recv}(M_j, h)$ \hfill $\triangleright$ Send and receive the image shard from offset $h$
    \State Compute diagonal $S_h, t_h$ such that $S_h x_{\min}^h + t_h = x_{\min}^\Omega$ and $S_h x_{\max}^h + t_h = x_{\max}^\Omega$
    \State Rescale affine transform $A_h \leftarrow S_h A, t_h \leftarrow S_h t + t_h$
    \If{$\texttt{requires\_grad}(M_j)$}
        \State $g_{inp} \leftarrow \texttt{zeros\_like}(M_j)$
    \Else
        \State $g_{inp} \leftarrow \texttt{None}$
    \EndIf
    \State Compute \texttt{backward\_grid\_sampler}($g, J_h, A_h, t_h, S_h, \gridu_j, g_{\gridu_j}, g_A, g_t, g_{inp}$)
    \If{$\texttt{requires\_grad}(M_j)$}
        \State $g'_{M_j} = \texttt{send\_and\_recv}(g_{inp}, -h)$
        \State $g_{M_j} \leftarrow g_{M_j} + g'_{M_j}$
    \EndIf
\EndFor
\State \Return ${g}_{\gridu_j}, {g}_A, {g}_t, {g}_{M_j}$
\EndFunction

\end{algorithmic}
\end{algorithm}

%% file: tables/fireants-speedup.tex
\begin{table}[H]
\centering
\caption{\textbf{Extended Results on accelerated registration on FireANTs}: Accelerating FireANTs registration with various computation backends and registration algorithms (Greedy and SyN). Our implementations maintain accuracy while substantially reducing runtime and peak memory usage.
\label{tab:fireants-speedup}
\colorbox{green!25}{\strut\ }(Green)/\colorbox{yellow!25}{\strut\ }(Yellow) = best/second;
Speedup and memory reduction are computed with respect to our kernels.
Our fused kernels maintain accuracy while substantially reducing runtime and peak memory usage.}
\setlength{\tabcolsep}{6pt}
\renewcommand{\arraystretch}{1.1}
\resizebox{0.98\linewidth}{!}{%
\begin{tabular}{lllccccc}
    \hline\hline
    \textbf{Algorithm} & \textbf{Method} & \textbf{Backend} & \textbf{Dice Score} $\uparrow$ & \textbf{Runtime (s)} $\downarrow$ & \textbf{Memory (MB)} $\downarrow$ & \textbf{Speedup} $\uparrow$ & \textbf{Mem. Reduction (\%)} $\uparrow$ \\
    \hline
    \multirow{5}{*}{Greedy} & LNCC & VXM/TM & \second{76.96 $\pm$ 3.60} & {57.08 $\pm$ 2.45} & {1418.5 $\pm$ 0.0} & 113.47 & 59.29 \\
    & LNCC & FastLNCC & \second{76.96 $\pm$ 3.60} & 3.76 $\pm$ 0.16 & 1026.3 $\pm$ 0.0 & 7.48 & 43.73 \\
    & LNCC & FireANTs & 72.81 $\pm$ 3.87 & 1.44 $\pm$ 0.08 & 1044.5 $\pm$ 0.0 & 2.87 & 44.71 \\
    & LNCC & \texttt{torch.compile} & 69.35 $\pm$ 4.09 & \second{0.82 $\pm$ 0.04} & \second{860.7 $\pm$ 0.0} & 1.63 & 32.90 \\
    & LNCC & Ours & \best{78.67 $\pm$ 3.04} & \best{0.50 $\pm$ 0.01} & \best{577.5 $\pm$ 0.0} & 1.00 & 0.00 \\
    \hline
    \multirow{4}{*}{Greedy} & MI & PyTorch & \second{75.88 $\pm$ 3.45} & 7.51 $\pm$ 0.37 & 12206.3 $\pm$ 0.0 & 2.59 & 95.27 \\
    & MI & \texttt{torch.compile} & \second{75.88 $\pm$ 3.45} & \best{1.05 $\pm$ 0.05} & 3865.5 $\pm$ 0.0 & 0.36 & 85.06 \\
    & MI & Ours & 75.87 $\pm$ 3.44 & \second{2.90 $\pm$ 0.16} & \best{577.5 $\pm$ 0.0} & 1.00 & 0.00 \\
    & MI & Ours + \texttt{torch.compile} & \best{75.93 $\pm$ 3.47} & 2.95 $\pm$ 0.16 & \second{657.3 $\pm$ 0.0} & 1.02 & 12.13 \\
    \hline \hline
    \multirow{5}{*}{SyN} & LNCC & VXM/TM & 76.69 $\pm$ 2.88 & 63.57 $\pm$ 0.58 & 1892.0 $\pm$ 0.0 & 65.92 & 50.05 \\
    & LNCC & FastLNCC & \second{76.70 $\pm$ 2.88} & 4.27 $\pm$ 0.05 & 1486.7 $\pm$ 0.0 & 4.43 & 36.43 \\
    & LNCC & FireANTs & 74.70 $\pm$ 2.93 & 2.55 $\pm$ 0.10 & 1616.4 $\pm$ 0.0 & 2.65 & 41.54 \\
    & LNCC & \texttt{torch.compile} & 71.65 $\pm$ 3.41 & \second{1.46 $\pm$ 0.04} & \second{1472.0 $\pm$ 0.0} & 1.51 & 35.80 \\
    & LNCC & Ours & \best{78.79 $\pm$ 2.82} & \best{0.96 $\pm$ 0.08} & \best{945.0 $\pm$ 0.0} & 1.00 & 0.00 \\
    \hline
    \multirow{4}{*}{SyN} & MI & PyTorch & 76.74 $\pm$ 2.58 & 12.84 $\pm$ 0.66 & 17720.8 $\pm$ 0.0 & 2.96 & 94.67 \\
    & MI & \texttt{torch.compile} & 76.76 $\pm$ 2.58 & \best{2.40 $\pm$ 0.13} & 7758.9 $\pm$ 0.0 & 0.55 & 87.82 \\
    & MI & Ours & 76.86 $\pm$ 2.59 & \second{4.34 $\pm$ 0.28} & \best{945.0 $\pm$ 0.0} & 1.00 & 0.00 \\
    & MI & Ours + \texttt{torch.compile} & \best{77.00 $\pm$ 2.57} & 4.56 $\pm$ 0.24 & \second{1104.5 $\pm$ 0.0} & 1.05 & 14.44 \\
    \hline
\end{tabular}
}
\end{table}

%% file: tables/faux-oasis-efficiency.tex
\begin{table}[H]
  \centering
  \caption{\textbf{Extended Efficiency Results on faux-OASIS-dataset}: Comparison of registration methods across multiple resolutions. Reported metrics include average Dice similarity coefficient (higher is better), wall-clock runtime, GPU cost (measured in GB-hours), relative speedup, and GPU cost reduction with respect to {FireANTs + \methodname (Ours)}. GPU usage (e.g., single GPU, multi-GPU, or CPU) is annotated alongside the cost values.}
  \label{tab:exp1_allmethods_grouped_runtime}
  \renewcommand{\arraystretch}{1.15}
  \resizebox{\linewidth}{!}{%
  \begin{tabular}{l
                  l
                  S[table-format=1.3(2)]
                  S[table-format=3.3]
                  S[table-format=4.3]
                  S[table-format=3.2,detect-weight=true]
                  S[table-format=3.2,detect-weight=true]}
    \toprule
    \textbf{Resolution} & \textbf{Method} &
    {\makecell{\textbf{Avg Dice Score} $\uparrow$}} &
    {\makecell{\textbf{Wall Clock} $\downarrow$ \\ \textbf{($10^{-2}$ Hours)}}} &
    {\makecell{\textbf{GPU Cost} $\downarrow$ \\ \textbf{($10^{-2}$ GB-Hours)}}} &
    {\makecell{\textbf{Speedup}}} &
    {\makecell{\textbf{GPU Cost} \\ \textbf{Reduction} (\%)}} \\
    \midrule
\multirow{9}{*}{\SI{1}{\milli\meter}}      & TransMorph             & \best{\num{0.851 +- 0.016}} & \best{\num{0.015}}  & \num{0.262}\gpu{1}  & 0.56$\times$ & \num{87.81} \\
      & VFA                    & \second{\num{0.851 +- 0.023}} & \second{\num{0.017}}  & \second{\num{0.216}}\gpu{1}  & 0.63$\times$ & \num{85.18} \\
      & Ours               & \num{0.838 +- 0.028} & \num{0.027}  & \best{\num{0.032}}\gpu{1}  & 1.00$\times$ & \num{0.00} \\
      & UniGradICON-IO             & \num{0.826 +- 0.022} & \num{5.167}  & \num{58.498}\gpu{1}  & 194.07$\times$ & \num{99.95} \\
      & UniGradICON-noIO           & \num{0.815 +- 0.026} & \num{0.067}  & \num{0.238}\gpu{1}  & 2.50$\times$ & \num{86.55} \\
      & SynthMorph             & \num{0.801 +- 0.022} & \num{2.155}  & \num{99.061}\gpu{1}  & 80.93$\times$ & \num{99.97} \\
      & Anatomix               & \num{0.796 +- 0.035} & \num{0.379}  & \num{2.656}\gpu{1}  & 14.24$\times$ & \num{98.80} \\
      & CLAIRE          & \num{0.776 +- 0.044} & \num{0.518}  & \num{1.389}\gpu{1}  & 19.47$\times$ & \num{97.70} \\
      & ITK-dreg     & \num{0.662 +- 0.055} & \num{1.527}  & \num{1.017}\gpu{CPU}  & 57.37$\times$ & \text{--} \\
    \midrule
    \multirow{9}{*}{\SI{500}{\micro\meter}}      & Ours               & \best{\num{0.872 +- 0.028}} & \best{\num{0.109}}  & \best{\num{0.862}}\gpu{1}  & 1.00$\times$ & \num{0.00} \\
      & VFA                    & \second{\num{0.805 +- 0.044}} & \num{0.302}  & \num{3.896}\gpu{1}  & 2.78$\times$ & \num{77.87} \\
      & CLAIRE          & \num{0.779 +- 0.051} & \num{25.903}  & \num{396.169}\gpu{1}  & 238.04$\times$ & \num{99.78} \\
      & SynthMorph             & \num{0.771 +- 0.035} & \num{4.068}  & \num{187.049}\gpu{1}  & 37.39$\times$ & \num{99.54} \\
      & TransMorph             & \num{0.759 +- 0.028} & \second{\num{0.198}}  & \second{\num{3.501}}\gpu{1}  & 1.82$\times$ & \num{75.38} \\
      & Anatomix               & \num{0.758 +- 0.040} & \num{8.837}  & \num{310.818}\gpu{1}  & 81.21$\times$ & \num{99.72} \\
      & ITK-dreg     & \num{0.699 +- 0.056} & \num{41.259}  & \num{207.466}\gpu{CPU}  & 379.17$\times$ & \text{--} \\
      & UniGradICON-IO             & \num{0.615 +- 0.047} & \num{84.538}  & \num{1072.657}\gpu{1}  & 776.89$\times$ & \num{99.92} \\
      & UniGradICON           & \num{0.610 +- 0.044} & \num{0.842}  & \num{3.545}\gpu{1}  & 7.73$\times$ & \num{75.69} \\
    \midrule
    \multirow{7}{*}{\SI{250}{\micro\meter}}
      & Ours              & \best{\num{0.895 +- 0.029}} & \best{1.065}  & \second{47.059}\gpu{1}  & 1.00$\times$ & 0.00 \\
      & CLAIRE          & \second{\num{0.809 +- 0.054}} & \num{1207.536}  & \num{159046.981}\gpu{4}  & 1133.84$\times$ & \num{99.97} \\
      & VFA                   & \num{0.714 +- 0.066} & 3.872  & 49.939\gpu{1}  & 3.64$\times$ & 5.77 \\
      & SynthMorph            & \num{0.690 +- 0.052} & 32.808 & 1507.133\gpu{1} & 30.80$\times$ & 96.88 \\
      & TransMorph            & \num{0.689 +- 0.044} &\second{2.597}  & \best{45.965}\gpu{1}  & 2.44$\times$ & -2.38 \\
      & Anatomix              & \num{0.620 +- 0.031} & 88.480 & 3112.015\gpu{1} & 83.07$\times$ & 98.49 \\
      & UniGradICON-IO            & \num{0.398 +- 0.062} & 163.812& 2539.721\gpu{1} & 153.80$\times$ & 98.15 \\
      & UniGradICON          & \num{0.359 +- 0.044} & 7.811  & 55.057\gpu{1}  & \num{7.33}$\times$ & 14.53 \\
      & ITK-dreg              & \num{0.758 +- 0.046} & 1363.868  & \num{33065.677}\gpu{CPU}  & 1280.63$\times$ & \text{--} \\
    \bottomrule
  \end{tabular}
  }
\end{table}

%% file: sections/app/faux-oasis.tex
\section{Additional Details on the simulated ex-vivo brain MRI dataset}
\label{app:faux-oasis}

In this section, we provide additional details on the synthetic data generation pipeline for the faux-OASIS dataset, followed by baseline configurations, and finally compare performance-efficiency tradeoffs and show qualitative results.    

% \subsection{Faux Oasis: Synthetic Data Generation Pipeline}
\subsection{Synthetic Data Generation Pipeline}

\begin{figure}[t!]
% \begin{wrapfigure}{r}{0.5\textwidth}
  \centering
  \vspace{-10pt}
  \includegraphics[width=0.9\textwidth]{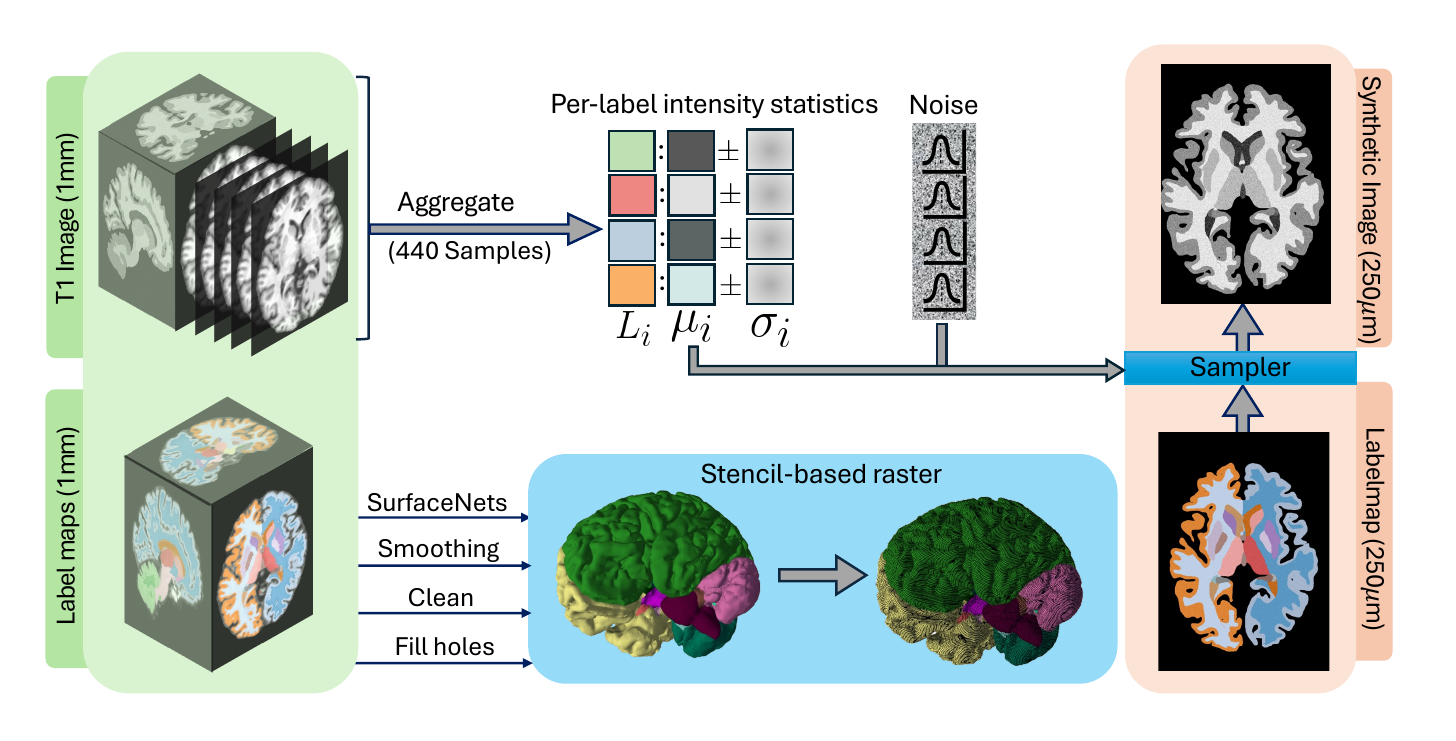}
\caption{\textbf{Synthetic data generation pipeline for faux-OASIS}. Coarse anatomical labels undergo geometry-preserving upsampling via surface reconstruction, followed by statistical intensity painting to produce high-resolution MR images at \SI{0.25}{mm}.}
  \label{fig:pipeline}
  \vspace{-10pt}
% \end{wrapfigure}
\end{figure}

To emulate high resolution (250$\um$ isotropic) T1 weighted images, we use the standard OASIS validation dataset to generate synthetic images.
Our method is inspired by \cite{synthseg,anatomix} to use the labelmaps as a starting point and synthesize images that are faithful to the labelmaps.
The synthetic data generation pipeline is illustrated in \cref{fig:pipeline}.
Specifically, our pipeline has three stages:
\begin{enumerate}
  \item \textbf{Compute per-label intensity statistics}: For each label, we consider all the intensities in the voxels belonging to the label. We store mean and standard deviation of the intensities for each label computed over the entire OASIS validation set.
  \item \textbf{Geometry-preserving upsampling of labels}: We use the labelmaps at 1mm isotropic and perform surface-based upsampling to resample the labelmaps with subvoxel accuracy \citep{pyvista}. 
  \item \textbf{Intensity painting}: We use the per-label intensity statistics and the voxelized labelmaps at  250$\um$ isotropic to synthesize the images.
\end{enumerate}
Following the generation of $250\um$ images, we downsample the images to $500\um$ and $1mm$ isotropic to show the effect on performance with downsampled images.

% We synthesize high–resolution MR images from coarse anatomical segmentations in two steps:  
% (i) \emph{geometry-preserving} super-resolution of the label map via surface extraction and voxelization, and  
% (ii) \emph{label-wise intensity painting} from cohort statistics. This yields paired image and synthetic label maps $(I_{\text{syn}}, L_\uparrow)$ at \SI{0.25}{mm} isotropic resolution. We downsample the synthetic label maps to \SI{0.5}{mm} and \SI{1}{mm} isotropic to show the effect on performance with downsampled images.
We describe the pipeline in detail below.

\paragraph{Per-label intensity statistics.}
Contrary to other synthetic data generation pipelines\cite{anatomix,synthseg}, we do not want to generate randomized intensities for each image and want to simulate the T1-weighted images.
Towards this end, we compute the per-label intensity statistics for all images in the OASIS validation set.

\paragraph{Geometry-preserving upsampling of labelmaps.}
Given an image volume and labelmap pair $(I,L)$, we upsample the labelmap to 250$\um$ isotropic $L_{\uparrow}$.
However, naively upsampling the label voxel grid and thresholding typically causes blocky artifacts \citep{surfacenets,marchingcubes,kitwarefastiso}, which has led to many sophisticated subvoxel-accurate surface reconstruction algorithms.
We use PyVista's SurfaceNets algorithm \citep{surfacenets} to extract surface contours from 3D image label maps. 
Specifically, an \texttt{ImageData} object with labels is converted into \emph{cell data} using \texttt{contour\_labels} (VTK SurfaceNets) to obtain per-label surfaces $S_\ell$ that respect voxel geometry and avoid block artifacts with voxel based interpolation.
The generated surface is smoothed using a constrained Taubin/Windowed-Sinc smoothing with conservative iterations (typically 16-30, relaxation $\approx 0.5$), then use \texttt{clean} and \texttt{fill\_holes} to remove slivers and pinholes while preserving anatomical shape fidelity.
The surface $S_\ell$ is voxelized to obtain a binary mask $M_\ell$, then the labelmap $L_\uparrow$ is assembled as
\[
L_\uparrow(\mathbf{p}) \;=\;
\begin{cases}
\ell & \text{if } M_\ell(\mathbf{p})=1 \text{ for some }\ell\ge1,\\[2pt]
0 & \text{otherwise}.
\end{cases}
\]
Finally, image-stencil-based rasterization (\texttt{voxelize\_binary\_mask}) is performed into the target \texttt{ImageData} at $t=\SI{0.25}{mm}$.% The generated surface is smoothed using a constrained smoothing filter, we run the smoothing for 16 iterations with a relaxation factor of 0.5.
When surfaces overlap, later labels in the loop take precedence; we process labels in anatomical priority order to ensure critical structures are preserved.
All steps for labelmap upsampling are implemented with PyVista/VTK for robustness and reproducibility.
% \begin{itemize}[leftmargin=1.2em,itemsep=2pt,topsep=2pt]
%   \item \textbf{Surface extraction:} 
%   \item \textbf{Smoothing \& cleanup:} 
%   \item \textbf{Voxelization:} 
% \end{itemize}

\paragraph{Synthesizing the image.}
For each label, we fill the voxels with intensities sampled from a normal distribution with the mean and standard deviation of the intensities corresponding to the label.

% Interpolating labels causes aliasing; instead we reconstruct a smooth, watertight surface per label and re-voxelize on the high-res grid $\Omega_t$. For each $\ell\in\{1,\dots,K\}$ we extract a surface $S_\ell$ (\texttt{contour\_labels} on cell data; SurfaceNets-style extraction), apply light Taubin/Windowed-Sinc smoothing (typically 16–30 iterations, relaxation $\approx0.5$), and fill small holes. We voxelize $S_\ell$ on $\Omega_t$ to obtain a binary mask $M_\ell$, then assemble

% \paragraph{Inputs.}
% Let $L \in \{0,\dots,K\}^{N_x \times N_y \times N_z}$ be a coarse segmentation with voxel spacings $\mathbf{s}=(s_x,s_y,s_z)$ (mm) and affine $A\!\in\!\mathbb{R}^{4\times4}$. 
% Per-label intensity statistics are given by $\{(\mu_\ell,\sigma_\ell)\}_{\ell=1}^K$, estimated once from a template cohort.

% \paragraph{Target grid and affine.}
% We construct an isotropic lattice with spacing $t=\SI{0.25}{mm}$ and extents
% \[
% \tilde N_i \;=\; \Big\lfloor (N_i-1)\,\frac{s_i}{t} \Big\rfloor + 1,
% \qquad i\in\{x,y,z\},
% \]
% preserving the physical field of view. The output affine $\tilde A$ inherits the origin and orientation of $A$, while rescaling its axis columns:
% \[
% \tilde A_{:,i} \;\leftarrow\; A_{:,i}\cdot \frac{t}{s_i},
% \qquad i\in\{1,2,3\}.
% \]

% 
% \paragraph{Intensity painting.}
Given $\{(\mu_\ell,\sigma_\ell)\}$, we synthesize the image by i.i.d.\ draws within each region:
\[
I_{\text{syn}}(\mathbf{p})
\;\sim\;
\mathcal{N}\!\big(\mu_{L_\uparrow(\mathbf{p})},\,\sigma_{L_\uparrow(\mathbf{p})}^{2}\big),
\qquad \mathbf{p}\in\Omega_t.
\]
We follow this step with a Gaussian smoothing with $\sigma=0.75$ voxels to impart local coherence without washing out label edges. Background ($L_\uparrow{=}0$) is set to zero.

\begin{algorithm}[t]
\caption{High-Resolution MR Synthesis Pipeline}
\label{alg:intensity-painting}
\begin{algorithmic}[1]
\Require $L$, spacings $\mathbf{s}$, affine $A$, stats $\{(\mu_\ell,\sigma_\ell)\}_{\ell=1}^K$, target spacing $t$
\State Compute $\tilde N_x,\tilde N_y,\tilde N_z$ and $\tilde A$ as above
\State $L_\uparrow \gets 0$ on $\Omega_t$
\For{$\ell = 1$ {\bf to} $K$}
    \State $S_\ell \gets \textsc{SurfaceNets}(L{=}\ell)$; smooth \& fill holes
    \State $M_\ell \gets \textsc{Voxelize}(S_\ell, \Omega_t)$
    \State $L_\uparrow[\text{where } M_\ell{=}1] \gets \ell$ \Comment{Assign label to voxelized region}
\EndFor
\State $I_{\text{syn}} \gets 0$
\For{$\ell = 1$ {\bf to} $K$}
    \State $U \gets \{\mathbf{p} \mid L_\uparrow(\mathbf{p})=\ell\}$
    \State $I_{\text{syn}}[U] \gets \textsc{Normal}(\mu_\ell,\sigma_\ell^2)$ \Comment{IID draws}
\EndFor
\State $I_{\text{syn}} \gets \textsc{GaussianBlur}(I_{\text{syn}},\,\sigma{=}0.75)$
\State \Return $(I_{\text{syn}}, L_\uparrow, \tilde A)$
\end{algorithmic}
\end{algorithm}

\paragraph{Randomization and metadata.}
All stochastic draws are seeded per subject (seed $=$ \texttt{base\_seed} $+$ \texttt{subject\_id}) for exact reproducibility.\footnote{We use \texttt{base\_seed} = 2025 in our experiments.}
All outputs are written as NIfTI files with same origin and directions as the original images, but with a voxel spacing of $t=\SI{0.25}{mm}$.

\subsection{Baselines}
We augment FireANTs \citep{fireants} with {\methodname} to enable scalable image registration at high resolutions.
The methods and their hyperparameter settings are described below:
\begin{itemize}
\item \textbf{CLAIRE}\citep{claire}:
CLAIRE is a velocity-based diffeomorphic registration framework optimized for distributed GPU/CPU execution via MPI. We use the official repository inside a custom multi-GPU Docker image that adds CUDA-aware Open MPI (v4.0.3; CUDA~11), since the official container supports only single-GPU runs. We launch one MPI rank per GPU and bind each rank to a distinct device via a lightweight wrapper that maps \texttt{OMPI\_COMM\_WORLD\_RANK} to \texttt{CUDA\_VISIBLE\_DEVICES}, enabling data-parallel execution across $N$ GPUs. We keep default solver settings, request deformation maps (\texttt{-defmap}), and set the continuation parameter \texttt{-betacont 7.75e-04} following the official examples; all other hyperparameters use documented defaults, including the iteration cap (\texttt{-maxit 50}). Full-resolution runs use 4 GPUs (4 ranks), while half/quarter resolutions use a single GPU (1 rank).

\item \textbf{ITK-DReg}\citep{itkdreg}:
ITK-DReg is a CPU-based, distributed, out-of-memory registration framework built on ITK and \texttt{dask.distributed}, formulating registration as blockwise map–reduce. We use the \texttt{itk\_dreg} pipeline with Elastix in deformable-only B-spline mode: the metric is AdvancedNormalizedCorrelation with three pyramid levels (\texttt{NumberOfResolutions=3}, \texttt{GridSpacingSchedule=[4,2,1]}), optimized via AdaptiveStochasticGradientDescent with \texttt{MaximumNumberOfIterations=500}. We use random sampling with \texttt{NumberOfSpatialSamples=5000} (refreshed each iteration). Registration operates in voxel units with \texttt{FinalGridSpacingInVoxels=20} and \texttt{BSplineTransformSplineOrder=3}. To scale to high resolutions, the fixed image is tiled into $256^3$-voxel chunks with $25\%$ overlap per axis; per-block results are reduced to a global displacement field defined on a grid subsampled by a factor of $4$. ITK threading is set via \texttt{SetGlobalDefaultNumberOfThreads}=24 (reported \texttt{GetGlobalMaximumNumberOfThreads}=128).

\item \textbf{FireANTs + {\methodname} (Ours)}\citep{fireants}: We use the official repository and scripts, except for our proposed modules (grid sampler, LNCC, and Mutual Information). We perform registration using the multi-scale of $4mm, 2mm, 1mm, 500\um$, and $250\um$ for $200, 200, 200, 100, 25$ iterations. 
We also truncate the optimization at $1mm$ and $500\um$ resolutions to verify the performance of the method at downsampled resolutions.
We use our Fused LNCC implementation with a window size of $7$, and a learning rate of 0.5.
The smoothing kernels are chosen with a $\sigma_{warp} = 0.5$ pixels, and $\sigma_{grad} = 1.0$ pixels.
\end{itemize}

We also evaluate against state-of-the-art deep learning methods:
\begin{itemize}
\item \textbf{SynthMorph}\citep{synthmorph}: 
SynthMorph uses an acquisition-free synthetic data generation pipeline to train a registration network.
We use the default \texttt{mri\_synthmorph} script provided by the vendor. Since all images are affine-aligned, we use the deformable registration mode \texttt{-m deform} with a regularization weight of \texttt{-r 0.25}.

\item \textbf{Vector-Field Attention} \citep{vfa}:
Vector-Field Attention (VFA) is a weakly-supervised learning-based method utilizing a novel attention  module to retrieve per-pixel correspondence based on feature similarity.
We evaluate using the pretrained model (trained on OASIS data with weak label supervision) provided in the official repository.

\item \textbf{UnigradICON} \citep{unigradicon}:
UnigradICON is a foundational registration model by training on a composite dataset consisting of lung CT, knee MRI, Abdomen CT, brain MRI, totalling more than 3 million image pairs, of which 4000 image pairs are sampled per epoch to mitigate data imbalance.
The model is trained with a bidirectional similarity loss and an inverse consistency loss.
UnigradICON also provides an instance optimization based postprocessing step to improve the registration performance.
We use the pretrained model and scripts provided in the official repository, and compare performance with and without the instance optimization step.

\item \textbf{TransMorph} \citep{transmorph}:
TransMorph is one of the first successful application of transformer-based architectures for image registration, marking a departure from  traditional convolutional architectures.
Compared to other convolutional architectures, TransMorph demonstrates higher performance under domain shift \cite{jian2024mamba,dio,magicormirage} among the deep learning methods.
We use the pretrained model (\texttt{TransMorph-Large} trained on the OASIS dataset) that is provided in the official repository.

\item \textbf{Anatomix + ConvexAdam} \citep{anatomix}:
Anatomix is a feature extractor that is trained to anticipate strong domain shift at training time and uses contrastive learning to extract domain-agnostic features that mitigate the effect of nuisance factors.
Anatomix shows strong results on zero-shot registration on abdomen and myocardium.
We use the pretrained model and scripts provided in the official repository.
\end{itemize}

We also acknowledge Quicksilver \citep{quicksilver} as a relevant baseline that performs patch-based registration. 
However, despite our best efforts with containerizing the environment (the dependencies are no longer available or supported on modern hardware), we were unable to run this baseline on our system. 

All deep learning methods are tested on $1mm, 500\um$, and $250\um$ resolutions.
On $500\um$ and $250\um$ resolutions, all methods run out of memory on a single NVIDIA A6000 GPU, and the methods do not provide infrastructure to run on multiple GPUs.
We adopt the patch-based registration strategy adopted by the literature on high-resolution registration methods for histology \citep{deeperhistreg,patchbasedhistoreg,patchbasedhisto3} as additional baselines with the above deep learning models as registration backends.
We choose \citep{synthmorph,anatomix,unigradicon} as general-purpose deep learning methods to mitigate the effect of domain shift due to patch-based registration at higher resolutions, and \citep{vfa,transmorph} as methods that are trained with weak label supervision on the OASIS dataset to verify performance at 1mm resolution and observe the performance at higher resolutions.

\paragraph{Robust HD90 (Cumulative)}
\label{subsec:hd90}
Hausdorff distance is a widely adopted boundary-based metric in medical image registration.
The conventional definition of HD90 (the 90th percentile Hausdorff distance) simply reports the 90th percentile, but does not provide an average performance for all surface boundaries.
% reports the distance value below which 90\% of the sorted surface-to-surface distances fall.
In contrast, we employ a modified formulation, which we denote as \emph{cumulative HD90}, designed to provide a more stable and comprehensive estimate.
Specifically, rather than selecting the single distance value at the 90th percentile, we compute the mean of all surface distances up to the 90th percentile.
Formally, given sorted distances $\{d_i\}_{i=1}^N$, we compute
\[
\mathrm{HD}_{90}^{\text{cu}} = \frac{1}{k}\sum_{i=1}^{k} d_i,
\quad k = \lfloor 0.9 \, N \rfloor .
\]
Distances are computed bidirectionally between ground-truth and predicted surfaces using isotropic voxel spacing, and the final HD90 is defined as the maximum of the two directional estimates.
% This definition yields more stable and consistent results across runs while preserving the standard interpretation of HD90 as a robust boundary-distance measure.

\subsection{Additional Results and Discussion}
\cref{tab:exp1_allmethods_grouped_runtime} shows the performance comparison for all methods at different resolutions.
All methods using full-context (CLAIRE, ITK-DReg, FireANTs) show improvement in performance with resolution, while all deep learning methods degrade in performance due to (a) progressive domain shift at higher resolutions, even for models trained on multiple or synthetic data, and (b) unlike image registration for histology slides, volumetric datasets like these require large deformations, and patch-based methods do not provide the context to perform well at higher resolutions.
In terms of efficiency, our method is substantially more efficient, both on terms of wall clock time, and the total GPU-hours consumed.

Although CLAIRE proposes a distributed GPU framework, the usage of scaling-and-squaring (which requires performing an integral and its adjoint computation every iteration) and other line search subroutines consume a considerable amount of resources.
On the faux-OASIS dataset at full resolution, CLAIRE runs out of memory with 1 and 2 GPUs, and does not work on 3 GPUs due to indivisibility of the image size by 3.
So the minimum number of GPUs required to run CLAIRE is 4.
Our method runs on a single GPU, but does not require the image sizes to be divisible by the number of GPUs, or any other qualitative constraints, allowing researchers to simply plug in their inputs and run their workflows.
For large-scale volumetric image registration problems, our method achieves three orders of magnitude of speedup over CLAIRE while enabling multimodal support and arbitrarily loss functions of choice.

\subsection{Qualitative Results}
Qualitative results are shown in \cref{fig:mosaic_250um,fig:mosaic_500um,fig:mosaic_1mm}.
With the exception of CLAIRE, ITK-DReg, and Ours, all methods get progressively worse as the resolution increases.
\include{figures/qualitative_results_method_wise}

%% file: figures/qualitative_results_method_wise.tex
\graphicspath{{qualitative_images/}}
% -------------------------------
% 1 mm resolution
% -------------------------------
\begin{figure*}[t]
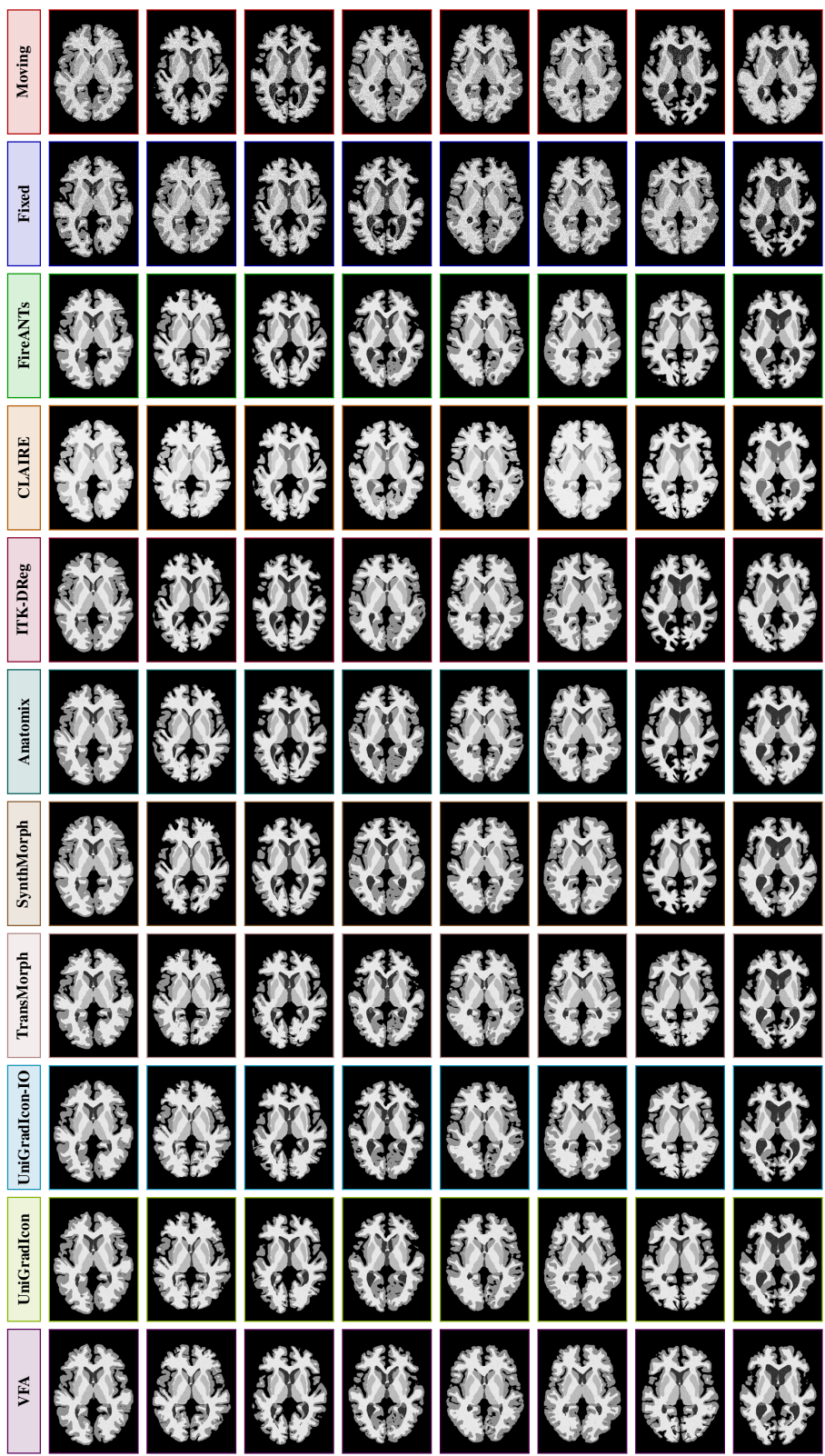

  \centering
      \scalebox{0.95}{  % Scale to 80% of original size
  \begin{minipage}{0.98\linewidth}
    \CompleteRow[red!70!black]{moving_1mm}{Moving}{Moving}\\[2pt]
    \CompleteRow[blue!70!black]{fixed_1mm}{Fixed}{Fixed}\\[2pt]
    \CompleteRow[green!60!black]{fireants_1mm}{FireANTs}{FireANTs}\\[2pt]
    \CompleteRow[orange!70!black]{claire_1mm}{CLAIRE}{CLAIRE}\\[2pt]
    \CompleteRow[purple!70!black]{itk_dreg_1mm}{itk\_dreg}{ITK-DReg}\\[2pt]
    \CompleteRow[teal!70!black]{anatomix_1mm}{Anatomix}{Anatomix}\\[2pt]
    \CompleteRow[brown!70!black]{synthmorph_1mm}{SynthMorph}{SynthMorph}\\[2pt]
    \CompleteRow[pink!70!black]{transmorph_1mm}{TransMorph}{TransMorph}\\[2pt]
    \CompleteRow[cyan!70!black]{unigradicon-io_1mm}{UniGradIcon-IO}{UniGradIcon-IO}\\[2pt]
    \CompleteRow[lime!70!black]{unigradicon_1mm}{UniGradIcon}{UniGradIcon}\\[2pt]
    \CompleteRow[violet!70!black]{vfa_1mm}{VFA}{VFA}
    \vspace{4pt}
\caption{Qualitative comparison of registration results at \textbf{1 mm}. 
Each row corresponds to the moving image, fixed image, or one of the registration methods, 
with 8 representative slices per row. The comparisons illustrate visual alignment quality 
and anatomical consistency across methods.}
    \label{fig:mosaic_1mm}
  \end{minipage}
  }
\end{figure*}

% -------------------------------
% 500 µm resolution
% -------------------------------
\begin{figure*}[t]
  \centering
    \scalebox{0.95}{  % Scale to 80% of original size
  \begin{minipage}{0.98\linewidth}
    \CompleteRow[red!70!black]{moving_500um}{Moving}{Moving}\\[2pt]
    \CompleteRow[blue!70!black]{fixed_500um}{Fixed}{Fixed}\\[2pt]
    \CompleteRow[green!60!black]{fireants_500um}{FireANTs}{FireANTs}\\[2pt]
    \CompleteRow[orange!70!black]{claire_500um}{CLAIRE}{CLAIRE}\\[2pt]
    \CompleteRow[purple!70!black]{itk_dreg_500um}{itk\_dreg}{ITK-DReg}\\[2pt]
    \CompleteRow[teal!70!black]{anatomix_500um}{Anatomix}{Anatomix}\\[2pt]
    \CompleteRow[brown!70!black]{synthmorph_500um}{SynthMorph}{SynthMorph}\\[2pt]
    \CompleteRow[pink!70!black]{transmorph_500um}{TransMorph}{TransMorph}\\[2pt]
    \CompleteRow[cyan!70!black]{unigradicon-io_500um}{UniGradIcon-IO}{UniGradIcon-IO}\\[2pt]
    \CompleteRow[lime!70!black]{unigradicon_500um}{UniGradIcon}{UniGradIcon}\\[2pt]
    \CompleteRow[violet!70!black]{vfa_500um}{VFA}{VFA}
    \vspace{4pt}
\caption{Qualitative comparison of registration results at \textbf{500um}. 
Each row corresponds to the moving image, fixed image, or one of the registration methods, 
with 8 representative slices per row. The comparisons illustrate visual alignment quality 
and anatomical consistency across methods.}
    \label{fig:mosaic_500um}
  \end{minipage}
  }
\end{figure*}

% -------------------------------
% 250 µm resolution
% -------------------------------
\begin{figure*}[t]
  \centering
      \scalebox{0.95}{  % Scale to 80% of original size
  \begin{minipage}{0.98\linewidth}
    \CompleteRow[red!70!black]{moving_250um}{Moving}{Moving}\\[2pt]
    \CompleteRow[blue!70!black]{fixed_250um}{Fixed}{Fixed}\\[2pt]
    \CompleteRow[green!60!black]{fireants_250um}{FireANTs}{FireANTs}\\[2pt]
    \CompleteRow[orange!70!black]{claire_500um}{CLAIRE}{CLAIRE}\\[2pt]
    \CompleteRow[purple!70!black]{itk_dreg_250um}{itk\_dreg}{ITK-DReg}\\[2pt]
    \CompleteRow[teal!70!black]{anatomix_250um}{Anatomix}{Anatomix}\\[2pt]
    \CompleteRow[brown!70!black]{synthmorph_250um}{SynthMorph}{SynthMorph}\\[2pt]
    \CompleteRow[pink!70!black]{transmorph_250um}{TransMorph}{TransMorph}\\[2pt]
    \CompleteRow[cyan!70!black]{unigradicon-io_250um}{UniGradIcon-IO}{UniGradIcon-IO}\\[2pt]
    \CompleteRow[lime!70!black]{unigradicon_250um}{UniGradIcon}{UniGradIcon}\\[2pt]
    \CompleteRow[violet!70!black]{vfa_250um}{VFA}{VFA}
    \vspace{4pt}
\caption{Qualitative comparison of registration results at \textbf{250um}. 
Each row corresponds to the moving image, fixed image, or one of the registration methods, 
with 8 representative slices per row. The comparisons illustrate visual alignment quality 
and anatomical consistency across methods.}
    \label{fig:mosaic_250um}
  \end{minipage}
  }
\end{figure*}

%% file: figures/synthetic-patches.tex
\graphicspath{{patch_images/}}
\begin{figure*}[t]
  \centering

  % -------- Quarter row --------
  \begin{subfigure}[b]{\linewidth}
    \centering
    \patchpair{quarter_fixed_1.png}{quarter_moving_1.png}\hfill
    \patchpair{quarter_fixed_2.png}{quarter_moving_2.png}\hfill
    \patchpair{quarter_fixed_3.png}{quarter_moving_3.png}\hfill
    \patchpair{quarter_fixed_4.png}{quarter_moving_4.png}
    \caption{1mm isotropic}
  \end{subfigure}

  \vspace{0.25em}

  % -------- Half row --------
  \begin{subfigure}[b]{\linewidth}
    \centering
    \patchpair{half_fixed_1.png}{half_moving_1.png}\hfill
    \patchpair{half_fixed_2.png}{half_moving_2.png}\hfill
    \patchpair{half_fixed_3.png}{half_moving_3.png}\hfill
    \patchpair{half_fixed_4.png}{half_moving_4.png}
    \caption{500$\um$ isotropic}
  \end{subfigure}

  \vspace{0.25em}

  % -------- Full row --------
  \begin{subfigure}[b]{\linewidth}
    \centering
    \patchpair{full_fixed_1.png}{full_moving_1.png}\hfill
    \patchpair{full_fixed_2.png}{full_moving_2.png}\hfill
    \patchpair{full_fixed_3.png}{full_moving_3.png}\hfill
    \patchpair{full_fixed_4.png}{full_moving_4.png}
    \caption{250$\um$ isotropic}
  \end{subfigure}

  \vspace{-0.25em}
  \caption{\textbf{Patch pairs seen during registration for patch-based methods}:
  For the 1mm isotropic images, there is only a single patch, i.e. the entire image.
  Deep learning methods utilize the global spatial context to perform accurate registration.
  At 500$\um$ isotropic, the patches still have large spatial context, but the images are out-of-distribution, leading to \textit{degraded} performance \cref{tab:exp1_allmethods_grouped}.
  At 250$\um$ isotropic, there is no meaningful spatial context and the patches are completely out-of-distribution, leading to poor performance for all patch-based methods.
  %   Patch-based registration lacks global spatial context across resolutions. 
  % Each tile shows a fixed (blue frame) and moving (red frame) patch side-by-side (central axial slice). 
  % Tiles are the highest-variance pairs according to label-histogram mismatch, illustrating how patch-based registration can attempt to align anatomically dissimilar regions due to lack
% of spatial context.
}
  \label{fig:patch_analysis}
\end{figure*}

%% file: figures/flamegraph.tex
\begin{figure}[t!]
    \centering
    \includegraphics[width=\linewidth]{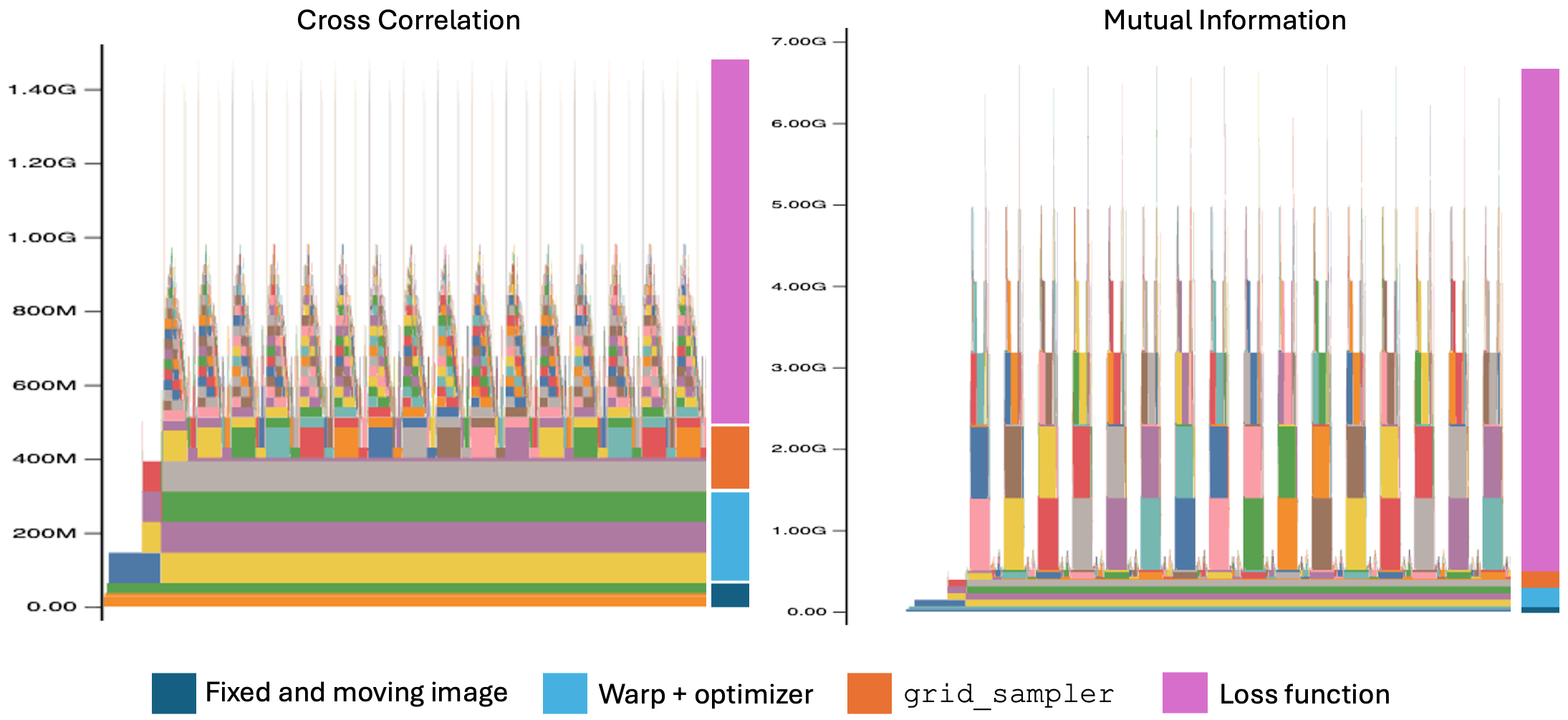}
    \vspace{0.5em}
    \caption{Flamegraph of FireANTs for Cross Correlation (left) and Mutual Information (right) losses on the OASIS dataset. 
    The flamegraph is annotated on the right with colored blocks denoting the memory overheads for the fixed and moving images, the warp field and its optimizer state, the \texttt{grid\_sampler} operation, and the loss function.
    Most of the computational overhead is due to the loss function, followed by the \texttt{grid\_sampler} operation.
    This motivates the use of fused kernels to eliminate intermediate memory overheads.
    }
    \label{fig:flamegraph}
\end{figure}